\pdfoutput=1
\documentclass[10pt,twocolumn,letterpaper]{article}

\usepackage[pagenumbers]{cvpr}

\usepackage{cuted}
\usepackage{caption}
\usepackage{subcaption}
\usepackage{booktabs}
\usepackage{multirow}
\usepackage{makecell}
\usepackage{colortbl}
\usepackage{xspace}
\usepackage{amsmath,amssymb,bm}
\usepackage{graphicx}
\usepackage{adjustbox}
\usepackage{enumitem}
\usepackage{microtype}
\graphicspath{{figures/}}

\newcommand{\ours}{ViDiHand2\xspace}
\newcommand{\vidihand}{ViDiHand\xspace}
\newcommand{\ace}{ACE-Ego-Hand\xspace}
\newcommand{\genhuman}{Gen-HumanEgo\xspace}
\newcommand{\rayfit}{Ray Fitting\xspace}

\newcommand{\best}[1]{\textbf{#1}}

\DeclareMathOperator*{\argmin}{arg\,min}
\newcommand{\R}{\mathbb{R}}
\newcommand{\vect}[1]{\bm{#1}}
\newcommand{\K}{\mathbf{K}}

\setlist{nosep,leftmargin=*}
\usepackage{tikz}
\usetikzlibrary{positioning,arrows.meta}
 
\definecolor{cvprblue}{rgb}{0.21,0.49,0.74}
\usepackage[pagebackref,breaklinks,colorlinks,allcolors=cvprblue]{hyperref}

\def\confName{CVPR}
\def\confYear{2027}

\title{\ours: Unifying Video Diffusion Priors and Hand Geometry\\for 3D Hand Motion Reconstruction in the Wild}

\author{%
Yuxi Wang$^{1*}$\quad Yufei Liu$^{2*}$\quad Chengkai Jin$^{1}$\quad Qingyue Chen$^{3}$\quad Wenqi Ouyang$^{1}$\quad Tianyi Wei$^{1}$\\
Kerui Ren$^{2}$\quad Siyuan Huang$^{2}$\quad Zhiqi Shen$^{1}$\quad Hongsheng Li$^{4}$\quad Xingang Pan$^{1\dagger}$\\[0.1em]
{\small $^{1}$Nanyang Technological University\quad $^{2}$Shanghai Jiao Tong University\quad $^{3}$Wuhan University\quad
$^{4}$The Chinese University of Hong Kong}\\[-0.1em]
{\small $^{*}$Equal contribution\quad $^{\dagger}$Corresponding author}%
}
\makeatletter
\def\@maketitle{\newpage\null\vskip 0pt
  \begin{center}{\Large\bf \@title\par}\vspace*{8pt}%
  {\large\lineskip .4em\begin{tabular}[t]{c}\@author\end{tabular}\par}\vspace*{2pt}\end{center}}
\makeatother

\usepackage{float}
\usepackage{placeins}
\makeatletter\patchcmd{\l@section}{1.0em}{0.40em}{}{}\makeatother

\renewcommand{\topfraction}{0.95}\renewcommand{\textfraction}{0.03}\renewcommand{\floatpagefraction}{0.9}
\makeatletter\@clubpenalty=10000\makeatother
\begin{document}
\addtocontents{toc}{\protect\setcounter{tocdepth}{-1}}
\maketitle
\begin{strip}
\centering
\vspace{-1.6em}
\noindent\includegraphics[width=\linewidth]{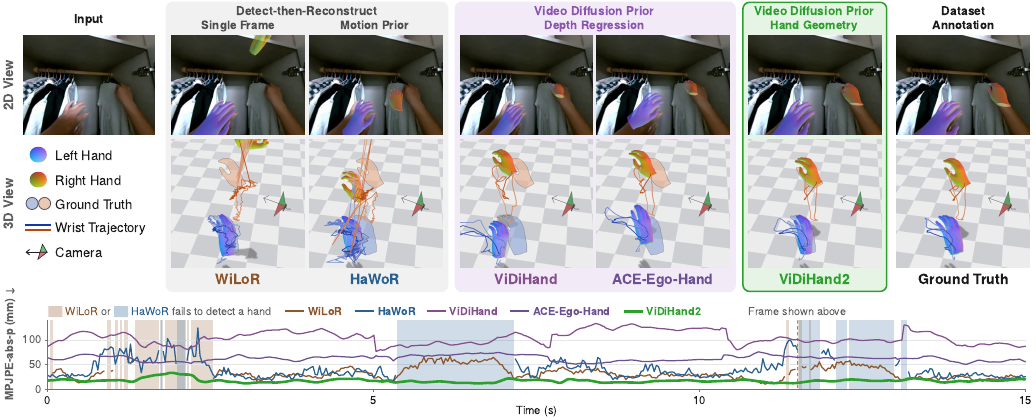}\par\vspace{0.3em}
\captionsetup{hypcap=false}
\captionof{figure}{\textbf{An in-the-wild clip of hands reaching for clothes in a wardrobe.} The 2D view shows the mesh overlays, the 3D view the
hands with the ground truth in light colors, and the plot the MPJPE-abs-p over 15\,s, with shaded bands where
WiLoR~\cite{potamias2025wilor} or HaWoR~\cite{zhang2025hawor} fails to detect a hand. \vidihand~\cite{wang2026vidihand} and
\ace~\cite{liu2026aceegohand}, which regress the depth, place the hands far from the ground truth in 3D, whereas \ours stays closest to it.}
\label{fig:teaser}
\captionsetup{hypcap=true}
\vspace{-0.6em}
\end{strip}
 \begin{abstract}

Learning robot manipulation from egocentric human videos requires accurate metric 3D hand motion. Recent methods leverage pretrained video diffusion model (VDM) priors to recover occluded hands directly from full-frame videos, but their appearance-based depth regression generalizes poorly to unseen cameras. In this work, we propose \ours, a geometry-driven framework that reformulates VDM-based hand depth estimation from direct regression into fine-grained geometric constraint solving. Specifically, we introduce Fine-Grained Geometric Lifting and Clip-Level Ray Fitting, which recover metric depth by enforcing geometric consistency between pixel-aligned 2D joints and the metric structure of MANO hands, while leveraging temporal observations across the clip. Since the lifting relies on these 2D joints, to improve robustness under occlusion and image truncation, we further develop \mbox{Confidence-Gated} Fusion to adaptively combine geometrically lifted joints with MANO predictions according to localization confidence. Finally, we enhance the VDM backbone to better capture spatial-temporal representations and produce the per-frame 2D joint and MANO predictions that both steps rely on. Extensive experiments on seven public datasets demonstrate that \ours consistently outperforms 18 baselines, reducing in-the-wild absolute joint error by 42\% and detection failures by 8.82$\times$ compared with the best baseline on each metric, while running at 57 FPS. The project page is available at \url{https://vidihand2.github.io/}.

\end{abstract}

 \section{Introduction}
\label{sec:intro}

Human videos offer a scalable source of training data for robot manipulation~\cite{kareer2025egomimic,yang2025egovla,luo2025beingh0,zheng2026egoscale} and egocentric world models~\cite{wang2026hand2world,yu2025egosim,tu2025playerone}. Accurate metric 3D hand motion is essential for extracting high-quality supervision from these videos. Egocentric hand motion reconstruction aims to recover the 3D joint positions of both hands in camera coordinates. Detect-then-reconstruct methods~\cite{pavlakos2024hamer,potamias2025wilor,zhang2025hawor} first localize hands and then estimate their 3D poses from image crops. Within this paradigm, single-frame approaches lack temporal consistency, while motion-prior-based methods model temporal coherence but remain vulnerable to localization and tracking failures under occlusion.
The current
state-of-the-art methods \vidihand~\cite{wang2026vidihand} and \ace~\cite{liu2026aceegohand} instead adapt pretrained video diffusion models (VDMs) to reconstruct hands directly from full-frame videos, eliminating the need for explicit hand detection and enabling recovery of occluded hand motion.

\begin{figure}[t]
\centering
\includegraphics[width=\linewidth]{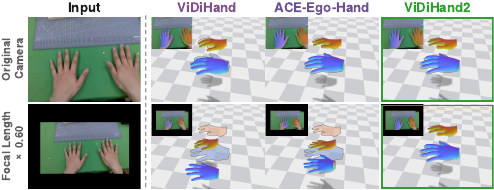}
\caption{\textbf{Sensitivity to the focal length.} An in-domain frame under its original camera and with the video and
intrinsics rescaled by 0.60. \vidihand and \ace, which regress the depth, misplace the 3D hands after rescaling, while
\ours stays closest to the ground truth. Insets show the mesh overlays.}
\label{fig:focal_intro}
\end{figure}

However, both methods exhibit substantial errors in recovering metric 3D hand trajectories from in-the-wild videos~\cite{genrobot2026genhumanego}, particularly in depth estimation and spatial localization, as \cref{fig:teaser} shows. A key limitation is that directly regressing metric depth from appearance requires learning a complex mapping that is highly sensitive to focal length and generalizes poorly to unseen cameras~\cite{yin2023metric3d,piccinelli2024unidepth}, as illustrated in \cref{fig:focal_intro}. Moreover, occlusion and hand truncation at image boundaries weaken the visual cues for hand localization, while coarse tokenization in both methods limits the preservation of fine-grained spatial details. Addressing these challenges further calls for richer spatial-temporal representations from VDMs and more consistent hand trajectory prediction over time.

In this work, we propose \ours, which reformulates VDM-based hand depth estimation from direct regression into geometric constraint solving. Our key insight is that metric depth can be recovered by enforcing geometric consistency between pixel-aligned 2D joint predictions and the metric structure of MANO hands~\cite{romero2017mano}, both of which are more reliable than directly regressed depth.  %

To realize this idea, we introduce Fine-Grained Geometric Lifting (\cref{sec:depth}), which places each joint along its camera ray and constrains its depth using MANO joint distances. Unlike prior methods~\cite{iqbal2018latent25d,sarandi2021metrabs,valassakis2024handdgp} that recover metric scale through bone-length fitting or 2D–3D alignment, our approach exploits fine-grained constraints across multiple joints. We further introduce Clip-Level Ray Fitting to jointly estimate depth across a video clip, leveraging temporal observations to constrain frames with unreliable joints.

Accurate joint localization is essential for maintaining reliable geometric constraints, particularly under occlusion and truncation. In \cref{sec:loc}, we introduce the Joint-Heatmap Refiner to recover fine-grained joint positions using heat maps at eight times the resolution of the VDM token grid. To further handle ambiguous visual evidence, Confidence-Gated Fusion adaptively combines the refined 2D predictions with the fitted MANO hand, assigning greater weight to MANO estimates when joint localization is uncertain.

We further enhance the VDM-based architecture with improved feature aggregation and temporal modeling in \cref{sec:backbone}. The Multi-Level Latent Encoder aggregates features across all adapted VDM blocks to exploit multi-level representations. The Hand Latent Decoder decodes each latent frame into its video frames, while Autoregressive Rollout Adaptation lets each chunk attend to the previous one.

Extensive experiments demonstrate the effectiveness of \ours. We consistently outperform existing methods on every metric. On in-the-wild data, \ours reduces absolute joint error by over 40\% compared with the strongest baseline. On in-domain benchmarks, it lowers the absolute and 2D joint errors of \ace by 20\% and 31\%.

 \begin{figure*}[t]
\centering
\includegraphics[width=\linewidth]{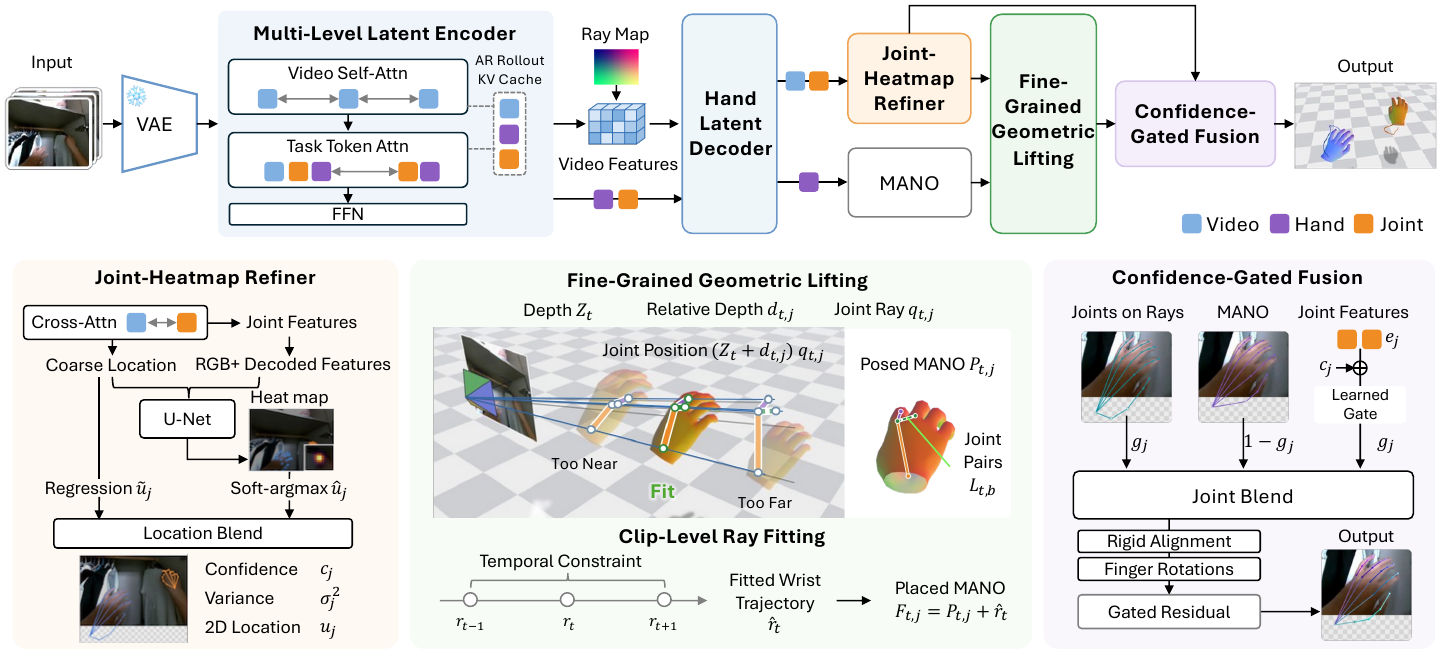}
\caption{\textbf{Overview of \ours.} 
The Multi-Level Latent Encoder and the Hand Latent Decoder turn the input clip into video features
and hand and joint tokens. In autoregressive (AR) rollouts, each chunk also attends to the key-value (KV) cache of the previous chunk.
The Joint-Heatmap Refiner predicts the 2D joint locations, and MANO gives the posed hand. Fine-Grained Geometric Lifting places this
hand in camera coordinates, and Confidence-Gated Fusion fuses it with the 2D locations. In the bottom row, the refiner blends heat-map
and regressed locations. The lifting fits the depth $Z_t$ at which the joint pairs on the rays keep their MANO distances $L_{t,b}$.
Clip-Level Ray Fitting adds a temporal constraint. The fusion blends the joints on the rays and the MANO joints with a gate $g_j$.
}
\label{fig:pipeline}
\end{figure*}
 Our contributions are summarized as follows:
\begin{itemize}[leftmargin=*,itemsep=1pt,topsep=2pt]

\item We identify a key limitation of existing VDM-based hand reconstruction methods: direct depth regression generalizes poorly to unseen cameras. To address this, we reformulate depth estimation as fine-grained geometric constraint solving, recovering metric depth by enforcing geometric consistency between pixel-aligned joint predictions and MANO hand structure.

\item To support reliable geometric reconstruction, we introduce fine-grained joint localization and confidence-gated fusion for handling occlusion and truncation, along with multi-level feature aggregation and autoregressive rollout for stronger spatial-temporal modeling.

\item We conduct extensive experiments on seven public datasets against 18 baselines. Evaluated on all ground-truth hands, \ours ranks first on every metric of every dataset, both against their officially released weights and against the re-trained baselines, showing superior detection, 3D pose accuracy and smoothness.

\end{itemize}
 \section{Related Work}
\label{sec:related}

\noindent\textbf{3D Hand Reconstruction.}
Detect-then-reconstruct methods regress MANO~\cite{romero2017mano} parameters or mesh vertices from each detected hand
crop~\cite{rong2021frankmocap,lin2021metro,lin2021meshgraphormer,park2022handoccnet,moon2022hand4whole,moon2023interwild,
zhou2024simplehand,pavlakos2024hamer,dong2024hamba,potamias2025wilor,si2026anyhand,prakash2024wildhands,xie2026egohandicl}.
Video methods aggregate crops over
time~\cite{fu2023deformer,lin2026omnihands}, learn motion
priors~\cite{duran2024hmp,ye2026haptic,ismayilzada2026padhand,xu2026handflow,zhu2026mint} or estimate world-space
motion~\cite{yu2025dynhamr,zhang2025hawor,ye2023slahmr,shin2024wham,wang2024tram,ren2026infinihand}.
Beyond pose and shape, the hand must also be placed in the camera frame. A common practice is to regress a camera translation alongside the hand under an assumed focal length~\cite{pavlakos2024hamer,potamias2025wilor}. Other methods compute the placement by geometry, from image-scale cues~\cite{moon2019rootnet}, bone lengths~\cite{iqbal2018latent25d,pavlakos2017volumetric}, or a per-frame least-squares fit of metric 3D joints to the 2D joints inside a detected crop~\cite{sarandi2021metrabs,valassakis2024handdgp,millerdurai2026egoforce}. 
\ours also uses the geometry, but fits the predicted MANO hand to the camera rays of its joints using metric joint-pair distances and reprojection constraints. The fitting requires no hand crop, spans an entire video clip, and includes joints beyond the image boundaries.

\noindent\textbf{Parametric Models and Keypoints.}
Parametric models such as SMPL~\cite{loper2015smpl} and
MANO~\cite{romero2017mano} accumulate rotation errors along the kinematic
chain. To avoid this, HybrIK~\cite{li2021hybrik}, NIKI~\cite{li2023niki}
and Minimal-Hand~\cite{zhou2022mono} derive the rotations from
predicted 3D keypoints through inverse kinematics.
MeTRAbs~\cite{sarandi2021metrabs} uses the back-projected 2D location for
joints inside the image and the root-relative 3D prediction for truncated
ones. \ours replaces this hard switch with learned gates that move the
placed MANO hand toward the 2D locations according to their reliability.

\noindent\textbf{Diffusion Models for Perception.}
Recent work adapts pretrained diffusion models to correspondence, depth
estimation, tracking and 3D
perception~\cite{tang2023dift,ke2024marigold,fu2024geowizard,he2025lotus,hu2025depthcrafter,nam2025difftrack,son2025ditracker,yuan2025denoisetotrack,huang2026howmuch3d,velez2025imagetovideo}.
\vidihand~\cite{wang2026vidihand} and \ace~\cite{liu2026aceegohand}
reconstruct hands from video diffusion features without a hand detector,
but both regress the hand depth. In monocular depth estimation and human
mesh recovery, a regressed metric depth is known to depend on the
training cameras~\cite{yin2023metric3d,piccinelli2024unidepth,li2022cliff}.
\ours therefore replaces this regression by a fit along the rays of the
test camera.
\section{\ours}
\label{sec:method}

\subsection{Overview}
\label{sec:overview}

Given only a monocular egocentric video clip $\mathcal{V}=\{I_t\}_{t=1}^{T}$ and the camera intrinsics $\K_t$,
we reconstruct both hands in every frame. For each hand, we estimate the MANO~\cite{romero2017mano} pose $\vect{\theta}_t$, the shape
$\vect{\beta}$ and the 21 joints $\vect{X}_{t,j}\in\R^3$ in metric camera coordinates. \ours reformulates hand depth estimation from regression
into geometric constraint solving. Specifically, we predict the MANO hand and the 2D joint locations with a network $\mathcal{F}_\phi$ and
fit the wrist trajectory,
\begin{equation}
\begin{aligned}
&\big(\vect{\theta}_{1:T},\vect{\beta},\vect{u}_{1:T}\big)=\mathcal{F}_\phi\big(I_{1:T},\K_{1:T}\big),\\
&\hat{\vect{r}}_{1:T}=\argmin_{\vect{r}_{1:T}}\,E\big(\vect{r}_{1:T};\,\vect{\theta}_{1:T},\vect{\beta},\vect{u}_{1:T},\K_{1:T}\big),
\end{aligned}
\label{eq:task}
\end{equation}
where $\vect{u}_t$ denotes the 2D joint locations, predicted with their variances and confidences, $\vect{r}_t$ the wrist position,
whose depth is the hand depth, and $E$ the energy of \cref{eq:rayfit}, which has a bone term, a reprojection term and a temporal term. We omit the hand index and write $\K$ for $\K_t$. MANO takes the pose
$\vect{\theta}_t$ and one shape $\vect{\beta}$ per clip and returns the wrist-relative joints $\vect{P}_{t,j}$ with the metric bone
lengths on which the fit relies.

Building on pretrained VDMs, \vidihand and \ace reconstruct hands from full-frame videos without a hand detector. Both predict
the MANO hand and the 2D joint locations from VDM features and regress the hand depth. \ours revisits this design starting from
the hand depth, as \cref{fig:pipeline} shows. \Cref{sec:depth} replaces the depth regression with a geometric fit. Because this
fit depends on accurate 2D evidence, \cref{sec:loc} localizes the joints beyond the resolution of the token grid, even on occluded or
truncated hands, and fuses them with the placed MANO hand into the output joints $\vect{X}_{t,j}$. Finally, \cref{sec:backbone}
strengthens the VDM-based architecture for the features that this localization requires.
\subsection{Hand Depth Estimation Revisited}
\label{sec:problem}
\label{sec:problem_model}
\label{sec:depth}
\label{sec:exp_depth}

\looseness=-1 We begin by revisiting the mixed projection of \vidihand and \ace~\cite{wang2026vidihand,liu2026aceegohand}, which regresses only
the depth and solves the in-plane shift in closed form. We pinpoint the depth error to this regression. Regressing a metric depth
requires learning the metric scale of all visible objects, which is hard for general
scenes~\cite{yin2023metric3d,piccinelli2024unidepth}. The learned mapping from appearance to depth thus depends on the focal
lengths of the training cameras, as \cref{fig:focal_intro} shows. We therefore seek a depth estimate that is easier to
learn and follows the test camera.

\noindent\textbf{Fine-Grained Geometric Lifting.}
Let us start by considering the 2D joint location, which is pixel-aligned and easier to learn than the depth. With $\K$, the location
$\vect{u}_{t,j}$ of joint $j$ in frame $t$ from \cref{sec:loc} defines the joint ray
\begin{equation}
\vect{q}_{t,j}=\K^{-1}[\vect{u}_{t,j};1],
\label{eq:ray}
\end{equation}
but not the depth along this ray. Over the central 98\% of our training hands, the depth varies by 3.37$\times$, whereas the metric
hand size varies by 1.28$\times$. A bone of metric length $L$ parallel to the image plane has the image length $h=fL/Z$ at
depth $Z$ and focal length $f$, so that
\begin{equation}
Z=\frac{fL}{h},\qquad\frac{\delta Z}{Z}\approx\frac{\delta L}{L}-\frac{\delta h}{h}
\label{eq:scale}
\end{equation}
for small errors $\delta$. This depth follows the test camera through $f$. Its relative error comes only from those of $L$ and
$h$. We therefore propose fine-grained geometric lifting, which computes the hand depth from the 2D joint locations and the metric
MANO hand. Every joint lies at its own depth on its own ray and is weighted by its 2D uncertainty, even if occluded. The
resulting depth is more accurate than a regressed one, as \cref{sec:ablation} confirms. Prior geometric
lifting~\cite{iqbal2018latent25d,sarandi2021metrabs,valassakis2024handdgp} solves each frame inside a detected crop with one bone
length or a reprojection fit. We instead use the full frame with the true intrinsics and keep the joints beyond the border.
Besides the reprojection, we fit the depth to many metric joint distances.

\noindent\textbf{Clip-Level Ray Fitting.}
We solve this lifting for the wrist trajectory $\vect{r}_{1:T}$ of a whole clip with one energy. Let $d_{t,j}$ and $Z_t$ be the depth
components of $\vect{P}_{t,j}$ and $\vect{r}_t$, respectively. Joint $j$ then lies at $(Z_t+d_{t,j})\,\vect{q}_{t,j}$. For a joint
pair $b=(a,p)$, the residual compares the distance between the joints on their rays with their distance $L_{t,b}$ in the posed
MANO hand,
\begin{equation}
\rho_{t,b}=\big\|(Z_t{+}d_{t,a})\,\vect{q}_{t,a}-(Z_t{+}d_{t,p})\,\vect{q}_{t,p}\big\|-L_{t,b} .
\label{eq:bone}
\end{equation}
The pairs are the 20 bones, the knuckle pairs and the pairs of the wrist with every non-adjacent joint. The energy is
\begin{equation}
\begin{aligned}
E(\vect{r}_{1:T})={}&\sum_{t,b} w^{\mathrm{bone}}_{t,b}\,\psi\big(\rho_{t,b}\big)\\
&+\sum_{t,j} w^{\mathrm{proj}}_{t,j}\,\psi\big(\big\|\pi(\vect{P}_{t,j}{+}\vect{r}_t)-\vect{u}_{t,j}\big\|\big)\\
&+\sum_{t}\big\|\vect{\Lambda}_t\,(\vect{r}_{t-1}-2\vect{r}_t+\vect{r}_{t+1})\big\|^2 ,
\end{aligned}
\label{eq:rayfit}
\end{equation}
where $\psi$ is the Huber penalty and $\pi$ the perspective projection with $\K$. The bone term depends on the wrist only through
$Z_t$ and constrains the depth in metric units. With a depth per joint, this term also holds for bones inclined to the image plane,
for which \cref{eq:scale} overestimates the depth~\cite{iqbal2018latent25d}. By aligning the projected MANO hand with the 2D
locations, the reprojection term fixes the lateral position and adds a depth constraint in pixels. The temporal term penalizes the
wrist acceleration with the weights $\vect{\Lambda}_t$, so that neighboring frames support a frame with few reliable joints. Both
$w^{\mathrm{bone}}_{t,b}$ and $w^{\mathrm{proj}}_{t,j}$ decrease with the 2D variances involved. We minimize $E$
with damped Gauss--Newton iterations. Two small heads correct the 2D variances and the lateral weights of $\vect{\Lambda}_t$. A loss
on the output hand trains these heads by implicit differentiation through the minimizer of
$E$~\cite{chen2020bpnp,chen2022epropnp}. The energy contains no regressed absolute depth or depth prior and uses $\K$ only
through the joint rays and $\pi$. The fitted trajectory gives the placed MANO hand $\vect{F}_{t,j}=\vect{P}_{t,j}+\hat{\vect{r}}_t$.
\subsection{Joint Localization Revisited}
\label{sec:loc}
\label{sec:evidence}
\label{sec:exp_loc}
\begin{figure}[t]
\centering
\includegraphics[width=0.86\linewidth]{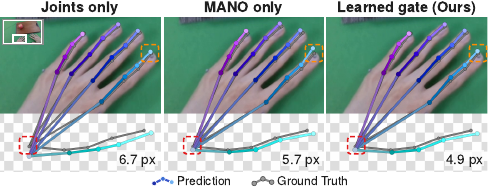}
\caption{\textbf{Confidence-Gated Fusion on a hand truncated by the image border.} Projected joints of the variants of \cref{tab:abl_fusion}
and the ground truth, shown in gray, with mean 2D errors. The checkerboard lies beyond the image. The fusion follows the 2D location
where the MANO finger leaves the image finger in the orange box, and keeps the MANO hand beyond the border in the red box.}
\label{fig:fusion}
\end{figure}
The video prior of \vidihand and \ace maintains a 3D hand state through occlusion, but an occluded hand offers little
evidence in the current frame. The key issue is that the joints in view cover few cells of a 16- or 32-pixel token grid.
Joints hidden by an object or the other hand have no pixels of their own. Truncation by the image border is the extreme case. By
\cref{eq:scale}, the lifting of such a hand relies on the few reliable 2D locations. We therefore localize every joint inside the image on heat maps eight times finer than our 32-pixel grid and regress the others. We then fuse them
with the placed MANO hand, which covers the hidden joints.

\noindent\textbf{Joint-Heatmap Refiner.}
We refine the coarse location of each joint on heat maps with 4-pixel cells. We omit the index $t$. Each joint token gives this
coarse location and the joint feature $\vect{e}_j$ by attending to the feature map of \cref{sec:backbone}. From the RGB frame and
the decoded features, we compute 42 heat maps, which a U-Net refines jointly from Gaussians at the coarse locations.
A null score models joints outside the image. A soft-argmax~\cite{sun2018integral,nibali2018dsnt} around the
peak~\cite{bulat2021subpixel} yields the heat-map location $\hat{\vect{u}}_j$ and its variance. Since heat maps end at the image
border~\cite{sarandi2021metrabs,li2021rle}, a regression branch predicts an unbounded location $\tilde{\vect{u}}_j$ so that
joints beyond the border also have a ray. We blend the two locations,
\begin{equation}
\vect{u}_j=\alpha_j\,\hat{\vect{u}}_j+(1-\alpha_j)\,\tilde{\vect{u}}_j ,
\label{eq:read}
\end{equation}
where $\alpha_j\in[0,1]$ decreases near the border and on uncertain heat maps. A confidence $c_j$ predicts whether $\hat{\vect{u}}_j$
lies within 16 pixels of the ground truth. The location $\vect{u}_j$ defines the joint ray of \cref{eq:ray}, its variance
$\sigma_j^2$ sets the weights of \cref{eq:rayfit}, and $c_j$ enters the gate of \cref{eq:gate}.

\noindent\textbf{Confidence-Gated Fusion.}
\label{sec:fusion}
We fuse the placed MANO hand with the 2D locations through learned gates. Each joint has two estimates at the same depth
$Z+d_j$: the placed MANO joint $\vect{F}_j=\vect{P}_j+\hat{\vect{r}}$ and the back-projected 2D location $(Z+d_j)\,\bar{\vect{q}}_j$.
The ray $\bar{\vect{q}}_j$ comes from \cref{eq:read} evaluated on heat maps pooled over neighboring frames. A gate
$g_j\in[0,1]$ blends the two,
\begin{equation}
\vect{B}_j=g_j\,(Z+d_j)\,\bar{\vect{q}}_j+(1-g_j)\,\vect{F}_j,
\label{eq:lock}
\end{equation}
which moves each joint only parallel to the image plane. The gate recalibrates $c_j$ and adds a correction from $\vect{e}_j$,
\begin{equation}
g_j=\mathrm{sigmoid}\big(\gamma_1\,\mathrm{logit}(c_j)+\gamma_0+\vect{v}^\top\vect{e}_j\big).
\label{eq:gate}
\end{equation}
The gate thus weighs each joint by the reliability of its 2D location, not its visibility. Where this reliability is low, as beyond
the border in the red box of \cref{fig:fusion}, the joint follows the MANO hand. Where accumulated rotation errors along the
kinematic chain~\cite{li2021hybrik} place the MANO finger off the image finger, the joint follows the 2D location, as in the
orange box. The blend treats each joint independently. We therefore move the MANO hand toward $\vect{B}$ by a rigid motion and
finger rotations, which preserve bone lengths. Each finger joint then moves by a gated share of the remaining offset, without
preserving bone lengths. The supplement details both steps in \cref{S-sec:supp_fusion}. All gates are trained on the 3D joint
error. Closing them returns the placed MANO hand. All metrics use the output joints $\vect{X}_j$.
 \begin{table*}[t]
\centering
\caption{\textbf{Comparison with existing methods.} All 18 baselines are re-trained with their official code on our six training splits.
The suffix -p adds a penalty for each detection failure. FAcc and F1 in \%, EPE2D-p in px, Jitter in mm/frame$^2$,
other errors in mm, best values in bold. \ours is the \textbf{best} on every metric, also against the officially released weights
in \cref{S-sec:supp_official} of the supplement.}
\label{tab:main}
\setlength{\tabcolsep}{3.2pt}\renewcommand{\arraystretch}{0.90}
\resizebox{\textwidth}{!}{%
\begin{tabular}{@{}lcccccccccccccc@{}}
\toprule
& \multicolumn{7}{c}{In-domain: mean of ARCTIC, H2O, HOT3D, OakInk2, Ego-Pro, Ego-Std} & \multicolumn{7}{c}{In-the-wild: \genhuman} \\
\cmidrule(lr){2-8}\cmidrule(lr){9-15}
& \multicolumn{2}{c}{Detection} & \multicolumn{2}{c}{Absolute pose} & \multicolumn{2}{c}{Relative pose} & \multicolumn{1}{c}{Temporal} & \multicolumn{2}{c}{Detection} & \multicolumn{2}{c}{Absolute pose} & \multicolumn{2}{c}{Relative pose} & \multicolumn{1}{c}{Temporal} \\
\cmidrule(lr){2-3}\cmidrule(lr){4-5}\cmidrule(lr){6-7}\cmidrule(lr){8-8}\cmidrule(lr){9-10}\cmidrule(lr){11-12}\cmidrule(lr){13-14}\cmidrule(lr){15-15}
Method & FAcc$\uparrow$ & F1$\uparrow$ & MPJPE-abs-p$\downarrow$ & EPE2D-p$\downarrow$ & MPJPE-p$\downarrow$ & PA-p$\downarrow$ & Jitter$\downarrow$ & FAcc$\uparrow$ & F1$\uparrow$ & MPJPE-abs-p$\downarrow$ & EPE2D-p$\downarrow$ & MPJPE-p$\downarrow$ & PA-p$\downarrow$ & Jitter$\downarrow$ \\
\midrule
\multicolumn{15}{@{}l}{\textit{Detect-then-reconstruct, single frame}} \\
Hamba~\cite{dong2024hamba} & 61.90 & 82.20 & 101.43 & 47.14 & 36.02 & 15.60 & 11.97 & 70.73 & 88.73 & 121.17 & 31.84 & 39.15 & 14.41 & 15.45 \\
WildHands~\cite{prakash2024wildhands} & 83.55 & 92.59 & 80.92 & 34.80 & 36.66 & 12.77 & 16.42 & 92.46 & 96.86 & 106.70 & 21.43 & 39.70 & 9.51 & 15.69 \\
InterWild~\cite{moon2023interwild} & 84.07 & 93.46 & 67.76 & 21.09 & 28.54 & 10.14 & 22.31 & 92.15 & 95.79 & 100.85 & 25.56 & 37.74 & 9.35 & 26.98 \\
AnyHand~\cite{si2026anyhand} & 92.26 & 97.21 & 65.54 & 14.98 & 18.05 & 8.56 & 10.20 & 91.93 & 96.90 & 96.46 & 15.17 & 28.23 & 9.51 & 15.91 \\
HaMeR~\cite{pavlakos2024hamer} & 84.76 & 93.18 & 73.06 & 25.74 & 22.50 & 10.11 & 13.99 & 94.65 & 97.61 & 86.82 & 15.60 & 26.49 & 9.02 & 18.87 \\
EgoForce~\cite{millerdurai2026egoforce} & 84.25 & 94.17 & 68.90 & 27.51 & 25.77 & 10.37 & 13.92 & 84.87 & 95.10 & 79.94 & 27.94 & 32.98 & 11.29 & 19.11 \\
HandDGP~\cite{valassakis2024handdgp} & 92.26 & 97.21 & 71.03 & 19.52 & 35.04 & 13.75 & 12.80 & 91.98 & 96.88 & 71.18 & 17.16 & 36.74 & 13.15 & 16.15 \\
EgoHandICL~\cite{xie2026egohandicl} & 92.27 & 97.21 & 46.89 & 13.29 & 17.14 & 7.89 & 12.32 & 91.94 & 96.89 & 57.07 & 14.55 & 28.21 & 9.66 & 15.05 \\
WiLoR~\cite{potamias2025wilor} & 92.26 & 97.21 & 47.10 & 13.01 & 16.89 & 7.83 & 11.17 & 91.95 & 96.89 & 56.55 & 14.45 & 28.07 & 9.76 & 14.61 \\
\midrule
\multicolumn{15}{@{}l}{\textit{Detect-then-reconstruct with a motion prior}} \\
Dyn-HaMR~\cite{yu2025dynhamr} & 92.50 & 97.34 & 159.19 & 14.72 & 33.25 & 8.00 & 12.05 & 96.31 & 98.06 & 151.24 & 15.46 & 50.84 & 10.51 & 17.48 \\
OmniHands~\cite{lin2026omnihands} & 77.89 & 89.62 & 99.61 & 29.19 & 27.69 & 10.43 & 53.55 & 90.77 & 95.25 & 104.65 & 21.70 & 31.03 & 10.01 & 42.21 \\
HaWoR~\cite{zhang2025hawor} & 76.91 & 86.91 & 74.04 & 29.07 & 31.11 & 12.24 & 22.19 & 76.03 & 83.49 & 96.69 & 32.10 & 42.60 & 14.98 & 27.85 \\
HandFlow~\cite{xu2026handflow} & 92.52 & 97.49 & 44.31 & 17.11 & 21.35 & 8.96 & 13.98 & 92.54 & 97.50 & 69.53 & 16.06 & 27.57 & 9.70 & 17.00 \\
MINT~\cite{zhu2026mint} & 98.80 & 99.54 & 54.07 & 22.41 & 19.82 & 9.08 & 9.64 & 97.80 & 99.00 & 68.74 & 16.67 & 27.73 & 8.06 & 8.58 \\
PAD-Hand~\cite{ismayilzada2026padhand} & 91.39 & 96.90 & 70.75 & 24.53 & 21.53 & 9.30 & 12.23 & 91.50 & 96.70 & 66.46 & 22.90 & 29.33 & 10.12 & 12.51 \\
InfiniHand~\cite{ren2026infinihand} & 98.77 & 99.52 & 40.67 & 8.32 & 15.30 & 6.11 & 13.06 & 98.03 & 99.24 & 59.91 & 10.17 & 24.39 & 9.42 & 22.65 \\
\midrule
\multicolumn{15}{@{}l}{\textit{Video diffusion with depth regression}} \\
ViDiHand~\cite{wang2026vidihand} & 98.98 & 99.59 & 36.60 & 9.92 & 18.64 & 8.29 & 2.93 & 99.03 & 99.60 & 94.53 & 11.84 & 29.30 & 7.45 & 4.08 \\
ACE-Ego-Hand~\cite{liu2026aceegohand} & 98.88 & 99.64 & 24.46 & 6.81 & 10.84 & 5.60 & 1.44 & 95.99 & 97.65 & 92.13 & 26.74 & 22.98 & 8.73 & 1.76 \\
\midrule
\multicolumn{15}{@{}l}{\textit{Video diffusion with hand geometry}} \\
\textbf{\ours (Ours)} & \best{99.93} & \best{99.98} & \best{19.62} & \best{4.67} & \best{9.50} & \best{4.92} & \best{1.26} & \best{99.89} & \best{99.96} & \best{32.59} & \best{6.50} & \best{18.59} & \best{6.99} & \best{1.61} \\
\bottomrule
\end{tabular}
 }
\end{table*}
 \begin{table}[!t]
\centering
\caption{\textbf{Comparison on hands detected by all methods.} \ours and the re-trained baselines of \cref{tab:main},
evaluated only on the hands that all 19 methods detect, 59\% of the hand-frames. Units
as in \cref{tab:main}, best in bold. \ours is the \textbf{best} on every metric.}
\label{tab:tp}
\setlength{\tabcolsep}{3pt}
\renewcommand{\arraystretch}{0.88}
\footnotesize
\resizebox{\columnwidth}{!}{%
\begin{tabular}{@{}lcccccc@{}}
\toprule
& \multicolumn{3}{c}{In-domain} & \multicolumn{3}{c}{In-the-wild} \\
\cmidrule(lr){2-4}\cmidrule(lr){5-7}
Method & MPJPE-abs$\downarrow$ & MPJPE$\downarrow$ & Jitter$\downarrow$ & MPJPE-abs$\downarrow$ & MPJPE$\downarrow$ & Jitter$\downarrow$ \\
\midrule
\multicolumn{7}{@{}l}{\textit{Detect-then-reconstruct, single frame}} \\
Hamba~\cite{dong2024hamba} & 41.74 & 12.98 & 10.77 & 61.27 & 22.45 & 13.53 \\
WildHands~\cite{prakash2024wildhands} & 49.14 & 27.73 & 14.25 & 74.37 & 35.78 & 13.11 \\
InterWild~\cite{moon2023interwild} & 53.28 & 24.46 & 17.57 & 67.66 & 32.46 & 21.10 \\
AnyHand~\cite{si2026anyhand} & 45.81 & 12.14 & 9.04 & 65.30 & 22.31 & 12.47 \\
HaMeR~\cite{pavlakos2024hamer} & 43.55 & 13.55 & 11.44 & 61.11 & 22.59 & 13.68 \\
EgoForce~\cite{millerdurai2026egoforce} & 32.81 & 13.47 & 10.88 & 36.67 & 20.04 & 14.08 \\
HandDGP~\cite{valassakis2024handdgp} & 45.28 & 28.85 & 9.99 & 45.45 & 29.89 & 11.57 \\
EgoHandICL~\cite{xie2026egohandicl} & 28.54 & 11.16 & 10.64 & 30.67 & 22.01 & 11.49 \\
WiLoR~\cite{potamias2025wilor} & 28.89 & 10.97 & 9.88 & 30.26 & 21.88 & 11.17 \\
\midrule
\multicolumn{7}{@{}l}{\textit{Detect-then-reconstruct with a motion prior}} \\
Dyn-HaMR~\cite{yu2025dynhamr} & 148.41 & 30.22 & 10.25 & 106.99 & 47.75 & 12.58 \\
OmniHands~\cite{lin2026omnihands} & 64.96 & 19.64 & 30.63 & 68.93 & 24.93 & 28.17 \\
HaWoR~\cite{zhang2025hawor} & 36.69 & 17.80 & 21.54 & 42.87 & 26.77 & 26.26 \\
HandFlow~\cite{xu2026handflow} & 26.88 & 16.03 & 13.18 & 50.92 & 22.70 & 14.95 \\
MINT~\cite{zhu2026mint} & 48.73 & 17.99 & 9.10 & 49.32 & 26.36 & 8.02 \\
PAD-Hand~\cite{ismayilzada2026padhand} & 48.20 & 15.18 & 10.70 & 41.69 & 23.44 & 9.90 \\
InfiniHand~\cite{ren2026infinihand} & 36.38 & 13.38 & 10.15 & 50.13 & 21.84 & 15.62 \\
\midrule
\multicolumn{7}{@{}l}{\textit{Video diffusion with depth regression}} \\
ViDiHand~\cite{wang2026vidihand} & 32.63 & 17.14 & 2.67 & 90.47 & 28.33 & 3.53 \\
ACE-Ego-Hand~\cite{liu2026aceegohand} & 20.33 & 9.45 & 1.35 & 79.15 & 19.55 & 1.60 \\
\midrule
\multicolumn{7}{@{}l}{\textit{Video diffusion with hand geometry}} \\
\textbf{\ours (Ours)} & \best{17.42} & \best{8.88} & \best{1.23} & \best{28.09} & \best{17.82} & \best{1.46} \\
\bottomrule
\end{tabular}
 }
\end{table}
\subsection{VDM-Based Architecture Revisited}
\label{sec:backbone}

\noindent\textbf{Multi-Level Latent Encoder.}
\looseness=-1 We aggregate the features of every block of a pretrained VDM. Like \ace~\cite{liu2026aceegohand}, we repurpose this VDM into a
deterministic, detector-free encoder. We apply the first 16 blocks of the Wan2.2 transformer~\cite{wan2025wan}, adapted with
LoRA~\cite{hu2022lora}, to noise-free VAE latents without diffusion sampling. Each latent frame after the first covers four video frames. The
transformer processes a clip in chunks of 21 latent frames, whereas the fit of \cref{sec:depth} spans the whole clip. The task
tokens of each latent frame are two hand tokens, 42 joint tokens and four registers. In \ace, the same tokens enter a separate
decoder after the last adapted block and see only that block's features. Our task tokens instead pass through every adapted block, where
they use the block weights~\cite{wang2026genception} and read the video tokens without changing them~\cite{lee2023rpo}. The hand
tokens predict the MANO hand and the hand presence. The video tokens of the last block and a ray encoding of $\K$ form the feature
map.

\noindent\textbf{Hand Latent Decoder.}
\label{sec:decoding}%
We decode the hand states and features of each latent frame into its video frames without a fixed interpolation. Our temporal
decoders build on the Wan VAE decoder~\cite{wan2025wan} and also read latent frames of the adjacent chunks. In contrast, \ace
linearly interpolates the hand states between latent frames.
\vidihand~\cite{wang2026vidihand} adds a learned residual to this interpolation.

\noindent\textbf{Autoregressive Rollout Adaptation.}
\label{sec:exp_temporal}%
In the last four blocks of our transformer, each chunk also attends to the cached keys and values of the previous
chunk~\cite{dai2019transformerxl,wu2022memvit,zhuo2025streamvggt}. \vidihand instead processes each chunk independently, so the
hand pose can jump at chunk boundaries. On autoregressive rollouts~\cite{karaev2024cotracker,huang2025selfforcing}, we train only a
bias on this attention. The supplement details chunked inference in \cref{S-sec:supp_clip}.
 \section{Experiments}
\makeatletter\@clubpenalty=10000\makeatother
\clubpenalty=10000
\widowpenalty=10000
\displaywidowpenalty=10000
\label{sec:main}
\label{sec:exp}

\subsection{Experimental Setup}
\label{sec:setup}
\label{sec:bench}
\label{sec:problem_data}

\label{sec:bench_build}%
\label{sec:revisit_data}%
\noindent\textbf{Datasets.} We train and test on six egocentric datasets, ARCTIC~\cite{fan2023arctic}, H2O~\cite{kwon2021h2o},
HOT3D~\cite{banerjee2025hot3d}, OakInk2~\cite{zhan2024oakink2} and the two parts of EgoDemo~\cite{lightwheel2026egosuite}, EgoDemo
Pro (Ego-Pro) and EgoDemo Standard (Ego-Std), with 38.97 and 6.22 hours of training and \emph{in-domain} test video. To test
\emph{in the wild}, on an unseen camera and scenes, we sample about one hour of \genhuman~\cite{genrobot2026genhumanego}, unused
in training. Rule-based checks and visual inspection remove 3.79\% of the frames of the seven datasets, mostly with inaccurate or missing labels
or unusable images, as \cref{S-sec:supp_protocol} of the supplement details.

\label{sec:bench_protocol}%
\noindent\textbf{Baselines.}
Most methods release weights trained on different datasets, so we evaluate the baselines in two versions. For \cref{tab:main},
we re-train all baselines with their official code on the same six training splits as \ours, and \cref{S-sec:supp_official} of the
supplement evaluates their officially released weights.

\noindent\textbf{Metrics.} A detection failure is a visible ground-truth hand that a method does not detect. We report F1 and the
frame accuracy FAcc, the share of frames with neither a false positive nor a detection failure, the joint errors MPJPE-abs in
camera coordinates, MPJPE after wrist alignment and PA-MPJPE (PA) after Procrustes alignment, EPE2D of the projected joints, and
Jitter, the mean joint acceleration of a detected hand, which measures smoothness. The suffix -p averages an error over all
visible ground-truth hands and scores each detection failure as a mean MANO hand at 0.5\,m depth, or in PA-p by the ground-truth spread.

\label{sec:training}%
\noindent\textbf{Implementation.}
\ours is trained on 8 A100 GPUs with a total batch size of 8, for 50k steps, then 3k steps on autoregressive rollouts.
The hand presence threshold is 0.05, and \cref{S-sec:supp_impl} of the supplement gives all details.

\subsection{Comparison with Existing Methods}
\label{sec:comparison}
\label{sec:analysis}
\label{sec:effect_ehb}

\noindent\textbf{Comparison on All Hands with Detection Failure Penalty.} Against the 18 re-trained baselines of \cref{tab:main}
and their officially released weights in \cref{S-sec:supp_official} of the supplement, \ours is the best on every metric,
in-domain and in the wild. In the wild, it outperforms all baselines by a large margin. On MPJPE-abs-p, which penalizes detection
failures and misplaced hands alike, it reduces the error of the best baseline, WiLoR, by 42\%, from 56.55 to 32.59\,mm. In-domain,
it reduces that of \ace by 20\%, from 24.46 to 19.62\,mm. A detection failure or a false positive occurs in only 0.07\% of its
in-domain and 0.11\% of its in-the-wild frames, 14.57$\times$ and 8.82$\times$ less often than for \vidihand, the baseline with
the highest FAcc.

\begin{figure*}[t]
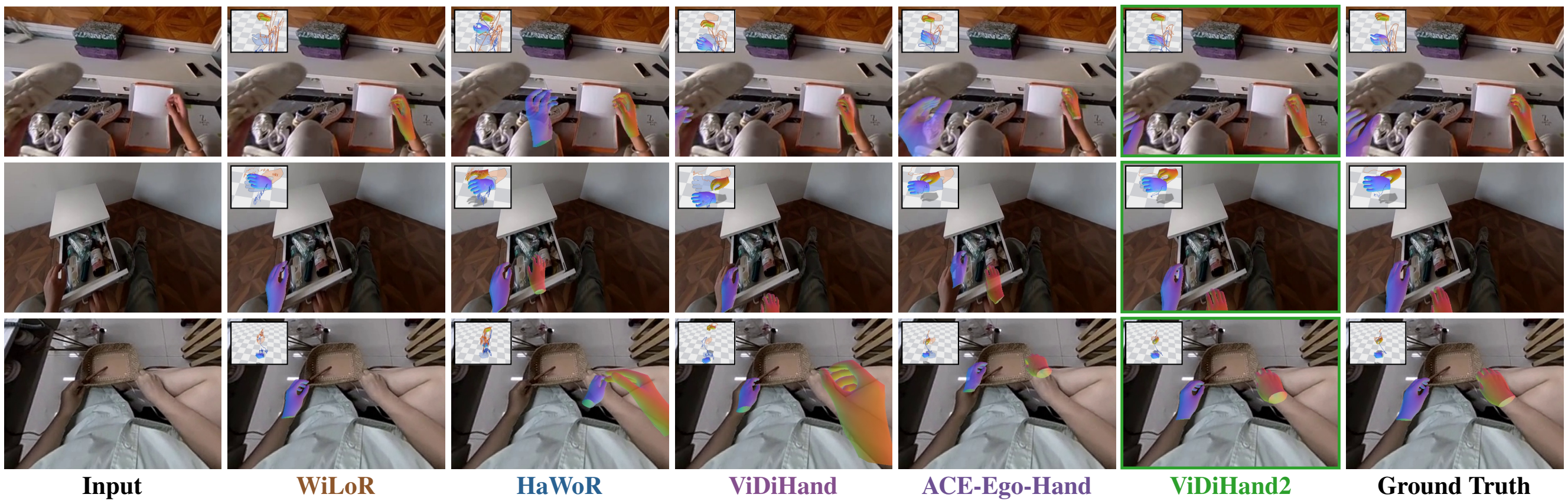

\centering
\definecolor{tvW}{HTML}{8D562B}\definecolor{tvM}{HTML}{2A5B90}\definecolor{tvA}{HTML}{6D5092}\definecolor{tvO}{HTML}{2CA02C}
\definecolor{tvV}{HTML}{844F8D}\definecolor{tvH}{HTML}{296191}
\begin{tikzpicture}[x=1pt, y=1pt, font=\footnotesize,
  img/.style={anchor=north west, inner sep=0pt},
  hdr/.style={anchor=base, inner sep=0pt, font=\footnotesize\bfseries},
  lab/.style={anchor=north west, inner sep=0.8pt, text=black!80, fill=white, fill opacity=0.82, text opacity=1,
       font=\fontsize{4.3}{4.3}\selectfont}]
\def\tw{69.22}\def\th{47.7}\def\sx{71.28}\def\sy{49.7}\def\yz{0}\def\iw{18}\def\ih{13.5}
\foreach \n/\c [count=\i from 0] in {Input/black, WiLoR/tvW, HaWoR/tvH, \vidihand/tvV, \ace/tvA, \textbf{\ours}/tvO, Ground Truth/black} {
  \node[hdr, text=\c] at (\i*\sx+0.5*\tw,-155.3) {\n};}
\foreach \case [count=\r from 0] in {genhumanego_shoe, genhumanego_drawer, genhumanego_basket} {
  \pgfmathsetmacro{\yy}{\yz-\r*\sy}
  \node[img] at (0,\yy) {\includegraphics[width=\tw pt,height=\th pt]{qualitative/\case_overlay_input.jpg}};
  \foreach \m [count=\i from 1] in {wilor, hawor, vidihand, aceegohand, ours, groundtruth} {   %
    \pgfmathsetmacro{\xx}{\i*\sx}
    \node[img] at (\xx,\yy) {\includegraphics[width=\tw pt,height=\th pt]{qualitative/\case_overlay_\m.jpg}};
    \node[img] at (\xx+1.2,\yy-1.2) {\includegraphics[width=\iw pt,height=\ih pt]{qualitative/\case_3d_\m.jpg}};
    \draw[black, line width=0.4pt] (\xx+1.2,\yy-1.2) rectangle ++(\iw,-\ih);   %
    }
  \draw[tvO, line width=0.9pt] (5*\sx,\yy) rectangle ++(\tw,-\th);   %
  }  %
\end{tikzpicture}
\caption{\textbf{Qualitative comparison in the wild on \genhuman.} MANO overlays, where a visible hand without a mesh is a
detection failure. Insets: the hands in 3D with 10\,s wrist paths, with the ground truth in light colors.
Top: a left hand behind a shoe, which HaWoR, \vidihand and \ace misplace. Middle: the hands open a drawer and take something
out, with the right hand in dim light. HaWoR misplaces the right hand, \ace both hands and \vidihand the left hand. Bottom: at a basket,
HaWoR puts the left hand on the right leg and \vidihand draws the right one far too large. WiLoR fails to detect one hand in every row. \ours places both hands correctly.}
\label{fig:qual}
\end{figure*}
 \begin{figure*}[t]
\centering
\definecolor{tvW}{HTML}{8D562B}\definecolor{tvM}{HTML}{2A5B90}\definecolor{tvA}{HTML}{6D5092}\definecolor{tvO}{HTML}{2CA02C}
\definecolor{tvV}{HTML}{844F8D}\definecolor{tvH}{HTML}{296191}
\begin{tikzpicture}[x=1pt, y=1pt, font=\footnotesize,
  img/.style={anchor=north west, inner sep=0pt},
  hdr/.style={anchor=base, inner sep=0pt, font=\footnotesize\bfseries},
  side/.style={rotate=90, anchor=south, inner sep=0pt, font=\footnotesize}]
\def\tw{79.3}\def\th{39.65}\def\sx{81.36}\def\sy{41.65}\def\yz{0}
\foreach \case/\ds [count=\r from 0] in {ego4d_plant/Ego4D, egodex_rake/EgoDex, xperience_phone/Xperience} {
  \pgfmathsetmacro{\yy}{\yz-\r*\sy}
  \foreach \m [count=\i from 0] in {input, wilor, hawor, vidihand, aceegohand, ours} {   %
    \node[img] at (\i*\sx,\yy) {\includegraphics[width=\tw pt,height=\th pt]{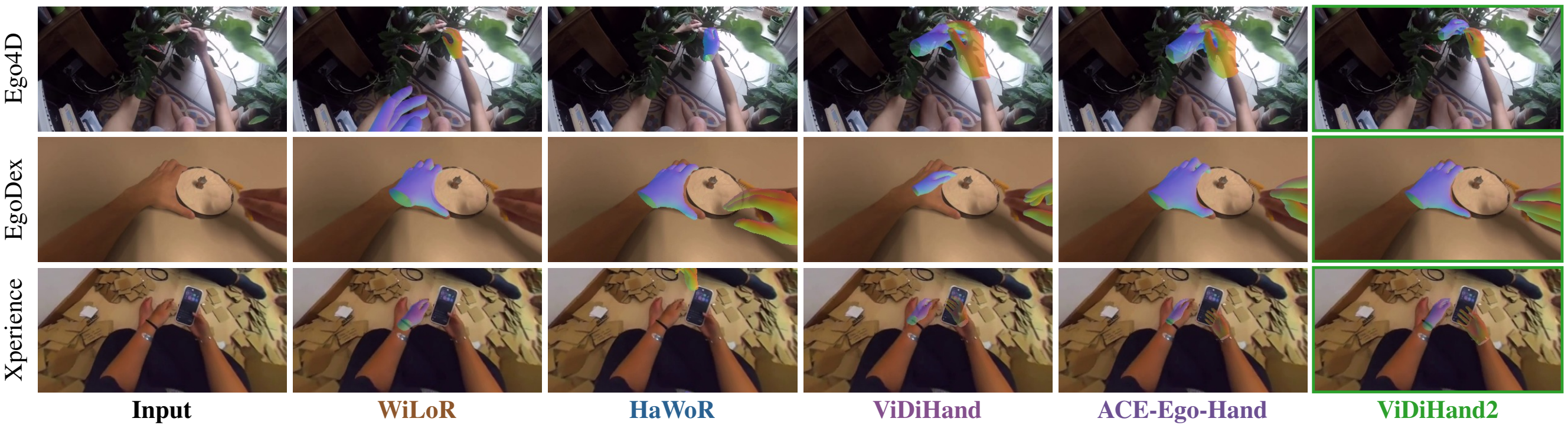}};}
  \draw[tvO, line width=0.9pt] (5*\sx,\yy) rectangle ++(\tw,-\th);   %
  \node[side] at (-3,\yy-0.5*\th) {\ds};}
\foreach \n/\c [count=\i from 0] in {Input/black, WiLoR/tvW, HaWoR/tvH, \vidihand/tvV, \ace/tvA, \textbf{\ours}/tvO} {
  \node[hdr, text=\c] at (\i*\sx+0.5*\tw,-131.2) {\n};}
\end{tikzpicture}
\caption{\textbf{Qualitative comparison in the wild on Ego4D~\cite{grauman2022ego4d}, EgoDex~\cite{hoque2026egodex} and Xperience~\cite{ropedia2026xperience}.} Without reliable ground
truth, the overlays show image alignment only. Top: hands in a plant. Middle: a tray and a rake. Bottom: a phone. WiLoR
places the left hand on the leg in the top row and fails to detect the right hand in the other two. HaWoR detects only one hand in the top row,
mis-poses the right hand in the middle row, fails to detect both hands at the phone and puts a right hand on the floor. \vidihand and \ace
draw the hands one size too large in the top row and misplace or mis-pose a hand in the other two. \ours aligns with the image
in every row.}
\label{fig:qual_nogt}
\end{figure*}
 \noindent\textbf{Comparison on Hands Detected by All Methods.} \Cref{tab:tp} removes the penalty and compares the methods only on
the hands that all 19 detect, 59\% of the hand-frames, a subset that favors the detect-then-reconstruct baselines. \ours remains
the best on every metric, and \cref{S-sec:supp_common_official} of the supplement repeats the test with the officially released
weights. In the wild, WiLoR comes close in MPJPE-abs, 30.26 against 28.09\,mm, but not in MPJPE, 21.88 against 17.82\,mm, or Jitter, 11.17
against 1.46\,mm/frame$^2$, so \ours gains on it mainly by detecting more hands. \ace and \vidihand still reach 79.15 and
90.47\,mm, so \ours gains on them by better placement.

\begin{table*}[t]
\centering
\fontsize{7.0}{6.8}\selectfont
\setlength{\tabcolsep}{0.5pt}
\renewcommand{\arraystretch}{0.92}
\newcommand{\mh}[2]{{\fontsize{6.1}{6.7}\selectfont #1#2}}
\newcommand{\mo}[1]{{\fontsize{6.1}{6.7}\selectfont #1}}
\newcommand{\blk}[5]{\ifnum#5>2 \par\vskip 5pt plus 1fill\fi\setcounter{table}{\numexpr\abbase+#5-1\relax}\centering
  \captionsetup{font=footnotesize,skip=2pt,position=top}\captionof{table}{#2}\label{#3}#1\par}
\edef\abbase{\the\value{table}}
\newsavebox\abL\newsavebox\abR
\setbox\abL=\vbox{\hsize=0.488\textwidth\linewidth=\hsize\parindent=0pt
{\fontsize{6.5}{6.7}\selectfont\blk{%
\begin{tabular*}{\linewidth}{@{\extracolsep{\fill}}lcccccccccc@{}}
\toprule
& \multicolumn{5}{c}{In-domain} & \multicolumn{5}{c}{In-the-wild} \\
\cmidrule(lr){2-6}\cmidrule(lr){7-11}
Focal factor & 0.60 & 0.80 & 1.00 & 1.25 & 1.60 & 0.60 & 0.80 & 1.00 & 1.25 & 1.60 \\
\midrule
\vidihand~\cite{wang2026vidihand} & 93.52 & 54.64 & 36.60 & 63.90 & 107.65 & 191.34 & 130.74 & 94.53 & 60.32 & 57.07 \\
\ace~\cite{liu2026aceegohand} & 178.86 & 80.42 & 24.46 & 68.97 & 151.98 & 314.15 & 217.86 & 92.13 & 42.59 & 54.79 \\
\textbf{\ours (Ours)} & \best{28.47} & \best{20.04} & \best{19.62} & \best{24.57} & \best{37.96} & \best{45.79} & \best{37.52} & \best{32.59} & \best{29.71} & \best{35.84} \\
\bottomrule
\end{tabular*}
 }{\textbf{Sensitivity to the focal length.} MPJPE-abs-p after rescaling the video and intrinsics, with padding below 1.00 and cropping above. At 1.00, results match \cref{tab:main}.}{tab:focal}{}{1}}
\blk{%
\begin{tabular*}{\linewidth}{@{\extracolsep{\fill}}lcccccc@{}}
\toprule
& \multicolumn{3}{c}{In-domain} & \multicolumn{3}{c}{In-the-wild} \\
\cmidrule(lr){2-4}\cmidrule(lr){5-7}
Backbone & \mh{MPJPE}{-abs-p$\downarrow$} & \mh{MPJPE}{-p$\downarrow$} & \mh{Jitter}{$\downarrow$} & \mh{MPJPE}{-abs-p$\downarrow$} & \mh{MPJPE}{-p$\downarrow$} & \mh{Jitter}{$\downarrow$} \\
\midrule
DINOv3~\cite{simeoni2025dinov3} & 27.74 & 13.49 & 1.54 & 36.17 & 20.78 & 1.79 \\
V-JEPA 2~\cite{assran2025vjepa2} & 31.59 & 14.15 & 1.59 & 34.91 & 19.74 & 1.81 \\
Wan2.2 Random weights & 31.62 & 19.35 & 1.53 & 41.49 & 23.93 & 1.70 \\
No task tokens in blocks & 20.50 & 11.73 & 1.30 & 34.52 & 20.37 & \best{1.60} \\
\textbf{Full (Ours)} & \best{19.62} & \best{9.50} & \best{1.26} & \best{32.59} & \best{18.59} & 1.61 \\
\bottomrule
\end{tabular*}
 }{\textbf{Ablation of the backbone.} Heads and losses are shared. The last three rows use Wan2.2. MPJPE in mm, jitter in mm/frame$^2$. Best in bold.}{tab:abl_backbone}{}{3}
\blk{%
\begin{tabular*}{\linewidth}{@{\extracolsep{\fill}}lcccccc@{}}
\toprule
& \multicolumn{3}{c}{In-domain} & \multicolumn{3}{c}{In-the-wild} \\
\cmidrule(lr){2-4}\cmidrule(lr){5-7}
2D locations & \mh{MPJPE}{-abs-p$\downarrow$} & \mh{MPJPE}{-p$\downarrow$} & \mh{Jitter}{$\downarrow$} & \mh{MPJPE}{-abs-p$\downarrow$} & \mh{MPJPE}{-p$\downarrow$} & \mh{Jitter}{$\downarrow$} \\
\midrule
Coarse token attention & 32.30 & 9.63 & \best{0.97} & 39.71 & 18.65 & \best{1.19} \\
Inside the image only & 20.20 & 9.57 & 1.26 & 32.60 & \best{18.59} & 1.60 \\
\textbf{All joints (Ours)} & \best{19.62} & \best{9.50} & 1.26 & \best{32.59} & \best{18.59} & 1.61 \\
\bottomrule
\end{tabular*}
 }{\textbf{Ablation of the Joint-Heatmap Refiner.} The coarse 2D locations, or those of \cref{eq:read}
inside the image only or for all joints.}{tab:abl_loc}{}{5}
}
\setbox\abR=\vbox{\hsize=0.488\textwidth\linewidth=\hsize\parindent=0pt
\blk{%
\begin{tabular*}{\linewidth}{@{\extracolsep{\fill}}lcccccc@{}}
\toprule
& \multicolumn{3}{c}{In-domain} & \multicolumn{3}{c}{In-the-wild} \\
\cmidrule(lr){2-4}\cmidrule(lr){5-7}
Chunks & \mo{MPJPE-abs-p$\downarrow$} & \mo{Jitter$\downarrow$} & \mo{Boundary jitter$\downarrow$} & \mo{MPJPE-abs-p$\downarrow$} & \mo{Jitter$\downarrow$} & \mo{Boundary jitter$\downarrow$} \\
\midrule
Independent & \best{19.62} & 1.27 & 0.98 & 32.62 & \best{1.61} & 1.34 \\
\textbf{AR (Ours)} & \best{19.62} & \best{1.26} & \best{0.94} & \best{32.59} & \best{1.61} & \best{1.29} \\
\bottomrule
\end{tabular*}
 }{\textbf{Ablation of Autoregressive Rollout Adaptation.} Cached AR uses prior chunks. Boundary jitter measures wrist-relative jitter across chunks.}{tab:abl_ar}{}{2}
\blk{%
\begin{tabular*}{\linewidth}{@{\extracolsep{\fill}}lcccccc@{}}
\toprule
& \multicolumn{3}{c}{In-domain} & \multicolumn{3}{c}{In-the-wild} \\
\cmidrule(lr){2-4}\cmidrule(lr){5-7}
Hand depth & \mh{MPJPE}{-abs-p$\downarrow$} & \mh{MPJPE}{-p$\downarrow$} & \mh{Jitter}{$\downarrow$} & \mh{MPJPE}{-abs-p$\downarrow$} & \mh{MPJPE}{-p$\downarrow$} & \mh{Jitter}{$\downarrow$} \\
\midrule
Regressed depth & 25.81 & 9.64 & 1.43 & 81.24 & \best{18.26} & 2.12   \\
Regressed depth/$f$ & 22.33 & 9.60 & 1.41 & 42.52 & 18.58 & 1.72   \\
\midrule
One bone (closed form) & 23.82 & 9.61 & 4.73 & 38.83 & 18.68 & 4.91   \\
Per-frame reprojection & 20.66 & 9.52 & 3.55 & 34.40 & 18.62 & 3.72   \\
Per-frame ray fit & 20.24 & \best{9.50} & 3.39 & 32.69 & 18.59 & 3.65   \\
\midrule
Clip-level reprojection & 20.32 & \best{9.50} & \best{1.24} & 32.86 & 18.60 & \best{1.58}   \\
\textbf{Clip-level ray fit (Ours)} & \best{19.62} & \best{9.50} & 1.26 & \best{32.59} & 18.59 & 1.61   \\
\bottomrule
\end{tabular*}
 }{\textbf{Ablation of Clip-Level \rayfit.} Depth regression (raw or normalized by focal length $f$) and per-frame or clip-level fitting.}{tab:abl_depth}{}{4}
\blk{%
\begin{tabular*}{\linewidth}{@{\extracolsep{\fill}}lcccccc@{}}
\toprule
& \multicolumn{3}{c}{In-domain} & \multicolumn{3}{c}{In-the-wild} \\
\cmidrule(lr){2-4}\cmidrule(lr){5-7}
Output joints & \mh{MPJPE}{-abs-p$\downarrow$} & \mh{MPJPE}{-p$\downarrow$} & \mh{Jitter}{$\downarrow$} & \mh{MPJPE}{-abs-p$\downarrow$} & \mh{MPJPE}{-p$\downarrow$} & \mh{Jitter}{$\downarrow$} \\
\midrule
Joints only & 19.81 & 10.40 & 1.60 & 32.92 & 20.95 & 2.01 \\
MANO only & 19.79 & 9.75 & \best{1.20} & 32.62 & 18.71 & \best{1.59} \\
Confidence ($g_j=c_j$) & 19.72 & 9.59 & 1.45 & 32.61 & 18.99 & 1.72 \\
\textbf{Learned gate (Ours)} & \best{19.62} & \best{9.50} & 1.26 & \best{32.59} & \best{18.59} & 1.61 \\
\bottomrule
\end{tabular*}
 }{\textbf{Ablation of Confidence-Gated Fusion.} Joints on their rays, MANO joints, or both refined
with $g_j=c_j$ or the learned gate.}{tab:abl_fusion}{}{6}
}
\dimen0=\ht\abL\advance\dimen0\dp\abL\dimen2=\ht\abR\advance\dimen2\dp\abR\ifdim\dimen2>\dimen0 \dimen0=\dimen2\fi
\hbox to \textwidth{\vtop to \dimen0{\unvbox\abL}\hfill\vtop to \dimen0{\unvbox\abR}}
\setcounter{table}{\numexpr\abbase+6\relax}
\end{table*}
 \noindent\textbf{Comparison on an Unseen Camera.} \ours transfers to an unseen camera far better than the video diffusion
baselines, which regress the depth. From in-domain to the wild, the MPJPE-abs-p of \vidihand and \ace rises by 57.93 and
67.67\,mm, that of \ours by only 12.97\,mm, within the range of the detect-then-reconstruct baselines.

On the hands
they detect in the wild, the two baselines predict the hand size to within 7\% yet place the hand 21\% and 26\% too far in the
median, whereas \ours places it 2\% too near, as \cref{S-sec:supp_lifting} of the supplement shows. A correct size with a wrong
placement points to the regressed depth, which \ours replaces by a fit.

Rescaling the videos and intrinsics by focal factors from
0.60 to 1.60 in \cref{tab:focal} gives the same picture, with \ours within 19.62 to 45.79\,mm and \ace
up~to~314.15\,mm.

\noindent\textbf{Qualitative Comparison.} \Cref{fig:qual,fig:qual_nogt} show occluded or truncated hands,
which the baselines fail to detect or misplace, whereas \ours aligns them with the image and, where ground truth exists, places
them close to it in 3D. \Cref{S-sec:supp_qual_wild} of the supplement shows more cases.

\subsection{Ablation Studies}
\label{sec:ablation}
\Cref{tab:abl_ar,tab:abl_backbone,tab:abl_depth,tab:abl_loc,tab:abl_fusion} each change one design of \ours, evaluated as in
\cref{tab:main}: Autoregressive Rollout Adaptation, the backbone, Clip-Level Ray Fitting, the Joint-Heatmap Refiner and
Confidence-Gated Fusion. In \cref{tab:abl_ar}, the rollout cache lowers the boundary jitter at equal or better accuracy.

\noindent\textbf{Importance of the Video Diffusion Prior.} In \cref{tab:abl_backbone}, the pretrained Wan2.2 with task tokens
in every block is the most accurate backbone in both settings. With the heads and losses of \ours,
DINOv3~\cite{simeoni2025dinov3}, V-JEPA~2~\cite{assran2025vjepa2} and the Wan2.2 transformer from random weights are less
accurate. Pretraining lowers the MPJPE-abs-p of the Wan2.2 transformer by 38\% in-domain and 21\% in the wild. Reading the task
tokens with a separate decoder after the last block, as \ace does, raises the MPJPE-p from 9.50 to 11.73\,mm in-domain and from
18.59 to 20.37\,mm in the wild, which shows the benefit of task tokens in every block. Even so, every variant, at 34.52 to
41.49\,mm in the wild, stays far below the 56.55\,mm of WiLoR, the best baseline.

\noindent\textbf{Fitting versus Regressing the Hand Depth.} As \cref{tab:abl_depth} shows, fitting the hand depth transfers to an
unseen camera far better than regressing it. A head trained with the features to regress the depth reaches an in-the-wild
MPJPE-abs-p of 81.24\,mm, close to \ace and \vidihand, and 42.52\,mm for the depth over the focal length, whereas every fit stays
at 38.83\,mm or below. On the features trained with the depth head, the fit still reaches 32.55\,mm, so the gap comes from how the
depth is obtained, not from the features.

\noindent\looseness=-1\textbf{Importance of Clip-Level Ray Fitting and Fine-Grained Geometric Lifting.} Fitting over the clip smooths the motion, and the bone term,
which ties the joints on their rays to the metric MANO joint-pair distances, improves the placement. One wrist trajectory per clip
lowers the jitter of the per-frame fit from 3.39 to 1.26\,mm/frame$^2$ in-domain and from 3.65 to 1.61\,mm/frame$^2$ in the wild.
Adding the bone term to the reprojection fit of MeTRAbs~\cite{sarandi2021metrabs} and HandDGP~\cite{valassakis2024handdgp} lowers
its MPJPE-abs-p in the wild from 34.40 to 32.69\,mm per frame and, over the clip, on every dataset by 0.17 to 1.73\,mm.

\noindent\textbf{Importance of the Joint-Heatmap Refiner.} In \cref{tab:abl_loc}, the 2D locations of \cref{sec:loc} have the largest
in-domain effect of all designs, as their heat maps are eight times finer than the feature map. Using only the coarse token-attention
locations, as \ace does, is smoother but less accurate and raises the MPJPE-abs-p from
19.62 to 32.30\,mm in-domain and from 32.59 to 39.71\,mm in the wild. Keeping only those inside the image gives
20.20\,mm in-domain.

\noindent\textbf{Confidence-Gated Fusion versus Either Source.} In \cref{tab:abl_fusion}, whose rows share one model and one wrist
trajectory and differ only in the output joints, the fusion is more accurate than either of its sources on every accuracy metric
in both settings. It combines the MANO hand, which is complete but need not align with the image, with the 2D locations, which
align with the image but are unreliable under occlusion and beyond the border. In-domain, it thus lowers the EPE2D-p of the MANO
hand by 7\%, from 5.03 to 4.67\,px, and its MPJPE-p from 9.75 to 9.50\,mm. The learned gate is also more accurate and smoother
than setting the gate to the confidence, $g_j=c_j$. \Cref{S-sec:supp_abl} of the supplement defines all variants and reports
more metrics.
 \section{Conclusion}
\label{sec:conclusion}
We propose \ours, which reformulates VDM-based hand depth estimation from regression into geometric constraint solving.
Fine-Grained Geometric Lifting and Clip-Level Ray Fitting compute the depth of a clip on the rays of the test camera, the
Joint-Heatmap Refiner and Confidence-Gated Fusion handle occlusion and truncation, and the Multi-Level Latent Encoder, Hand Latent
Decoder and Autoregressive Rollout Adaptation improve accuracy and smoothness. \ours substantially outperforms 18 methods on every
metric, in-domain and in the wild, at 57 FPS, enabling scalable hand motion data collection for embodied AI.

\noindent\textbf{Limitations.}
\ours needs the camera intrinsics, which egocentric devices provide in practice. It also runs offline, which suits
recorded videos but not closed-loop control.
 \paragraph{Acknowledgements.}
This research is supported by RIE2025 Industry Alignment Fund -- Industry Collaboration Projects (IAF-ICP) (Award I2301E0026),
administered by A*STAR, and by Alibaba Group and NTU Singapore through the Alibaba-NTU Global e-Sustainability CorpLab (ANGEL).

{
    \small
    \bibliographystyle{ieeenat_fullname}

}

\clearpage
\onecolumn
\raggedbottom
\renewcommand{\topfraction}{0.95}\renewcommand{\bottomfraction}{0.7}\renewcommand{\textfraction}{0.05}
\renewcommand{\floatpagefraction}{0.8}\setcounter{totalnumber}{5}\setcounter{topnumber}{4}\setcounter{bottomnumber}{3}
\setlength{\floatsep}{12pt plus 2pt minus 2pt}\setlength{\textfloatsep}{20pt plus 2pt minus 4pt}
\setlength{\aboverulesep}{0.4ex}\setlength{\belowrulesep}{0.65ex}
\setcounter{section}{0}\setcounter{figure}{0}\setcounter{table}{0}\setcounter{equation}{0}
\renewcommand{\theHsection}{supp.\arabic{section}}\renewcommand{\theHsubsection}{supp.\arabic{section}.\arabic{subsection}}
\renewcommand{\theHsubsubsection}{supp.\arabic{section}.\arabic{subsection}.\arabic{subsubsection}}
\renewcommand{\theHfigure}{supp.\arabic{figure}}\renewcommand{\theHtable}{supp.\arabic{table}}
\renewcommand{\theHequation}{supp.\arabic{equation}}
\addtocontents{toc}{\protect\setcounter{tocdepth}{2}}

{\centering\Large\textbf{\thetitle}\\[0.5em]Supplementary Material\par\vspace{0.4em}}
\section*{Overview}
This supplement gives the datasets and the evaluation protocol in \cref{S-sec:supp_protocol}, quantitative results in
\cref{S-sec:supp_tables}, more qualitative results in \cref{S-sec:supp_qual} and implementation details in \cref{S-sec:supp_impl}, which
cover every module of \cref{sec:method}: the Multi-Level Latent Encoder, the Joint-Heatmap Refiner, Fine-Grained Geometric Lifting with Clip-Level Ray Fitting, the Hand Latent Decoder, the Autoregressive Rollout Adaptation and Confidence-Gated Fusion. The variants of the
ablation studies and the runtime follow in Secs.~\ref{S-sec:supp_abl} and~\ref{S-sec:supp_speed}.
References such as ``Tab.~1'' or ``Eq.~(2)'' point to the main paper.
\vspace{0.5em}
{\hypersetup{linkcolor=black}\setcounter{tocdepth}{2}\renewcommand{\numberline}[1]{#1\hskip1em}\tableofcontents}
\vspace{1.0em}
\renewcommand{\thesection}{\Alph{section}}
\setcounter{section}{0}
\renewcommand{\thetable}{S\arabic{table}}
\renewcommand{\thefigure}{S\arabic{figure}}
\renewcommand{\theequation}{S\arabic{equation}}
\setcounter{table}{0}
\setcounter{figure}{0}

\section{Datasets and Evaluation Protocol}
\label{S-sec:supp_protocol}

\Cref{S-tab:supp_data} lists the datasets of Sec.~\ref{sec:bench}, and \cref{S-tab:supp_filter} lists the frames removed by label
filtering. We evaluate the baselines in two versions. For Tab.~\ref{tab:main}, we re-train all 18 baselines with their official
code on the training splits of \cref{S-tab:supp_data}, the same training data as \ours, so that every method is trained and tested on
the same data. For 15 of these baselines, \cref{S-tab:supp_official} also evaluates their officially released weights, which are
trained on other data. The released weights of \vidihand and \ace are themselves trained on these six datasets, and EgoHandICL has no
released weights for them. Every method receives the RGB video and the ground-truth intrinsics $\K$, and uses $\K$ wherever its
official code accepts intrinsics, for example to refit the translation or as a network input. No method receives the ground-truth
camera pose, hand shape or hand size.

\subsection{Labels}
We represent all labels by MANO in camera coordinates with per-frame pinhole intrinsics and the mirrored right-hand model for left
hands. For datasets that provide only 3D joints, we fit MANO to these joints, for \genhuman to its tracked joints with one shape per
clip after undistorting the fisheye images of one camera. Rule-based checks and visual
inspection remove 1{,}864 \genhuman frames, most of them with inaccurate labels, as \cref{S-tab:supp_filter} lists.

\subsection{Splits}
ARCTIC and H2O are split by subject and the other in-domain datasets by recording. HOT3D and OakInk2 share participants or scenes
between the splits, and the two EgoDemo datasets record no subject identity.

\begin{table}[H]
\centering
\caption{\textbf{Datasets after label filtering.} Clips and hours of the training splits, and clips, hours and frames of the test splits.}
\label{S-tab:supp_data}
\setlength{\tabcolsep}{3pt}
\adjustbox{max width=\textwidth}{%
\begin{tabular}{@{}llrrrrr@{}}
\toprule
Set & Setting & Training clips / h & Test clips & Test h & Test frames & Resolution \\
\midrule
ARCTIC~\cite{S-fan2023arctic} & In-domain & 277 / 1.78 & 37 & 0.24 & 25{,}783 & $672{\times}480$ \\
H2O~\cite{S-kwon2021h2o} & In-domain & 180 / 0.77 & 54 & 0.28 & 30{,}439 & $832{\times}480$ \\
HOT3D~\cite{S-banerjee2025hot3d} & In-domain & 562 / 3.36 & 29 & 0.24 & 25{,}740 & $480{\times}480$ \\
OakInk2~\cite{S-zhan2024oakink2} & In-domain & 680 / 6.94 & 258 & 2.20 & 237{,}591 & $832{\times}480$ \\
EgoDemo Pro (Ego-Pro)~\cite{S-lightwheel2026egosuite} & In-domain & 1{,}191 / 14.24 & 184 & 1.81 & 195{,}170 & $608{\times}480$ \\
EgoDemo Standard (Ego-Std)~\cite{S-lightwheel2026egosuite} & In-domain & 727 / 11.88 & 80 & 1.45 & 156{,}250 & $608{\times}480$ \\
\genhuman~\cite{S-genrobot2026genhumanego} & In-the-wild & -- & 46 & 1.01 & 109{,}580 & $576{\times}480$ \\
\midrule
Total & & 3{,}617 / 38.97 & 688 & 7.23 & 780{,}553 & \\
\bottomrule
\end{tabular}
 }
\end{table}

\subsection{Label Filtering}
\label{S-sec:supp_filtering}
We identify the frames to remove in two steps. Rule-based checks on the labels and the video
flag label jumps between consecutive frames, abnormal hand poses, labeled hands without a visible hand, visible frames without
labels and dark or blurred frames. We then render the labels over the video and inspect the flagged events and their neighboring
frames. The 196{,}120 removed frames of \cref{S-tab:supp_filter} fall into five groups. In 104{,}493 frames, the label is inaccurate,
for example misplaced, misshaped, jumping between frames or in an abnormal pose. In 1{,}413 frames, a hand is labeled where no hand
is visible, and in 6{,}570 frames, a visible hand has no label. In 25{,}487 frames, the dataset provides no label. In 58{,}157
frames, the image is unusable: clips with privacy mosaics and dark, blurred or missing frames. We remove these frames from training and
from the evaluation of every method and record the reason for each removal. Although the public datasets have largely accurate
labels, the percentage of frames with inaccurate labels ranges from 0.07\% on ARCTIC to 5.83\% on HOT3D.
\Cref{S-tab:supp_qc} gives the frames before and after filtering and the percentage of inaccurate labels of each dataset.

\begin{table}[!tbp]
\centering
\caption{\textbf{Label filtering.} Frames removed from each dataset, with the recorded reasons in the five groups of
\cref{S-sec:supp_filtering}.}
\label{S-tab:supp_filter}
\setlength{\tabcolsep}{3pt}
\adjustbox{max width=\textwidth}{%
\begin{tabular}{@{}llrrrrrr@{}}
\toprule
Set & Camera & Inaccurate label & Label without a hand & Hand without a label & No label & Unusable image & Removed \\
\midrule
ARCTIC & Ego camera 0 & 175 & -- & -- & 25{,}477 & 497 & 26{,}149 \\
H2O & Ego camera 4 & 1{,}136 & -- & -- & -- & 12 & 1{,}148 \\
HOT3D & RGB camera & 27{,}988 & 72 & 6{,}550 & -- & 57{,}648 & 92{,}258 \\
OakInk2 & Ego camera 0 & 16{,}971 & 1{,}084 & -- & -- & -- & 18{,}055 \\
Ego-Pro & Right-eye camera & 18{,}112 & 5 & -- & -- & -- & 18{,}117 \\
Ego-Std & Right-eye camera & 38{,}478 & 51 & -- & -- & -- & 38{,}529 \\
\genhuman & Right stereo camera 3 & 1{,}633 & 201 & 20 & 10 & -- & 1{,}864 \\
\midrule
Total & & 104{,}493 & 1{,}413 & 6{,}570 & 25{,}487 & 58{,}157 & 196{,}120 \\
\bottomrule
\end{tabular}
 }
\end{table}

\begin{table}[!tbp]
\centering
\caption{\textbf{Datasets before and after label filtering.} Clips, frames and hours of the camera stream used from each dataset.
Cutting a source clip at its removed frames yields several clips, so the total number of clips grows. The inaccurate-label columns give
the removed frames of the group ``Inaccurate label'' of \cref{S-tab:supp_filter} and their share of the frames before filtering.}
\label{S-tab:supp_qc}
\setlength{\tabcolsep}{4pt}
\adjustbox{max width=\textwidth}{%
\begin{tabular}{@{}lrrrrrrrrr@{}}
\toprule
& \multicolumn{2}{c}{Clips} & \multicolumn{3}{c}{Frames} & \multicolumn{2}{c}{Inaccurate labels} & \multicolumn{2}{c}{Hours} \\
\cmidrule(lr){2-3}\cmidrule(lr){4-6}\cmidrule(lr){7-8}\cmidrule(lr){9-10}
Set & Source & After & Before & After & Removed & Frames & \% & Before & After \\
\midrule
ARCTIC & 339 & 314 & 243{,}750 & 217{,}601 & 26{,}149 & 175 & 0.07 & 2.26 & 2.02 \\
H2O & 184 & 234 & 114{,}329 & 113{,}181 & 1{,}148 & 1{,}136 & 0.99 & 1.06 & 1.05 \\
HOT3D & 136 & 591 & 480{,}375 & 388{,}117 & 92{,}258 & 27{,}988 & 5.83 & 4.46 & 3.60 \\
OakInk2 & 627 & 938 & 1{,}004{,}631 & 986{,}576 & 18{,}055 & 16{,}971 & 1.69 & 9.31 & 9.14 \\
Ego-Pro & 668 & 1{,}375 & 1{,}747{,}030 & 1{,}728{,}913 & 18{,}117 & 18{,}112 & 1.04 & 16.22 & 16.05 \\
Ego-Std & 408 & 807 & 1{,}468{,}564 & 1{,}430{,}035 & 38{,}529 & 38{,}478 & 2.62 & 13.69 & 13.33 \\
\genhuman & 28 & 46 & 111{,}444 & 109{,}580 & 1{,}864 & 1{,}633 & 1.47 & 1.03 & 1.01 \\
\midrule
Total & 2{,}390 & 4{,}305 & 5{,}170{,}123 & 4{,}974{,}003 & 196{,}120 & 104{,}493 & 2.02 & 48.03 & 46.20 \\
\bottomrule
\end{tabular}
 }
\end{table}

\subsection{Joints and Matching}
\looseness=-1 In each frame, a method predicts at most one left and one right hand, each as 21 joints in camera coordinates. These are the 16 MANO
kinematic joints and five fingertip vertices. We compute the same joints from MANO predictions. A ground-truth hand is \emph{visible}
if at least one of its joints has a depth $z>1$\,cm and projects inside the image, and a predicted hand likewise. Only visible hands
are evaluated. Each predicted hand is matched to the ground-truth hand with the largest intersection over union (IoU) of the 2D
boxes around all 21 projected joints, where the ground-truth box is enlarged by 10\% and a hand of the same handedness receives a
bonus of 0.5. A \emph{detected hand} is a visible ground-truth hand that is matched to a prediction of the same handedness with nonzero
IoU. Every other visible ground-truth hand is a \emph{detection failure}. A visible prediction is a \emph{false positive} if its match has
zero IoU or the opposite handedness, so a prediction with swapped handedness counts as a false positive and leaves a detection
failure. A prediction that overlaps an annotated hand of the same handedness that is not visible is ignored.

\subsection{Metrics}
The frame accuracy FAcc is the fraction of test frames with neither a false positive nor a detection failure. F1 is the harmonic
mean of the precision and the recall of the detections, with the counts pooled over all hands of a dataset. False positives enter
FAcc and F1 only.
MPJPE-abs is the joint error in camera coordinates, MPJPE the joint error after subtracting the wrist of each hand, and PA-MPJPE the
joint error after aligning scale, rotation and translation. Each is the mean over the 21 joints of a hand, also of joints outside
the image, and then over the hands. EPE2D is the pixel distance between the predicted joints, projected with the ground-truth
intrinsics, and the ground-truth joints, over the visible ground-truth joints, with each distance capped at the image diagonal.
The suffix -p denotes that an error is computed over all visible ground-truth hands. Each detected hand is evaluated with the matched
prediction, and each detection failure receives a penalty. For MPJPE-abs-p, MPJPE-p and EPE2D-p, the penalty is the error of a default
prediction. This default prediction places the MANO mean hand of the same handedness with the wrist at a depth of 0.5\,m on the camera
ray through the image center. Since PA-MPJPE
aligns each prediction to the ground truth, the default prediction can obtain a lower error than an actual one. The penalty of PA-p is therefore
the root-mean-square distance of the ground-truth joints from their centroid, an upper bound on the Procrustes-aligned error of any
non-degenerate prediction.
Jitter is the mean acceleration $\|\vect{X}_{t+1,j}-2\vect{X}_{t,j}+\vect{X}_{t-1,j}\|$ of the predicted joints $\vect{X}_{t,j}$ of a method in camera coordinates, in mm/frame$^2$,
over every three consecutive frames in which the hand is detected. It has no penalty and no ground-truth term, so it measures
smoothness and not accuracy. In the
comparisons on hands detected by all methods, a hand-frame is kept when every method detects it, the errors have no penalty, and
Jitter uses the triplets of consecutive kept frames. Every metric is computed over all test clips of a dataset, and the in-domain
value is the unweighted mean over the six in-domain datasets.

\section{Quantitative Results}
\label{S-sec:supp_tables}

This section gives the comparison of Tab.~\ref{tab:main} on every in-domain dataset with the re-trained baselines in
\cref{S-sec:supp_baselines} and with their officially released weights in \cref{S-sec:supp_official}. \Cref{S-sec:supp_common,S-sec:supp_common_official}
give the comparison on hands detected by all methods for both versions. \ours is the best on every metric of
every dataset in both versions, and it ties only where several methods reach 100\% FAcc or F1, on H2O, Ego-Pro and Ego-Std. On the
hands detected by all methods, the one exception is the in-the-wild PA of ViDiHand$^\ast$ with its released weights in
\cref{S-tab:supp_common_tp_official}. All tables list the methods in the order of Tab.~\ref{tab:main} and use its units, FAcc
and F1 in \%, MPJPE-abs-p, MPJPE-p and PA-p in mm, EPE2D-p in px and Jitter in mm/frame$^2$, with the best value of each column in
bold.

\subsection{Baselines Trained on the Same Six Datasets}
\label{S-sec:supp_baselines}
\Cref{S-tab:supp_ehb_arctic,S-tab:supp_ehb_h2o,S-tab:supp_ehb_hot3d,S-tab:supp_ehb_oakink2,S-tab:supp_ehb_egopro,S-tab:supp_ehb_egostd} give
every metric of Tab.~\ref{tab:main} on each in-domain dataset, with all 18 baselines re-trained with their official code and
losses on the same training data as \ours. \ours is the best on every metric of every dataset.

\begin{table}[!tbp]
\centering
\setlength{\tabcolsep}{3.2pt}
\captionof{table}{\textbf{Comparison on ARCTIC with the re-trained baselines.} All 18 baselines are re-trained with their official code on the six training splits of \ours, and every method is tested on the ARCTIC test split. Metrics with the suffix -p add a penalty for each detection failure. FAcc and F1 in \%, EPE2D-p in px, Jitter in mm/frame$^2$, other errors in mm, best values in bold. \ours is the \textbf{best} on every metric.}
\label{S-tab:supp_ehb_arctic}
\adjustbox{max width=0.8\textwidth,max totalheight=0.37\textheight}{%
\begin{tabular}{@{}lccccccc@{}}
\toprule
& \multicolumn{2}{c}{Detection} & \multicolumn{2}{c}{Absolute pose} & \multicolumn{2}{c}{Relative pose} & \multicolumn{1}{c}{Temporal} \\
\cmidrule(lr){2-3}\cmidrule(lr){4-5}\cmidrule(lr){6-7}\cmidrule(lr){8-8}
Method & FAcc$\uparrow$ & F1$\uparrow$ & MPJPE-abs-p$\downarrow$ & EPE2D-p$\downarrow$ & MPJPE-p$\downarrow$ & PA-p$\downarrow$ & Jitter$\downarrow$ \\
\midrule
\multicolumn{8}{@{}l}{\textit{Detect-then-reconstruct, single frame}} \\
Hamba~\cite{S-dong2024hamba} & 76.23 & 89.66 & 69.84 & 37.12 & 27.18 & 14.09 & 18.40 \\
WildHands~\cite{S-prakash2024wildhands} & 84.04 & 93.70 & 71.40 & 37.98 & 40.20 & 15.47 & 20.59 \\
InterWild~\cite{S-moon2023interwild} & 80.18 & 90.91 & 76.06 & 28.39 & 33.03 & 13.54 & 32.39 \\
AnyHand~\cite{S-si2026anyhand} & 89.75 & 96.56 & 56.57 & 21.60 & 21.01 & 11.59 & 16.19 \\
HaMeR~\cite{S-pavlakos2024hamer} & 85.53 & 94.35 & 56.66 & 26.54 & 23.77 & 12.29 & 19.20 \\
EgoForce~\cite{S-millerdurai2026egoforce} & 85.62 & 95.21 & 59.12 & 27.94 & 25.51 & 12.51 & 18.71 \\
HandDGP~\cite{S-valassakis2024handdgp} & 89.75 & 96.56 & 82.35 & 28.73 & 37.82 & 17.32 & 21.68 \\
EgoHandICL~\cite{S-xie2026egohandicl} & 89.75 & 96.56 & 61.09 & 20.50 & 20.69 & 10.93 & 17.91 \\
WiLoR~\cite{S-potamias2025wilor} & 89.75 & 96.56 & 62.19 & 20.30 & 20.56 & 10.88 & 17.42 \\
\midrule
\multicolumn{8}{@{}l}{\textit{Detect-then-reconstruct with a motion prior}} \\
Dyn-HaMR~\cite{S-yu2025dynhamr} & 94.97 & 98.04 & 191.17 & 19.08 & 32.84 & 9.75 & 18.70 \\
OmniHands~\cite{S-lin2026omnihands} & 77.60 & 89.35 & 93.37 & 28.39 & 24.27 & 11.44 & 43.85 \\
HaWoR~\cite{S-zhang2025hawor} & 85.30 & 92.94 & 60.00 & 26.18 & 26.44 & 11.60 & 28.26 \\
HandFlow~\cite{S-xu2026handflow} & 89.39 & 96.50 & 51.65 & 28.68 & 28.68 & 12.68 & 20.52 \\
MINT~\cite{S-zhu2026mint} & 97.67 & 99.07 & 61.26 & 29.20 & 26.27 & 12.09 & 12.87 \\
PAD-Hand~\cite{S-ismayilzada2026padhand} & 86.25 & 95.43 & 83.86 & 31.52 & 27.07 & 13.58 & 19.65 \\
InfiniHand~\cite{S-ren2026infinihand} & 97.96 & 99.13 & 45.51 & 11.91 & 19.58 & 7.52 & 27.97 \\
\midrule
\multicolumn{8}{@{}l}{\textit{Video diffusion with depth regression}} \\
ViDiHand~\cite{S-wang2026vidihand} & 98.89 & 99.67 & 46.55 & 12.99 & 21.66 & 10.02 & 3.81 \\
ACE-Ego-Hand~\cite{S-liu2026aceegohand} & 98.81 & 99.65 & 37.95 & 11.00 & 15.90 & 7.25 & 2.52 \\
\midrule
\multicolumn{8}{@{}l}{\textit{Video diffusion with hand geometry}} \\
\textbf{\ours (Ours)} & \best{99.90} & \best{99.97} & \best{33.33} & \best{7.72} & \best{14.19} & \best{6.89} & \best{2.30} \\
\bottomrule
\end{tabular}
 }
\end{table}
\begin{table}[!tbp]
\centering
\setlength{\tabcolsep}{3.2pt}
\captionof{table}{\textbf{Comparison on H2O with the re-trained baselines.} All 18 baselines are re-trained with their official code on the six training splits of \ours, and every method is tested on the H2O test split. Metrics with the suffix -p add a penalty for each detection failure. FAcc and F1 in \%, EPE2D-p in px, Jitter in mm/frame$^2$, other errors in mm, best values in bold. \ours is the \textbf{best} on every metric. Other methods tie with it only where they also reach 100\% FAcc or F1.}
\label{S-tab:supp_ehb_h2o}
\adjustbox{max width=0.8\textwidth,max totalheight=0.37\textheight}{%
\begin{tabular}{@{}lccccccc@{}}
\toprule
& \multicolumn{2}{c}{Detection} & \multicolumn{2}{c}{Absolute pose} & \multicolumn{2}{c}{Relative pose} & \multicolumn{1}{c}{Temporal} \\
\cmidrule(lr){2-3}\cmidrule(lr){4-5}\cmidrule(lr){6-7}\cmidrule(lr){8-8}
Method & FAcc$\uparrow$ & F1$\uparrow$ & MPJPE-abs-p$\downarrow$ & EPE2D-p$\downarrow$ & MPJPE-p$\downarrow$ & PA-p$\downarrow$ & Jitter$\downarrow$ \\
\midrule
\multicolumn{8}{@{}l}{\textit{Detect-then-reconstruct, single frame}} \\
Hamba~\cite{S-dong2024hamba} & 88.72 & 95.98 & 31.92 & 19.51 & 17.48 & 8.25 & 6.49 \\
WildHands~\cite{S-prakash2024wildhands} & 99.22 & 99.68 & 39.50 & 16.75 & 29.16 & 7.69 & 10.25 \\
InterWild~\cite{S-moon2023interwild} & 94.73 & 97.33 & 59.38 & 25.44 & 31.55 & 9.37 & 15.62 \\
AnyHand~\cite{S-si2026anyhand} & 99.67 & 99.90 & 22.48 & 7.31 & 11.88 & 5.37 & 5.86 \\
HaMeR~\cite{S-pavlakos2024hamer} & 98.54 & 99.38 & 21.21 & 9.97 & 13.65 & 5.89 & 7.32 \\
EgoForce~\cite{S-millerdurai2026egoforce} & 99.46 & 99.86 & 25.96 & 8.12 & 12.12 & 5.12 & 6.84 \\
HandDGP~\cite{S-valassakis2024handdgp} & 99.67 & 99.90 & 36.69 & 13.00 & 28.12 & 11.02 & 6.68 \\
EgoHandICL~\cite{S-xie2026egohandicl} & 99.67 & 99.90 & 28.09 & 7.22 & 10.76 & 4.83 & 7.26 \\
WiLoR~\cite{S-potamias2025wilor} & 99.67 & 99.90 & 28.35 & 7.15 & 10.82 & 4.80 & 7.10 \\
\midrule
\multicolumn{8}{@{}l}{\textit{Detect-then-reconstruct with a motion prior}} \\
Dyn-HaMR~\cite{S-yu2025dynhamr} & 99.95 & 99.97 & 23.51 & 8.80 & 17.50 & 6.29 & 5.34 \\
OmniHands~\cite{S-lin2026omnihands} & 98.91 & 99.46 & 52.23 & 16.43 & 15.74 & 6.18 & 12.57 \\
HaWoR~\cite{S-zhang2025hawor} & 99.92 & 99.96 & 19.10 & 7.42 & 12.50 & 5.91 & 12.29 \\
HandFlow~\cite{S-xu2026handflow} & 99.71 & 99.93 & 25.81 & 10.08 & 14.03 & 6.34 & 9.30 \\
MINT~\cite{S-zhu2026mint} & \best{100.00} & \best{100.00} & 66.75 & 17.96 & 15.64 & 7.24 & 7.96 \\
PAD-Hand~\cite{S-ismayilzada2026padhand} & 99.74 & 99.91 & 27.79 & 11.67 & 14.61 & 6.40 & 7.94 \\
InfiniHand~\cite{S-ren2026infinihand} & 99.89 & 99.94 & 21.79 & 7.33 & 13.65 & 5.45 & 6.94 \\
\midrule
\multicolumn{8}{@{}l}{\textit{Video diffusion with depth regression}} \\
ViDiHand~\cite{S-wang2026vidihand} & \best{100.00} & \best{100.00} & 27.27 & 7.96 & 14.95 & 6.01 & 1.74 \\
ACE-Ego-Hand~\cite{S-liu2026aceegohand} & \best{100.00} & \best{100.00} & 26.75 & 7.29 & 8.18 & 4.37 & 0.73 \\
\midrule
\multicolumn{8}{@{}l}{\textit{Video diffusion with hand geometry}} \\
\textbf{\ours (Ours)} & \best{100.00} & \best{100.00} & \best{11.99} & \best{4.10} & \best{7.84} & \best{4.10} & \best{0.59} \\
\bottomrule
\end{tabular}
 }
\end{table}
\begin{table}[!tbp]
\centering
\setlength{\tabcolsep}{3.2pt}
\captionof{table}{\textbf{Comparison on HOT3D with the re-trained baselines.} All 18 baselines are re-trained with their official code on the six training splits of \ours, and every method is tested on the HOT3D test split. Metrics with the suffix -p add a penalty for each detection failure. FAcc and F1 in \%, EPE2D-p in px, Jitter in mm/frame$^2$, other errors in mm, best values in bold. \ours is the \textbf{best} on every metric.}
\label{S-tab:supp_ehb_hot3d}
\adjustbox{max width=0.8\textwidth,max totalheight=0.37\textheight}{%
\begin{tabular}{@{}lccccccc@{}}
\toprule
& \multicolumn{2}{c}{Detection} & \multicolumn{2}{c}{Absolute pose} & \multicolumn{2}{c}{Relative pose} & \multicolumn{1}{c}{Temporal} \\
\cmidrule(lr){2-3}\cmidrule(lr){4-5}\cmidrule(lr){6-7}\cmidrule(lr){8-8}
Method & FAcc$\uparrow$ & F1$\uparrow$ & MPJPE-abs-p$\downarrow$ & EPE2D-p$\downarrow$ & MPJPE-p$\downarrow$ & PA-p$\downarrow$ & Jitter$\downarrow$ \\
\midrule
\multicolumn{8}{@{}l}{\textit{Detect-then-reconstruct, single frame}} \\
Hamba~\cite{S-dong2024hamba} & 40.92 & 74.07 & 203.20 & 53.03 & 47.25 & 19.62 & 18.54 \\
WildHands~\cite{S-prakash2024wildhands} & 64.94 & 85.91 & 159.60 & 36.30 & 47.97 & 16.43 & 22.51 \\
InterWild~\cite{S-moon2023interwild} & 64.06 & 87.54 & 99.37 & 19.91 & 39.34 & 13.24 & 42.40 \\
AnyHand~\cite{S-si2026anyhand} & 81.41 & 93.03 & 176.06 & 27.68 & 29.53 & 13.20 & 16.40 \\
HaMeR~\cite{S-pavlakos2024hamer} & 67.70 & 87.46 & 166.56 & 27.27 & 30.01 & 13.29 & 21.92 \\
EgoForce~\cite{S-millerdurai2026egoforce} & 73.26 & 89.49 & 139.09 & 48.11 & 41.14 & 15.00 & 25.01 \\
HandDGP~\cite{S-valassakis2024handdgp} & 81.41 & 93.02 & 121.09 & 27.05 & 52.60 & 18.18 & 21.31 \\
EgoHandICL~\cite{S-xie2026egohandicl} & 81.41 & 93.03 & 94.54 & 21.75 & 27.87 & 11.94 & 21.91 \\
WiLoR~\cite{S-potamias2025wilor} & 81.41 & 93.03 & 94.93 & 21.46 & 27.23 & 11.74 & 19.33 \\
\midrule
\multicolumn{8}{@{}l}{\textit{Detect-then-reconstruct with a motion prior}} \\
Dyn-HaMR~\cite{S-yu2025dynhamr} & 71.61 & 90.33 & 136.80 & 20.12 & 64.29 & 12.16 & 12.46 \\
OmniHands~\cite{S-lin2026omnihands} & 55.46 & 82.81 & 167.21 & 30.75 & 42.39 & 13.62 & 79.06 \\
HaWoR~\cite{S-zhang2025hawor} & 52.48 & 74.06 & 152.35 & 51.92 & 52.66 & 18.34 & 33.94 \\
HandFlow~\cite{S-xu2026handflow} & 82.65 & 93.83 & 86.55 & 25.16 & 33.84 & 12.68 & 22.49 \\
MINT~\cite{S-zhu2026mint} & 97.47 & 99.04 & 58.72 & 16.01 & 25.81 & 10.46 & 13.33 \\
PAD-Hand~\cite{S-ismayilzada2026padhand} & 80.73 & 92.64 & 129.49 & 40.90 & 33.24 & 12.99 & 21.87 \\
InfiniHand~\cite{S-ren2026infinihand} & 98.26 & 99.38 & 70.05 & 7.01 & 17.06 & 7.45 & 12.53 \\
\midrule
\multicolumn{8}{@{}l}{\textit{Video diffusion with depth regression}} \\
ViDiHand~\cite{S-wang2026vidihand} & 98.76 & 99.61 & 35.32 & 6.92 & 19.05 & 8.38 & 3.96 \\
ACE-Ego-Hand~\cite{S-liu2026aceegohand} & 98.18 & 99.44 & 24.26 & 5.33 & 12.78 & 6.42 & 2.31 \\
\midrule
\multicolumn{8}{@{}l}{\textit{Video diffusion with hand geometry}} \\
\textbf{\ours (Ours)} & \best{99.93} & \best{99.98} & \best{23.02} & \best{3.90} & \best{11.25} & \best{5.56} & \best{2.04} \\
\bottomrule
\end{tabular}
 }
\end{table}
\begin{table}[!tbp]
\centering
\setlength{\tabcolsep}{3.2pt}
\captionof{table}{\textbf{Comparison on OakInk2 with the re-trained baselines.} All 18 baselines are re-trained with their official code on the six training splits of \ours, and every method is tested on the OakInk2 test split. Metrics with the suffix -p add a penalty for each detection failure. FAcc and F1 in \%, EPE2D-p in px, Jitter in mm/frame$^2$, other errors in mm, best values in bold. \ours is the \textbf{best} on every metric.}
\label{S-tab:supp_ehb_oakink2}
\adjustbox{max width=0.8\textwidth,max totalheight=0.37\textheight}{%
\begin{tabular}{@{}lccccccc@{}}
\toprule
& \multicolumn{2}{c}{Detection} & \multicolumn{2}{c}{Absolute pose} & \multicolumn{2}{c}{Relative pose} & \multicolumn{1}{c}{Temporal} \\
\cmidrule(lr){2-3}\cmidrule(lr){4-5}\cmidrule(lr){6-7}\cmidrule(lr){8-8}
Method & FAcc$\uparrow$ & F1$\uparrow$ & MPJPE-abs-p$\downarrow$ & EPE2D-p$\downarrow$ & MPJPE-p$\downarrow$ & PA-p$\downarrow$ & Jitter$\downarrow$ \\
\midrule
\multicolumn{8}{@{}l}{\textit{Detect-then-reconstruct, single frame}} \\
Hamba~\cite{S-dong2024hamba} & 40.62 & 66.09 & 128.23 & 106.74 & 48.93 & 21.43 & 8.26 \\
WildHands~\cite{S-prakash2024wildhands} & 66.20 & 83.62 & 107.43 & 79.21 & 48.73 & 17.07 & 12.49 \\
InterWild~\cite{S-moon2023interwild} & 69.30 & 87.04 & 81.75 & 27.74 & 25.65 & 8.81 & 14.86 \\
AnyHand~\cite{S-si2026anyhand} & 89.96 & 96.17 & 45.39 & 18.43 & 18.45 & 8.28 & 6.60 \\
HaMeR~\cite{S-pavlakos2024hamer} & 68.08 & 84.96 & 92.06 & 64.76 & 34.44 & 14.69 & 11.20 \\
EgoForce~\cite{S-millerdurai2026egoforce} & 84.73 & 93.96 & 78.70 & 41.57 & 26.27 & 10.10 & 12.25 \\
HandDGP~\cite{S-valassakis2024handdgp} & 89.92 & 96.15 & 101.94 & 26.94 & 35.63 & 13.78 & 8.84 \\
EgoHandICL~\cite{S-xie2026egohandicl} & 89.97 & 96.19 & 45.01 & 18.21 & 17.65 & 8.06 & 10.36 \\
WiLoR~\cite{S-potamias2025wilor} & 89.96 & 96.17 & 44.24 & 17.05 & 16.87 & 7.82 & 7.12 \\
\midrule
\multicolumn{8}{@{}l}{\textit{Detect-then-reconstruct with a motion prior}} \\
Dyn-HaMR~\cite{S-yu2025dynhamr} & 90.85 & 96.63 & 70.48 & 19.22 & 23.32 & 7.11 & 7.47 \\
OmniHands~\cite{S-lin2026omnihands} & 48.15 & 74.59 & 157.55 & 67.74 & 36.62 & 13.77 & 118.42 \\
HaWoR~\cite{S-zhang2025hawor} & 79.06 & 90.27 & 48.94 & 22.22 & 20.21 & 8.28 & 18.91 \\
HandFlow~\cite{S-xu2026handflow} & 90.40 & 96.57 & 45.95 & 21.94 & 20.35 & 8.53 & 9.43 \\
MINT~\cite{S-zhu2026mint} & 97.69 & 99.16 & 66.79 & 48.56 & 18.75 & 7.94 & 8.16 \\
PAD-Hand~\cite{S-ismayilzada2026padhand} & 88.85 & 95.77 & 78.31 & 30.38 & 22.58 & 9.12 & 8.33 \\
InfiniHand~\cite{S-ren2026infinihand} & 96.81 & 98.81 & 56.97 & 14.38 & 17.52 & 5.73 & 11.41 \\
\midrule
\multicolumn{8}{@{}l}{\textit{Video diffusion with depth regression}} \\
ViDiHand~\cite{S-wang2026vidihand} & 98.33 & 99.32 & 29.31 & 13.13 & 15.16 & 6.39 & 2.23 \\
ACE-Ego-Hand~\cite{S-liu2026aceegohand} & 96.30 & 98.75 & 21.24 & 9.84 & 10.55 & 5.52 & 0.82 \\
\midrule
\multicolumn{8}{@{}l}{\textit{Video diffusion with hand geometry}} \\
\textbf{\ours (Ours)} & \best{99.75} & \best{99.91} & \best{14.54} & \best{6.66} & \best{7.56} & \best{3.89} & \best{0.59} \\
\bottomrule
\end{tabular}
 }
\end{table}
\begin{table}[!tbp]
\centering
\setlength{\tabcolsep}{3.2pt}
\captionof{table}{\textbf{Comparison on Ego-Pro with the re-trained baselines.} All 18 baselines are re-trained with their official code on the six training splits of \ours, and every method is tested on the Ego-Pro test split. Metrics with the suffix -p add a penalty for each detection failure. FAcc and F1 in \%, EPE2D-p in px, Jitter in mm/frame$^2$, other errors in mm, best values in bold. \ours is the \textbf{best} on every metric. Other methods tie with it only where they also reach 100\% FAcc or F1.}
\label{S-tab:supp_ehb_egopro}
\adjustbox{max width=0.8\textwidth,max totalheight=0.37\textheight}{%
\begin{tabular}{@{}lccccccc@{}}
\toprule
& \multicolumn{2}{c}{Detection} & \multicolumn{2}{c}{Absolute pose} & \multicolumn{2}{c}{Relative pose} & \multicolumn{1}{c}{Temporal} \\
\cmidrule(lr){2-3}\cmidrule(lr){4-5}\cmidrule(lr){6-7}\cmidrule(lr){8-8}
Method & FAcc$\uparrow$ & F1$\uparrow$ & MPJPE-abs-p$\downarrow$ & EPE2D-p$\downarrow$ & MPJPE-p$\downarrow$ & PA-p$\downarrow$ & Jitter$\downarrow$ \\
\midrule
\multicolumn{8}{@{}l}{\textit{Detect-then-reconstruct, single frame}} \\
Hamba~\cite{S-dong2024hamba} & 56.97 & 82.40 & 88.76 & 35.07 & 39.64 & 16.03 & 10.06 \\
WildHands~\cite{S-prakash2024wildhands} & 94.99 & 97.51 & 46.34 & 21.11 & 27.07 & 9.60 & 19.78 \\
InterWild~\cite{S-moon2023interwild} & 97.48 & 98.70 & 44.77 & 11.18 & 20.60 & 7.62 & 14.00 \\
AnyHand~\cite{S-si2026anyhand} & 94.15 & 98.05 & 42.05 & 8.35 & 13.96 & 6.80 & 7.96 \\
HaMeR~\cite{S-pavlakos2024hamer} & 96.51 & 98.19 & 42.50 & 10.75 & 13.68 & 6.41 & 11.84 \\
EgoForce~\cite{S-millerdurai2026egoforce} & 68.09 & 88.23 & 66.88 & 28.54 & 33.52 & 12.80 & 10.70 \\
HandDGP~\cite{S-valassakis2024handdgp} & 94.15 & 98.06 & 42.92 & 11.80 & 27.83 & 11.56 & 9.82 \\
EgoHandICL~\cite{S-xie2026egohandicl} & 94.15 & 98.06 & 27.00 & 7.18 & 13.33 & 6.18 & 8.36 \\
WiLoR~\cite{S-potamias2025wilor} & 94.15 & 98.05 & 27.28 & 7.17 & 13.29 & 6.24 & 8.14 \\
\midrule
\multicolumn{8}{@{}l}{\textit{Detect-then-reconstruct with a motion prior}} \\
Dyn-HaMR~\cite{S-yu2025dynhamr} & 98.22 & 99.26 & 60.02 & 9.88 & 27.36 & 6.25 & 8.64 \\
OmniHands~\cite{S-lin2026omnihands} & 97.35 & 98.45 & 46.77 & 12.81 & 20.20 & 7.53 & 24.04 \\
HaWoR~\cite{S-zhang2025hawor} & 74.38 & 83.71 & 75.07 & 29.27 & 34.18 & 13.54 & 19.60 \\
HandFlow~\cite{S-xu2026handflow} & 94.03 & 98.41 & 24.83 & 9.38 & 16.34 & 7.14 & 10.83 \\
MINT~\cite{S-zhu2026mint} & 99.96 & 99.98 & 34.74 & 11.23 & 15.79 & 8.38 & 7.75 \\
PAD-Hand~\cite{S-ismayilzada2026padhand} & 94.07 & 98.05 & 59.53 & 15.95 & 16.10 & 7.18 & 7.99 \\
InfiniHand~\cite{S-ren2026infinihand} & 99.88 & 99.94 & 20.25 & 4.68 & 12.72 & 5.40 & 9.48 \\
\midrule
\multicolumn{8}{@{}l}{\textit{Video diffusion with depth regression}} \\
ViDiHand~\cite{S-wang2026vidihand} & 98.36 & 99.18 & 40.39 & 9.71 & 21.31 & 9.52 & 3.06 \\
ACE-Ego-Hand~\cite{S-liu2026aceegohand} & \best{100.00} & \best{100.00} & 15.32 & 3.60 & 8.14 & 5.03 & 1.16 \\
\midrule
\multicolumn{8}{@{}l}{\textit{Video diffusion with hand geometry}} \\
\textbf{\ours (Ours)} & \best{100.00} & \best{100.00} & \best{13.90} & \best{2.71} & \best{7.29} & \best{4.51} & \best{1.06} \\
\bottomrule
\end{tabular}
 }
\end{table}
\begin{table}[!tbp]
\centering
\setlength{\tabcolsep}{3.2pt}
\captionof{table}{\textbf{Comparison on Ego-Std with the re-trained baselines.} All 18 baselines are re-trained with their official code on the six training splits of \ours, and every method is tested on the Ego-Std test split. Metrics with the suffix -p add a penalty for each detection failure. FAcc and F1 in \%, EPE2D-p in px, Jitter in mm/frame$^2$, other errors in mm, best values in bold. \ours is the \textbf{best} on every metric. Other methods tie with it only where they also reach 100\% FAcc or F1.}
\label{S-tab:supp_ehb_egostd}
\adjustbox{max width=0.8\textwidth,max totalheight=0.37\textheight}{%
\begin{tabular}{@{}lccccccc@{}}
\toprule
& \multicolumn{2}{c}{Detection} & \multicolumn{2}{c}{Absolute pose} & \multicolumn{2}{c}{Relative pose} & \multicolumn{1}{c}{Temporal} \\
\cmidrule(lr){2-3}\cmidrule(lr){4-5}\cmidrule(lr){6-7}\cmidrule(lr){8-8}
Method & FAcc$\uparrow$ & F1$\uparrow$ & MPJPE-abs-p$\downarrow$ & EPE2D-p$\downarrow$ & MPJPE-p$\downarrow$ & PA-p$\downarrow$ & Jitter$\downarrow$ \\
\midrule
\multicolumn{8}{@{}l}{\textit{Detect-then-reconstruct, single frame}} \\
Hamba~\cite{S-dong2024hamba} & 67.95 & 85.01 & 86.61 & 31.40 & 35.63 & 14.17 & 10.07 \\
WildHands~\cite{S-prakash2024wildhands} & 91.94 & 95.14 & 61.21 & 17.44 & 26.83 & 10.39 & 12.89 \\
InterWild~\cite{S-moon2023interwild} & 98.68 & 99.26 & 45.21 & 13.88 & 21.07 & 8.25 & 14.61 \\
AnyHand~\cite{S-si2026anyhand} & 98.65 & 99.55 & 50.69 & 6.52 & 13.46 & 6.15 & 8.18 \\
HaMeR~\cite{S-pavlakos2024hamer} & 92.18 & 94.72 & 59.37 & 15.16 & 19.48 & 8.11 & 12.48 \\
EgoForce~\cite{S-millerdurai2026egoforce} & 94.37 & 98.28 & 43.62 & 10.76 & 16.08 & 6.70 & 10.00 \\
HandDGP~\cite{S-valassakis2024handdgp} & 98.64 & 99.54 & 41.21 & 9.59 & 28.24 & 10.63 & 8.44 \\
EgoHandICL~\cite{S-xie2026egohandicl} & 98.65 & 99.55 & 25.59 & 4.88 & 12.56 & 5.40 & 8.09 \\
WiLoR~\cite{S-potamias2025wilor} & 98.65 & 99.55 & 25.59 & 4.89 & 12.57 & 5.50 & 7.91 \\
\midrule
\multicolumn{8}{@{}l}{\textit{Detect-then-reconstruct with a motion prior}} \\
Dyn-HaMR~\cite{S-yu2025dynhamr} & 99.39 & 99.79 & 473.15 & 11.20 & 34.17 & 6.42 & 19.72 \\
OmniHands~\cite{S-lin2026omnihands} & 89.87 & 93.03 & 80.54 & 19.01 & 26.90 & 10.00 & 43.36 \\
HaWoR~\cite{S-zhang2025hawor} & 70.33 & 80.51 & 88.75 & 37.39 & 40.65 & 15.77 & 20.16 \\
HandFlow~\cite{S-xu2026handflow} & 98.90 & 99.72 & 31.07 & 7.42 & 14.86 & 6.39 & 11.29 \\
MINT~\cite{S-zhu2026mint} & \best{100.00} & \best{100.00} & 36.18 & 11.51 & 16.68 & 8.35 & 7.76 \\
PAD-Hand~\cite{S-ismayilzada2026padhand} & 98.70 & 99.58 & 45.54 & 16.75 & 15.58 & 6.54 & 7.61 \\
InfiniHand~\cite{S-ren2026infinihand} & 99.80 & 99.90 & 29.48 & 4.61 & 11.28 & 5.11 & 10.00 \\
\midrule
\multicolumn{8}{@{}l}{\textit{Video diffusion with depth regression}} \\
ViDiHand~\cite{S-wang2026vidihand} & 99.52 & 99.76 & 40.75 & 8.80 & 19.70 & 9.43 & 2.79 \\
ACE-Ego-Hand~\cite{S-liu2026aceegohand} & \best{100.00} & \best{100.00} & 21.26 & 3.77 & 9.51 & 5.04 & 1.12 \\
\midrule
\multicolumn{8}{@{}l}{\textit{Video diffusion with hand geometry}} \\
\textbf{\ours (Ours)} & \best{100.00} & \best{100.00} & \best{20.93} & \best{2.89} & \best{8.87} & \best{4.58} & \best{0.99} \\
\bottomrule
\end{tabular}
 }
\end{table}

\subsection{Baselines with Their Officially Released Weights}
\label{S-sec:supp_official}
\Cref{S-tab:supp_official} repeats Tab.~\ref{tab:main} with the officially released weights of every baseline that provides them, and
\cref{S-tab:supp_off_arctic,S-tab:supp_off_h2o,S-tab:supp_off_hot3d,S-tab:supp_off_oakink2,S-tab:supp_off_egopro,S-tab:supp_off_egostd} give
these weights on each in-domain dataset. \ours is the best on every metric, overall and on every dataset.

\begin{table}[!tbp]
\centering
\caption{\textbf{Comparison with the officially released weights.} The setting and format of Tab.~\ref{tab:main}, with each
baseline using its officially released weights where they exist. A $^\ast$ marks the methods trained on the same data as \ours,
namely \ours itself, \vidihand and \ace, whose released weights use this data, and EgoHandICL, which has no released weights for
these datasets. \ours is the \textbf{best} on every metric.}
\label{S-tab:supp_official}
\setlength{\tabcolsep}{3.2pt}
\adjustbox{max width=\textwidth}{%
\begin{tabular}{@{}lcccccccccccccc@{}}
\toprule
& \multicolumn{7}{c}{In-domain: mean of ARCTIC, H2O, HOT3D, OakInk2, Ego-Pro, Ego-Std} & \multicolumn{7}{c}{In-the-wild: \genhuman} \\
\cmidrule(lr){2-8}\cmidrule(lr){9-15}
& \multicolumn{2}{c}{Detection} & \multicolumn{2}{c}{Absolute pose} & \multicolumn{2}{c}{Relative pose} & \multicolumn{1}{c}{Temporal} & \multicolumn{2}{c}{Detection} & \multicolumn{2}{c}{Absolute pose} & \multicolumn{2}{c}{Relative pose} & \multicolumn{1}{c}{Temporal} \\
\cmidrule(lr){2-3}\cmidrule(lr){4-5}\cmidrule(lr){6-7}\cmidrule(lr){8-8}\cmidrule(lr){9-10}\cmidrule(lr){11-12}\cmidrule(lr){13-14}\cmidrule(lr){15-15}
Method & FAcc$\uparrow$ & F1$\uparrow$ & MPJPE-abs-p$\downarrow$ & EPE2D-p$\downarrow$ & MPJPE-p$\downarrow$ & PA-p$\downarrow$ & Jitter$\downarrow$ & FAcc$\uparrow$ & F1$\uparrow$ & MPJPE-abs-p$\downarrow$ & EPE2D-p$\downarrow$ & MPJPE-p$\downarrow$ & PA-p$\downarrow$ & Jitter$\downarrow$ \\
\midrule
\multicolumn{15}{@{}l}{\textit{Detect-then-reconstruct, single frame}} \\
Hamba~\cite{S-dong2024hamba} & 61.35 & 81.79 & 103.91 & 48.56 & 51.17 & 16.76 & 13.25 & 70.38 & 88.49 & 103.38 & 32.81 & 57.27 & 15.27 & 17.73 \\
WildHands~\cite{S-prakash2024wildhands} & 75.52 & 86.97 & 115.37 & 56.68 & 44.00 & 16.14 & 16.70 & 42.35 & 58.37 & 210.94 & 94.82 & 65.90 & 25.42 & 18.66 \\
InterWild~\cite{S-moon2023interwild} & 76.25 & 87.56 & 91.40 & 33.08 & 45.38 & 13.34 & 40.75 & 83.11 & 89.78 & 104.39 & 28.83 & 58.87 & 13.17 & 44.71 \\
AnyHand~\cite{S-si2026anyhand} & 92.26 & 97.21 & 60.32 & 16.08 & 25.27 & 8.96 & 13.22 & 91.80 & 96.75 & 63.69 & 16.34 & 44.39 & 10.16 & 19.21 \\
HaMeR~\cite{S-pavlakos2024hamer} & 84.71 & 93.14 & 75.53 & 25.68 & 40.86 & 11.33 & 19.06 & 94.57 & 97.57 & 63.18 & 14.37 & 48.52 & 9.73 & 24.58 \\
EgoForce~\cite{S-millerdurai2026egoforce} & 83.25 & 93.78 & 84.91 & 38.59 & 37.31 & 12.33 & 19.82 & 82.28 & 94.07 & 102.04 & 44.77 & 46.04 & 12.40 & 35.55 \\
HandDGP~\cite{S-valassakis2024handdgp} & 92.21 & 97.19 & 87.52 & 21.63 & 45.37 & 13.04 & 26.14 & 92.06 & 96.98 & 85.89 & 19.95 & 51.86 & 12.48 & 34.98 \\
EgoHandICL$^\ast$~\cite{S-xie2026egohandicl} & 92.27 & 97.21 & 46.89 & 13.29 & 17.14 & 7.89 & 12.32 & 91.94 & 96.89 & 57.07 & 14.55 & 28.21 & 9.66 & 15.05 \\
WiLoR~\cite{S-potamias2025wilor} & 92.26 & 97.21 & 53.47 & 14.95 & 26.55 & 8.96 & 12.40 & 91.83 & 96.77 & 64.75 & 16.38 & 46.75 & 10.10 & 17.49 \\
\midrule
\multicolumn{15}{@{}l}{\textit{Detect-then-reconstruct with a motion prior}} \\
Dyn-HaMR~\cite{S-yu2025dynhamr} & 89.33 & 96.19 & 170.93 & 14.24 & 25.18 & 9.37 & 14.08 & 92.72 & 96.30 & 233.23 & 14.83 & 26.79 & 10.60 & 24.76 \\
OmniHands~\cite{S-lin2026omnihands} & 76.92 & 89.08 & 105.95 & 30.46 & 39.60 & 11.80 & 64.41 & 93.36 & 96.62 & 96.12 & 20.60 & 46.56 & 10.57 & 52.46 \\
HaWoR~\cite{S-zhang2025hawor} & 71.35 & 83.86 & 114.22 & 41.23 & 41.94 & 14.76 & 24.47 & 71.89 & 81.47 & 141.30 & 40.02 & 48.63 & 17.02 & 29.44 \\
HandFlow~\cite{S-xu2026handflow} & 92.43 & 97.43 & 100.53 & 20.92 & 28.53 & 10.25 & 4.54 & 91.38 & 96.95 & 132.59 & 16.07 & 25.26 & 10.86 & 6.25 \\
MINT~\cite{S-zhu2026mint} & 95.76 & 98.52 & 70.84 & 20.10 & 25.51 & 9.23 & 10.13 & 98.05 & 99.08 & 55.22 & 13.42 & 22.50 & 7.66 & 9.40 \\
PAD-Hand~\cite{S-ismayilzada2026padhand} & 91.47 & 96.95 & 71.59 & 25.99 & 31.60 & 11.05 & 12.05 & 91.09 & 96.40 & 73.19 & 23.05 & 49.96 & 10.33 & 14.96 \\
InfiniHand~\cite{S-ren2026infinihand} & 98.29 & 99.37 & 53.49 & 11.36 & 20.61 & 8.40 & 18.45 & 97.81 & 99.16 & 101.65 & 12.65 & 24.82 & 8.09 & 27.88 \\
\midrule
\multicolumn{15}{@{}l}{\textit{Video diffusion with depth regression}} \\
ViDiHand$^\ast$~\cite{S-wang2026vidihand} & 98.98 & 99.59 & 36.60 & 9.92 & 18.64 & 8.29 & 2.93 & 99.03 & 99.60 & 94.53 & 11.84 & 29.30 & 7.45 & 4.08 \\
ACE-Ego-Hand$^\ast$~\cite{S-liu2026aceegohand} & 98.88 & 99.64 & 24.46 & 6.81 & 10.84 & 5.60 & 1.44 & 95.99 & 97.65 & 92.13 & 26.74 & 22.98 & 8.73 & 1.76 \\
\midrule
\multicolumn{15}{@{}l}{\textit{Video diffusion with hand geometry}} \\
\textbf{\ours (Ours)}$^\ast$ & \best{99.93} & \best{99.98} & \best{19.62} & \best{4.67} & \best{9.50} & \best{4.92} & \best{1.26} & \best{99.89} & \best{99.96} & \best{32.59} & \best{6.50} & \best{18.59} & \best{6.99} & \best{1.61} \\
\bottomrule
\end{tabular}
 }
\end{table}

\begin{table}[!tbp]
\centering
\setlength{\tabcolsep}{3.2pt}
\captionof{table}{\textbf{Comparison on ARCTIC with the officially released weights.} Each baseline uses its officially released weights where they exist, and every method is tested on the ARCTIC test split. A $^\ast$ marks the methods trained on the same data as \ours. Metrics with the suffix -p add a penalty for each detection failure. FAcc and F1 in \%, EPE2D-p in px, Jitter in mm/frame$^2$, other errors in mm, best values in bold. \ours is the \textbf{best} on every metric.}
\label{S-tab:supp_off_arctic}
\adjustbox{max width=0.8\textwidth,max totalheight=0.37\textheight}{%
\begin{tabular}{@{}lccccccc@{}}
\toprule
& \multicolumn{2}{c}{Detection} & \multicolumn{2}{c}{Absolute pose} & \multicolumn{2}{c}{Relative pose} & \multicolumn{1}{c}{Temporal} \\
\cmidrule(lr){2-3}\cmidrule(lr){4-5}\cmidrule(lr){6-7}\cmidrule(lr){8-8}
Method & FAcc$\uparrow$ & F1$\uparrow$ & MPJPE-abs-p$\downarrow$ & EPE2D-p$\downarrow$ & MPJPE-p$\downarrow$ & PA-p$\downarrow$ & Jitter$\downarrow$ \\
\midrule
\multicolumn{8}{@{}l}{\textit{Detect-then-reconstruct, single frame}} \\
Hamba~\cite{S-dong2024hamba} & 75.19 & 89.00 & 75.02 & 37.31 & 35.12 & 14.97 & 15.33 \\
WildHands~\cite{S-prakash2024wildhands} & 84.26 & 93.87 & 55.93 & 30.65 & 28.67 & 12.93 & 12.69 \\
InterWild~\cite{S-moon2023interwild} & 72.24 & 85.83 & 83.74 & 35.32 & 34.00 & 14.59 & 49.13 \\
AnyHand~\cite{S-si2026anyhand} & 89.75 & 96.56 & 54.71 & 19.63 & 23.86 & 10.08 & 15.64 \\
HaMeR~\cite{S-pavlakos2024hamer} & 85.45 & 94.28 & 63.74 & 26.17 & 30.34 & 12.86 & 19.52 \\
EgoForce~\cite{S-millerdurai2026egoforce} & 85.52 & 95.12 & 56.92 & 28.00 & 24.46 & 12.20 & 15.65 \\
HandDGP~\cite{S-valassakis2024handdgp} & 89.72 & 96.54 & 89.62 & 28.98 & 44.42 & 15.71 & 35.14 \\
EgoHandICL$^\ast$~\cite{S-xie2026egohandicl} & 89.75 & 96.56 & 61.09 & 20.50 & 20.69 & 10.93 & 17.91 \\
WiLoR~\cite{S-potamias2025wilor} & 89.75 & 96.56 & 47.32 & 18.85 & 23.55 & 10.12 & 18.61 \\
\midrule
\multicolumn{8}{@{}l}{\textit{Detect-then-reconstruct with a motion prior}} \\
Dyn-HaMR~\cite{S-yu2025dynhamr} & 93.49 & 97.31 & 83.45 & 17.74 & 26.62 & 10.86 & 15.09 \\
OmniHands~\cite{S-lin2026omnihands} & 76.66 & 88.92 & 81.40 & 27.37 & 31.28 & 11.79 & 37.09 \\
HaWoR~\cite{S-zhang2025hawor} & 76.83 & 89.13 & 91.45 & 46.08 & 40.70 & 13.27 & 28.45 \\
HandFlow~\cite{S-xu2026handflow} & 89.61 & 96.56 & 122.60 & 34.67 & 39.48 & 12.94 & 5.65 \\
MINT~\cite{S-zhu2026mint} & 90.86 & 96.87 & 93.24 & 32.45 & 32.00 & 11.78 & 14.36 \\
PAD-Hand~\cite{S-ismayilzada2026padhand} & 86.24 & 95.43 & 63.71 & 33.91 & 30.25 & 14.75 & 18.12 \\
InfiniHand~\cite{S-ren2026infinihand} & 97.87 & 99.21 & 46.10 & 14.08 & 22.23 & 9.13 & 26.16 \\
\midrule
\multicolumn{8}{@{}l}{\textit{Video diffusion with depth regression}} \\
ViDiHand$^\ast$~\cite{S-wang2026vidihand} & 98.89 & 99.67 & 46.55 & 12.99 & 21.66 & 10.02 & 3.81 \\
ACE-Ego-Hand$^\ast$~\cite{S-liu2026aceegohand} & 98.81 & 99.65 & 37.95 & 11.00 & 15.90 & 7.25 & 2.52 \\
\midrule
\multicolumn{8}{@{}l}{\textit{Video diffusion with hand geometry}} \\
\textbf{\ours (Ours)}$^\ast$ & \best{99.90} & \best{99.97} & \best{33.33} & \best{7.72} & \best{14.19} & \best{6.89} & \best{2.30} \\
\bottomrule
\end{tabular}
 }
\end{table}
\begin{table}[!tbp]
\centering
\setlength{\tabcolsep}{3.2pt}
\captionof{table}{\textbf{Comparison on H2O with the officially released weights.} Each baseline uses its officially released weights where they exist, and every method is tested on the H2O test split. A $^\ast$ marks the methods trained on the same data as \ours. Metrics with the suffix -p add a penalty for each detection failure. FAcc and F1 in \%, EPE2D-p in px, Jitter in mm/frame$^2$, other errors in mm, best values in bold. \ours is the \textbf{best} on every metric. Other methods tie with it only where they also reach 100\% FAcc or F1.}
\label{S-tab:supp_off_h2o}
\adjustbox{max width=0.8\textwidth,max totalheight=0.37\textheight}{%
\begin{tabular}{@{}lccccccc@{}}
\toprule
& \multicolumn{2}{c}{Detection} & \multicolumn{2}{c}{Absolute pose} & \multicolumn{2}{c}{Relative pose} & \multicolumn{1}{c}{Temporal} \\
\cmidrule(lr){2-3}\cmidrule(lr){4-5}\cmidrule(lr){6-7}\cmidrule(lr){8-8}
Method & FAcc$\uparrow$ & F1$\uparrow$ & MPJPE-abs-p$\downarrow$ & EPE2D-p$\downarrow$ & MPJPE-p$\downarrow$ & PA-p$\downarrow$ & Jitter$\downarrow$ \\
\midrule
\multicolumn{8}{@{}l}{\textit{Detect-then-reconstruct, single frame}} \\
Hamba~\cite{S-dong2024hamba} & 88.59 & 95.90 & 43.64 & 19.80 & 25.18 & 9.06 & 7.67 \\
WildHands~\cite{S-prakash2024wildhands} & 99.24 & 99.69 & 71.59 & 21.55 & 30.76 & 9.85 & 10.56 \\
InterWild~\cite{S-moon2023interwild} & 88.59 & 91.53 & 74.16 & 33.85 & 29.48 & 11.92 & 22.63 \\
AnyHand~\cite{S-si2026anyhand} & 99.66 & 99.90 & 31.76 & 8.26 & 17.08 & 5.66 & 7.97 \\
HaMeR~\cite{S-pavlakos2024hamer} & 98.54 & 99.37 & 33.13 & 10.52 & 20.39 & 6.72 & 8.54 \\
EgoForce~\cite{S-millerdurai2026egoforce} & 99.42 & 99.85 & 26.06 & 12.10 & 15.55 & 5.95 & 5.51 \\
HandDGP~\cite{S-valassakis2024handdgp} & 99.67 & 99.90 & 73.50 & 12.83 & 30.27 & 8.24 & 8.72 \\
EgoHandICL$^\ast$~\cite{S-xie2026egohandicl} & 99.67 & 99.90 & 28.09 & 7.22 & 10.76 & 4.83 & 7.26 \\
WiLoR~\cite{S-potamias2025wilor} & 99.66 & 99.90 & 25.99 & 7.78 & 16.76 & 5.50 & 5.48 \\
\midrule
\multicolumn{8}{@{}l}{\textit{Detect-then-reconstruct with a motion prior}} \\
Dyn-HaMR~\cite{S-yu2025dynhamr} & 99.93 & 99.96 & 32.49 & 6.98 & 15.86 & 6.33 & 6.56 \\
OmniHands~\cite{S-lin2026omnihands} & 98.74 & 99.37 & 48.99 & 16.16 & 18.86 & 6.10 & 10.08 \\
HaWoR~\cite{S-zhang2025hawor} & 99.45 & 99.81 & 41.94 & 12.89 & 21.18 & 7.14 & 16.47 \\
HandFlow~\cite{S-xu2026handflow} & 99.71 & 99.93 & 50.57 & 14.78 & 23.93 & 7.33 & 3.45 \\
MINT~\cite{S-zhu2026mint} & 99.99 & 99.99 & 42.45 & 16.19 & 18.20 & 6.85 & 7.20 \\
PAD-Hand~\cite{S-ismayilzada2026padhand} & 99.74 & 99.91 & 33.08 & 13.86 & 20.53 & 6.98 & 6.78 \\
InfiniHand~\cite{S-ren2026infinihand} & 99.99 & \best{100.00} & 48.78 & 11.16 & 18.82 & 7.36 & 13.80 \\
\midrule
\multicolumn{8}{@{}l}{\textit{Video diffusion with depth regression}} \\
ViDiHand$^\ast$~\cite{S-wang2026vidihand} & \best{100.00} & \best{100.00} & 27.27 & 7.96 & 14.95 & 6.01 & 1.74 \\
ACE-Ego-Hand$^\ast$~\cite{S-liu2026aceegohand} & \best{100.00} & \best{100.00} & 26.75 & 7.29 & 8.18 & 4.37 & 0.73 \\
\midrule
\multicolumn{8}{@{}l}{\textit{Video diffusion with hand geometry}} \\
\textbf{\ours (Ours)}$^\ast$ & \best{100.00} & \best{100.00} & \best{11.99} & \best{4.10} & \best{7.84} & \best{4.10} & \best{0.59} \\
\bottomrule
\end{tabular}
 }
\end{table}
\begin{table}[!tbp]
\centering
\setlength{\tabcolsep}{3.2pt}
\captionof{table}{\textbf{Comparison on HOT3D with the officially released weights.} Each baseline uses its officially released weights where they exist, and every method is tested on the HOT3D test split. A $^\ast$ marks the methods trained on the same data as \ours. Metrics with the suffix -p add a penalty for each detection failure. FAcc and F1 in \%, EPE2D-p in px, Jitter in mm/frame$^2$, other errors in mm, best values in bold. \ours is the \textbf{best} on every metric.}
\label{S-tab:supp_off_hot3d}
\adjustbox{max width=0.8\textwidth,max totalheight=0.37\textheight}{%
\begin{tabular}{@{}lccccccc@{}}
\toprule
& \multicolumn{2}{c}{Detection} & \multicolumn{2}{c}{Absolute pose} & \multicolumn{2}{c}{Relative pose} & \multicolumn{1}{c}{Temporal} \\
\cmidrule(lr){2-3}\cmidrule(lr){4-5}\cmidrule(lr){6-7}\cmidrule(lr){8-8}
Method & FAcc$\uparrow$ & F1$\uparrow$ & MPJPE-abs-p$\downarrow$ & EPE2D-p$\downarrow$ & MPJPE-p$\downarrow$ & PA-p$\downarrow$ & Jitter$\downarrow$ \\
\midrule
\multicolumn{8}{@{}l}{\textit{Detect-then-reconstruct, single frame}} \\
Hamba~\cite{S-dong2024hamba} & 40.29 & 73.65 & 165.36 & 54.19 & 78.03 & 20.69 & 17.61 \\
WildHands~\cite{S-prakash2024wildhands} & 57.37 & 79.10 & 174.01 & 62.29 & 58.71 & 20.69 & 24.31 \\
InterWild~\cite{S-moon2023interwild} & 51.90 & 78.73 & 160.12 & 36.47 & 78.47 & 16.16 & 70.78 \\
AnyHand~\cite{S-si2026anyhand} & 81.41 & 93.04 & 117.16 & 24.46 & 27.64 & 11.69 & 17.13 \\
HaMeR~\cite{S-pavlakos2024hamer} & 67.55 & 87.39 & 112.33 & 26.32 & 69.61 & 14.48 & 25.32 \\
EgoForce~\cite{S-millerdurai2026egoforce} & 72.64 & 88.94 & 151.32 & 54.88 & 50.24 & 15.28 & 27.89 \\
HandDGP~\cite{S-valassakis2024handdgp} & 81.14 & 92.92 & 175.01 & 33.33 & 75.20 & 18.02 & 53.00 \\
EgoHandICL$^\ast$~\cite{S-xie2026egohandicl} & 81.41 & 93.03 & 94.54 & 21.75 & 27.87 & 11.94 & 21.91 \\
WiLoR~\cite{S-potamias2025wilor} & 81.41 & 93.03 & 86.06 & 21.71 & 30.93 & 11.88 & 17.88 \\
\midrule
\multicolumn{8}{@{}l}{\textit{Detect-then-reconstruct with a motion prior}} \\
Dyn-HaMR~\cite{S-yu2025dynhamr} & 62.40 & 86.88 & 114.59 & 22.83 & 40.92 & 14.41 & 14.82 \\
OmniHands~\cite{S-lin2026omnihands} & 54.79 & 82.67 & 169.43 & 29.33 & 63.26 & 13.71 & 99.21 \\
HaWoR~\cite{S-zhang2025hawor} & 52.97 & 75.27 & 158.35 & 51.94 & 49.19 & 17.83 & 24.10 \\
HandFlow~\cite{S-xu2026handflow} & 82.70 & 93.90 & 107.01 & 20.66 & 26.35 & 10.96 & 4.98 \\
MINT~\cite{S-zhu2026mint} & 92.50 & 97.53 & 68.18 & 17.30 & 25.17 & 10.02 & 12.35 \\
PAD-Hand~\cite{S-ismayilzada2026padhand} & 81.19 & 92.97 & 114.63 & 42.67 & 38.13 & 13.51 & 17.09 \\
InfiniHand~\cite{S-ren2026infinihand} & 97.20 & 99.08 & 46.64 & 9.61 & 24.97 & 9.68 & 18.04 \\
\midrule
\multicolumn{8}{@{}l}{\textit{Video diffusion with depth regression}} \\
ViDiHand$^\ast$~\cite{S-wang2026vidihand} & 98.76 & 99.61 & 35.32 & 6.92 & 19.05 & 8.38 & 3.96 \\
ACE-Ego-Hand$^\ast$~\cite{S-liu2026aceegohand} & 98.18 & 99.44 & 24.26 & 5.33 & 12.78 & 6.42 & 2.31 \\
\midrule
\multicolumn{8}{@{}l}{\textit{Video diffusion with hand geometry}} \\
\textbf{\ours (Ours)}$^\ast$ & \best{99.93} & \best{99.98} & \best{23.02} & \best{3.90} & \best{11.25} & \best{5.56} & \best{2.04} \\
\bottomrule
\end{tabular}
 }
\end{table}
\begin{table}[!tbp]
\centering
\setlength{\tabcolsep}{3.2pt}
\captionof{table}{\textbf{Comparison on OakInk2 with the officially released weights.} Each baseline uses its officially released weights where they exist, and every method is tested on the OakInk2 test split. A $^\ast$ marks the methods trained on the same data as \ours. Metrics with the suffix -p add a penalty for each detection failure. FAcc and F1 in \%, EPE2D-p in px, Jitter in mm/frame$^2$, other errors in mm, best values in bold. \ours is the \textbf{best} on every metric.}
\label{S-tab:supp_off_oakink2}
\adjustbox{max width=0.8\textwidth,max totalheight=0.37\textheight}{%
\begin{tabular}{@{}lccccccc@{}}
\toprule
& \multicolumn{2}{c}{Detection} & \multicolumn{2}{c}{Absolute pose} & \multicolumn{2}{c}{Relative pose} & \multicolumn{1}{c}{Temporal} \\
\cmidrule(lr){2-3}\cmidrule(lr){4-5}\cmidrule(lr){6-7}\cmidrule(lr){8-8}
Method & FAcc$\uparrow$ & F1$\uparrow$ & MPJPE-abs-p$\downarrow$ & EPE2D-p$\downarrow$ & MPJPE-p$\downarrow$ & PA-p$\downarrow$ & Jitter$\downarrow$ \\
\midrule
\multicolumn{8}{@{}l}{\textit{Detect-then-reconstruct, single frame}} \\
Hamba~\cite{S-dong2024hamba} & 39.91 & 65.52 & 144.60 & 111.16 & 57.71 & 22.94 & 12.60 \\
WildHands~\cite{S-prakash2024wildhands} & 66.05 & 83.42 & 137.46 & 91.37 & 49.64 & 19.34 & 13.56 \\
InterWild~\cite{S-moon2023interwild} & 54.26 & 74.59 & 112.01 & 62.83 & 44.43 & 15.94 & 40.10 \\
AnyHand~\cite{S-si2026anyhand} & 89.94 & 96.16 & 61.90 & 23.65 & 29.65 & 10.15 & 12.37 \\
HaMeR~\cite{S-pavlakos2024hamer} & 68.05 & 84.94 & 119.49 & 68.85 & 46.04 & 16.52 & 24.71 \\
EgoForce~\cite{S-millerdurai2026egoforce} & 83.42 & 93.28 & 104.15 & 58.85 & 42.18 & 13.46 & 19.83 \\
HandDGP~\cite{S-valassakis2024handdgp} & 89.92 & 96.15 & 84.89 & 30.46 & 49.87 & 13.00 & 15.91 \\
EgoHandICL$^\ast$~\cite{S-xie2026egohandicl} & 89.97 & 96.19 & 45.01 & 18.21 & 17.65 & 8.06 & 10.36 \\
WiLoR~\cite{S-potamias2025wilor} & 89.94 & 96.16 & 68.53 & 24.36 & 30.22 & 10.12 & 9.59 \\
\midrule
\multicolumn{8}{@{}l}{\textit{Detect-then-reconstruct with a motion prior}} \\
Dyn-HaMR~\cite{S-yu2025dynhamr} & 87.08 & 95.07 & 207.24 & 23.45 & 28.36 & 9.84 & 15.85 \\
OmniHands~\cite{S-lin2026omnihands} & 46.15 & 73.07 & 181.93 & 71.22 & 48.70 & 16.08 & 162.30 \\
HaWoR~\cite{S-zhang2025hawor} & 78.40 & 90.11 & 116.12 & 46.29 & 35.60 & 11.54 & 19.02 \\
HandFlow~\cite{S-xu2026handflow} & 89.94 & 96.33 & 93.84 & 37.34 & 33.89 & 11.47 & 4.67 \\
MINT~\cite{S-zhu2026mint} & 91.42 & 96.84 & 100.95 & 37.47 & 30.44 & 10.50 & 9.41 \\
PAD-Hand~\cite{S-ismayilzada2026padhand} & 88.86 & 95.78 & 86.64 & 32.90 & 33.08 & 11.85 & 9.65 \\
InfiniHand~\cite{S-ren2026infinihand} & 94.71 & 97.92 & 82.92 & 23.54 & 28.86 & 11.19 & 24.46 \\
\midrule
\multicolumn{8}{@{}l}{\textit{Video diffusion with depth regression}} \\
ViDiHand$^\ast$~\cite{S-wang2026vidihand} & 98.33 & 99.32 & 29.31 & 13.13 & 15.16 & 6.39 & 2.23 \\
ACE-Ego-Hand$^\ast$~\cite{S-liu2026aceegohand} & 96.30 & 98.75 & 21.24 & 9.84 & 10.55 & 5.52 & 0.82 \\
\midrule
\multicolumn{8}{@{}l}{\textit{Video diffusion with hand geometry}} \\
\textbf{\ours (Ours)}$^\ast$ & \best{99.75} & \best{99.91} & \best{14.54} & \best{6.66} & \best{7.56} & \best{3.89} & \best{0.59} \\
\bottomrule
\end{tabular}
 }
\end{table}
\begin{table}[!tbp]
\centering
\setlength{\tabcolsep}{3.2pt}
\captionof{table}{\textbf{Comparison on Ego-Pro with the officially released weights.} Each baseline uses its officially released weights where they exist, and every method is tested on the Ego-Pro test split. A $^\ast$ marks the methods trained on the same data as \ours. Metrics with the suffix -p add a penalty for each detection failure. FAcc and F1 in \%, EPE2D-p in px, Jitter in mm/frame$^2$, other errors in mm, best values in bold. \ours is the \textbf{best} on every metric. Other methods tie with it only where they also reach 100\% FAcc or F1.}
\label{S-tab:supp_off_egopro}
\adjustbox{max width=0.8\textwidth,max totalheight=0.37\textheight}{%
\begin{tabular}{@{}lccccccc@{}}
\toprule
& \multicolumn{2}{c}{Detection} & \multicolumn{2}{c}{Absolute pose} & \multicolumn{2}{c}{Relative pose} & \multicolumn{1}{c}{Temporal} \\
\cmidrule(lr){2-3}\cmidrule(lr){4-5}\cmidrule(lr){6-7}\cmidrule(lr){8-8}
Method & FAcc$\uparrow$ & F1$\uparrow$ & MPJPE-abs-p$\downarrow$ & EPE2D-p$\downarrow$ & MPJPE-p$\downarrow$ & PA-p$\downarrow$ & Jitter$\downarrow$ \\
\midrule
\multicolumn{8}{@{}l}{\textit{Detect-then-reconstruct, single frame}} \\
Hamba~\cite{S-dong2024hamba} & 56.77 & 82.19 & 111.01 & 35.87 & 56.01 & 17.08 & 14.12 \\
WildHands~\cite{S-prakash2024wildhands} & 69.69 & 80.36 & 127.61 & 65.37 & 52.39 & 17.70 & 24.00 \\
InterWild~\cite{S-moon2023interwild} & 94.91 & 97.22 & 58.11 & 14.59 & 44.35 & 10.64 & 35.00 \\
AnyHand~\cite{S-si2026anyhand} & 94.15 & 98.05 & 42.27 & 10.47 & 29.19 & 7.97 & 14.08 \\
HaMeR~\cite{S-pavlakos2024hamer} & 96.47 & 98.17 & 72.53 & 8.83 & 36.79 & 7.71 & 18.93 \\
EgoForce~\cite{S-millerdurai2026egoforce} & 67.31 & 88.04 & 106.56 & 46.31 & 54.00 & 15.67 & 28.01 \\
HandDGP~\cite{S-valassakis2024handdgp} & 94.15 & 98.06 & 57.95 & 13.55 & 42.48 & 12.31 & 27.33 \\
EgoHandICL$^\ast$~\cite{S-xie2026egohandicl} & 94.15 & 98.06 & 27.00 & 7.18 & 13.33 & 6.18 & 8.36 \\
WiLoR~\cite{S-potamias2025wilor} & 94.15 & 98.05 & 56.24 & 9.34 & 30.61 & 8.03 & 12.58 \\
\midrule
\multicolumn{8}{@{}l}{\textit{Detect-then-reconstruct with a motion prior}} \\
Dyn-HaMR~\cite{S-yu2025dynhamr} & 98.24 & 99.26 & 103.87 & 5.50 & 18.17 & 6.78 & 11.38 \\
OmniHands~\cite{S-lin2026omnihands} & 95.84 & 97.65 & 62.51 & 17.22 & 35.52 & 10.56 & 31.40 \\
HaWoR~\cite{S-zhang2025hawor} & 62.94 & 77.27 & 149.02 & 39.97 & 50.94 & 17.97 & 32.10 \\
HandFlow~\cite{S-xu2026handflow} & 93.81 & 98.22 & 135.83 & 10.35 & 24.30 & 9.46 & 4.27 \\
MINT~\cite{S-zhu2026mint} & 99.81 & 99.94 & 72.86 & 8.20 & 23.41 & 7.69 & 9.07 \\
PAD-Hand~\cite{S-ismayilzada2026padhand} & 94.08 & 98.04 & 80.05 & 15.73 & 34.89 & 9.67 & 10.91 \\
InfiniHand~\cite{S-ren2026infinihand} & 99.99 & \best{100.00} & 57.05 & 5.02 & 15.29 & 6.37 & 15.39 \\
\midrule
\multicolumn{8}{@{}l}{\textit{Video diffusion with depth regression}} \\
ViDiHand$^\ast$~\cite{S-wang2026vidihand} & 98.36 & 99.18 & 40.39 & 9.71 & 21.31 & 9.52 & 3.06 \\
ACE-Ego-Hand$^\ast$~\cite{S-liu2026aceegohand} & \best{100.00} & \best{100.00} & 15.32 & 3.60 & 8.14 & 5.03 & 1.16 \\
\midrule
\multicolumn{8}{@{}l}{\textit{Video diffusion with hand geometry}} \\
\textbf{\ours (Ours)}$^\ast$ & \best{100.00} & \best{100.00} & \best{13.90} & \best{2.71} & \best{7.29} & \best{4.51} & \best{1.06} \\
\bottomrule
\end{tabular}
 }
\end{table}
\begin{table}[!tbp]
\centering
\setlength{\tabcolsep}{3.2pt}
\captionof{table}{\textbf{Comparison on Ego-Std with the officially released weights.} Each baseline uses its officially released weights where they exist, and every method is tested on the Ego-Std test split. A $^\ast$ marks the methods trained on the same data as \ours. Metrics with the suffix -p add a penalty for each detection failure. FAcc and F1 in \%, EPE2D-p in px, Jitter in mm/frame$^2$, other errors in mm, best values in bold. \ours is the \textbf{best} on every metric. Other methods tie with it only where they also reach 100\% FAcc or F1.}
\label{S-tab:supp_off_egostd}
\adjustbox{max width=0.8\textwidth,max totalheight=0.37\textheight}{%
\begin{tabular}{@{}lccccccc@{}}
\toprule
& \multicolumn{2}{c}{Detection} & \multicolumn{2}{c}{Absolute pose} & \multicolumn{2}{c}{Relative pose} & \multicolumn{1}{c}{Temporal} \\
\cmidrule(lr){2-3}\cmidrule(lr){4-5}\cmidrule(lr){6-7}\cmidrule(lr){8-8}
Method & FAcc$\uparrow$ & F1$\uparrow$ & MPJPE-abs-p$\downarrow$ & EPE2D-p$\downarrow$ & MPJPE-p$\downarrow$ & PA-p$\downarrow$ & Jitter$\downarrow$ \\
\midrule
\multicolumn{8}{@{}l}{\textit{Detect-then-reconstruct, single frame}} \\
Hamba~\cite{S-dong2024hamba} & 67.35 & 84.49 & 83.82 & 33.01 & 54.99 & 15.81 & 12.19 \\
WildHands~\cite{S-prakash2024wildhands} & 76.48 & 85.37 & 125.60 & 68.82 & 43.83 & 16.35 & 15.09 \\
InterWild~\cite{S-moon2023interwild} & 95.59 & 97.48 & 60.24 & 15.44 & 41.56 & 10.78 & 26.84 \\
AnyHand~\cite{S-si2026anyhand} & 98.64 & 99.54 & 54.09 & 10.03 & 24.19 & 8.19 & 12.14 \\
HaMeR~\cite{S-pavlakos2024hamer} & 92.18 & 94.70 & 51.98 & 13.43 & 41.99 & 9.70 & 17.32 \\
EgoForce~\cite{S-millerdurai2026egoforce} & 91.20 & 97.45 & 64.42 & 31.39 & 37.47 & 11.41 & 22.04 \\
HandDGP~\cite{S-valassakis2024handdgp} & 98.64 & 99.55 & 44.13 & 10.64 & 29.96 & 10.95 & 16.73 \\
EgoHandICL$^\ast$~\cite{S-xie2026egohandicl} & 98.65 & 99.55 & 25.59 & 4.88 & 12.56 & 5.40 & 8.09 \\
WiLoR~\cite{S-potamias2025wilor} & 98.64 & 99.54 & 36.66 & 7.64 & 27.21 & 8.14 & 10.25 \\
\midrule
\multicolumn{8}{@{}l}{\textit{Detect-then-reconstruct with a motion prior}} \\
Dyn-HaMR~\cite{S-yu2025dynhamr} & 94.81 & 98.63 & 483.95 & 8.93 & 21.15 & 8.00 & 20.77 \\
OmniHands~\cite{S-lin2026omnihands} & 89.35 & 92.82 & 91.46 & 21.45 & 39.98 & 12.57 & 46.37 \\
HaWoR~\cite{S-zhang2025hawor} & 57.53 & 71.55 & 128.45 & 50.19 & 54.01 & 20.79 & 26.69 \\
HandFlow~\cite{S-xu2026handflow} & 98.80 & 99.66 & 93.32 & 7.72 & 23.26 & 9.36 & 4.19 \\
MINT~\cite{S-zhu2026mint} & 99.98 & 99.98 & 47.34 & 9.00 & 23.82 & 8.52 & 8.36 \\
PAD-Hand~\cite{S-ismayilzada2026padhand} & 98.70 & 99.57 & 51.40 & 16.88 & 32.69 & 9.54 & 9.75 \\
InfiniHand~\cite{S-ren2026infinihand} & \best{100.00} & \best{100.00} & 39.45 & 4.78 & 13.46 & 6.67 & 12.88 \\
\midrule
\multicolumn{8}{@{}l}{\textit{Video diffusion with depth regression}} \\
ViDiHand$^\ast$~\cite{S-wang2026vidihand} & 99.52 & 99.76 & 40.75 & 8.80 & 19.70 & 9.43 & 2.79 \\
ACE-Ego-Hand$^\ast$~\cite{S-liu2026aceegohand} & \best{100.00} & \best{100.00} & 21.26 & 3.77 & 9.51 & 5.04 & 1.12 \\
\midrule
\multicolumn{8}{@{}l}{\textit{Video diffusion with hand geometry}} \\
\textbf{\ours (Ours)}$^\ast$ & \best{100.00} & \best{100.00} & \best{20.93} & \best{2.89} & \best{8.87} & \best{4.58} & \best{0.99} \\
\bottomrule
\end{tabular}
 }
\end{table}

\subsection{Hands Detected by All Methods, Baselines Trained on the Same Six Datasets}
\label{S-sec:supp_common}
\Cref{S-tab:supp_common_tp} compares \ours and the re-trained baselines of Tab.~\ref{tab:main} only on the hands that all 19
methods detect, 848{,}345 of the 1{,}446{,}795 ground-truth hand-frames, or 59\%. Detection failures are not penalized, and the errors carry no -p suffix. \ours is the best in all ten columns, with an MPJPE-abs of 17.42\,mm
in-domain and 28.09\,mm in the wild. The two video diffusion baselines, which regress the depth, still reach 79.15 and 90.47\,mm in
the wild on these hands, so their large in-the-wild error does not come from detection.

\begin{table}[!tbp]
\centering
\caption{\textbf{Comparison on hands detected by all methods, re-trained baselines.} The setting and units of
Tab.~\ref{tab:main}, evaluated only on the hands that all 19 methods detect, so detection failures are not penalized and the
errors carry no -p suffix. \ours is the \textbf{best} on every metric.}
\label{S-tab:supp_common_tp}
\setlength{\tabcolsep}{3.2pt}
\adjustbox{max width=\textwidth,max totalheight=0.37\textheight}{%
\begin{tabular}{@{}lcccccccccc@{}}
\toprule
& \multicolumn{5}{c}{In-domain: mean of 6 datasets} & \multicolumn{5}{c}{In-the-wild: \genhuman} \\
\cmidrule(lr){2-6}\cmidrule(lr){7-11}
& \multicolumn{2}{c}{Absolute pose} & \multicolumn{2}{c}{Relative pose} & \multicolumn{1}{c}{Temporal} & \multicolumn{2}{c}{Absolute pose} & \multicolumn{2}{c}{Relative pose} & \multicolumn{1}{c}{Temporal} \\
\cmidrule(lr){2-3}\cmidrule(lr){4-5}\cmidrule(lr){6-6}\cmidrule(lr){7-8}\cmidrule(lr){9-10}\cmidrule(lr){11-11}
Method & MPJPE-abs$\downarrow$ & EPE2D$\downarrow$ & MPJPE$\downarrow$ & PA$\downarrow$ & Jitter$\downarrow$ & MPJPE-abs$\downarrow$ & EPE2D$\downarrow$ & MPJPE$\downarrow$ & PA$\downarrow$ & Jitter$\downarrow$ \\
\midrule
\multicolumn{11}{@{}l}{\textit{Detect-then-reconstruct, single frame}} \\
Hamba~\cite{S-dong2024hamba} & 41.74 & 8.55 & 12.98 & 6.21 & 10.77 & 61.27 & 8.44 & 22.45 & 7.18 & 13.53 \\
WildHands~\cite{S-prakash2024wildhands} & 49.14 & 16.78 & 27.73 & 8.91 & 14.25 & 74.37 & 14.53 & 35.78 & 7.55 & 13.11 \\
InterWild~\cite{S-moon2023interwild} & 53.28 & 17.04 & 24.46 & 8.92 & 17.57 & 67.66 & 18.26 & 32.46 & 7.44 & 21.10 \\
AnyHand~\cite{S-si2026anyhand} & 45.81 & 7.90 & 12.14 & 6.26 & 9.04 & 65.30 & 7.59 & 22.31 & 7.50 & 12.47 \\
HaMeR~\cite{S-pavlakos2024hamer} & 43.55 & 9.77 & 13.55 & 6.37 & 11.44 & 61.11 & 10.25 & 22.59 & 7.76 & 13.68 \\
EgoForce~\cite{S-millerdurai2026egoforce} & 32.81 & 9.69 & 13.47 & 5.98 & 10.88 & 36.67 & 11.00 & 20.04 & 7.59 & 14.08 \\
HandDGP~\cite{S-valassakis2024handdgp} & 45.28 & 12.30 & 28.85 & 11.64 & 9.99 & 45.45 & 9.87 & 29.89 & 11.16 & 11.57 \\
EgoHandICL~\cite{S-xie2026egohandicl} & 28.54 & 6.31 & 11.16 & 5.52 & 10.64 & 30.67 & 7.05 & 22.01 & 7.60 & 11.49 \\
WiLoR~\cite{S-potamias2025wilor} & 28.89 & 6.10 & 10.97 & 5.48 & 9.88 & 30.26 & 6.96 & 21.88 & 7.74 & 11.17 \\
\midrule
\multicolumn{11}{@{}l}{\textit{Detect-then-reconstruct with a motion prior}} \\
Dyn-HaMR~\cite{S-yu2025dynhamr} & 148.41 & 11.31 & 30.22 & 6.92 & 10.25 & 106.99 & 11.55 & 47.75 & 9.67 & 12.58 \\
OmniHands~\cite{S-lin2026omnihands} & 64.96 & 15.00 & 19.64 & 7.50 & 30.63 & 68.93 & 12.73 & 24.93 & 7.50 & 28.17 \\
HaWoR~\cite{S-zhang2025hawor} & 36.69 & 11.44 & 17.80 & 7.43 & 21.54 & 42.87 & 12.24 & 26.77 & 8.41 & 26.26 \\
HandFlow~\cite{S-xu2026handflow} & 26.88 & 10.37 & 16.03 & 6.81 & 13.18 & 50.92 & 9.64 & 22.70 & 8.02 & 14.95 \\
MINT~\cite{S-zhu2026mint} & 48.73 & 20.33 & 17.99 & 8.73 & 9.10 & 49.32 & 14.06 & 26.36 & 7.43 & 8.02 \\
PAD-Hand~\cite{S-ismayilzada2026padhand} & 48.20 & 16.24 & 15.18 & 6.85 & 10.70 & 41.69 & 15.31 & 23.44 & 7.98 & 9.90 \\
InfiniHand~\cite{S-ren2026infinihand} & 36.38 & 6.93 & 13.38 & 5.47 & 10.15 & 50.13 & 8.21 & 21.84 & 8.85 & 15.62 \\
\midrule
\multicolumn{11}{@{}l}{\textit{Video diffusion with depth regression}} \\
ViDiHand~\cite{S-wang2026vidihand} & 32.63 & 8.82 & 17.14 & 7.95 & 2.67 & 90.47 & 10.63 & 28.33 & 6.89 & 3.53 \\
ACE-Ego-Hand~\cite{S-liu2026aceegohand} & 20.33 & 6.06 & 9.45 & 5.07 & 1.35 & 79.15 & 23.19 & 19.55 & 7.48 & 1.60 \\
\midrule
\multicolumn{11}{@{}l}{\textit{Video diffusion with hand geometry}} \\
\textbf{\ours (Ours)} & \best{17.42} & \best{4.22} & \best{8.88} & \best{4.73} & \best{1.23} & \best{28.09} & \best{5.99} & \best{17.82} & \best{6.80} & \best{1.46} \\
\bottomrule
\end{tabular}
 }
\end{table}

\subsection{Hands Detected by All Methods, Baselines with Their Officially Released Weights}
\label{S-sec:supp_common_official}
\Cref{S-tab:supp_common_tp_official} repeats the comparison of \cref{S-sec:supp_common} with the officially released weights of
\cref{S-tab:supp_official}. With these weights, all 19 methods detect 699{,}221 of the 1{,}446{,}795 ground-truth hand-frames, or 48\%.
\ours achieves superior results. It has the lowest MPJPE-abs, EPE2D, MPJPE and Jitter in both settings and the lowest PA in-domain,
and only the in-the-wild PA of ViDiHand$^\ast$ is lower, 6.60 against 6.65\,mm. On these hands, \ours reaches an MPJPE-abs of
24.08\,mm in the wild, against 25.15\,mm for EgoHandICL and 31.95\,mm for WiLoR.

\begin{table}[!tbp]
\centering
\caption{\textbf{Comparison on hands detected by all methods, officially released weights.} The setting of
\cref{S-tab:supp_official}, with $^\ast$ marking the methods trained on the same data as \ours, and the format of
\cref{S-tab:supp_common_tp}. \ours achieves superior results and has the lowest value in every column except PA in the wild.}
\label{S-tab:supp_common_tp_official}
\setlength{\tabcolsep}{3.2pt}
\adjustbox{max width=\textwidth,max totalheight=0.37\textheight}{%
\begin{tabular}{@{}lcccccccccc@{}}
\toprule
& \multicolumn{5}{c}{In-domain: mean of 6 datasets} & \multicolumn{5}{c}{In-the-wild: \genhuman} \\
\cmidrule(lr){2-6}\cmidrule(lr){7-11}
& \multicolumn{2}{c}{Absolute pose} & \multicolumn{2}{c}{Relative pose} & \multicolumn{1}{c}{Temporal} & \multicolumn{2}{c}{Absolute pose} & \multicolumn{2}{c}{Relative pose} & \multicolumn{1}{c}{Temporal} \\
\cmidrule(lr){2-3}\cmidrule(lr){4-5}\cmidrule(lr){6-6}\cmidrule(lr){7-8}\cmidrule(lr){9-10}\cmidrule(lr){11-11}
Method & MPJPE-abs$\downarrow$ & EPE2D$\downarrow$ & MPJPE$\downarrow$ & PA$\downarrow$ & Jitter$\downarrow$ & MPJPE-abs$\downarrow$ & EPE2D$\downarrow$ & MPJPE$\downarrow$ & PA$\downarrow$ & Jitter$\downarrow$ \\
\midrule
\multicolumn{11}{@{}l}{\textit{Detect-then-reconstruct, single frame}} \\
Hamba~\cite{S-dong2024hamba} & 44.04 & 9.86 & 32.02 & 7.48 & 11.26 & 36.56 & 10.30 & 41.81 & 7.66 & 10.82 \\
WildHands~\cite{S-prakash2024wildhands} & 73.36 & 34.50 & 29.73 & 10.40 & 14.58 & 105.83 & 69.75 & 30.53 & 9.40 & 15.93 \\
InterWild~\cite{S-moon2023interwild} & 60.45 & 15.34 & 35.13 & 9.33 & 30.85 & 50.36 & 13.62 & 43.04 & 8.24 & 23.47 \\
AnyHand~\cite{S-si2026anyhand} & 41.67 & 9.12 & 18.56 & 6.61 & 11.14 & 33.61 & 9.47 & 35.35 & 7.70 & 8.56 \\
HaMeR~\cite{S-pavlakos2024hamer} & 43.72 & 9.99 & 32.19 & 7.63 & 12.27 & 36.55 & 10.52 & 41.81 & 7.79 & 10.35 \\
EgoForce~\cite{S-millerdurai2026egoforce} & 46.58 & 18.46 & 24.12 & 7.80 & 13.98 & 45.49 & 23.29 & 29.25 & 7.77 & 18.54 \\
HandDGP~\cite{S-valassakis2024handdgp} & 65.05 & 13.59 & 35.24 & 10.28 & 17.85 & 52.33 & 12.16 & 40.34 & 8.86 & 14.79 \\
EgoHandICL$^\ast$~\cite{S-xie2026egohandicl} & 28.36 & 6.25 & 11.07 & 5.49 & 10.37 & 25.15 & 7.36 & 21.54 & 7.23 & 8.82 \\
WiLoR~\cite{S-potamias2025wilor} & 33.53 & 7.92 & 19.85 & 6.64 & 10.34 & 31.95 & 9.50 & 37.72 & 7.56 & 8.27 \\
\midrule
\multicolumn{11}{@{}l}{\textit{Detect-then-reconstruct with a motion prior}} \\
Dyn-HaMR~\cite{S-yu2025dynhamr} & 167.06 & 8.56 & 19.21 & 7.65 & 11.43 & 73.47 & 8.30 & 19.94 & 8.41 & 9.67 \\
OmniHands~\cite{S-lin2026omnihands} & 61.09 & 15.28 & 30.29 & 8.36 & 28.40 & 51.42 & 13.79 & 37.92 & 8.22 & 19.93 \\
HaWoR~\cite{S-zhang2025hawor} & 73.23 & 19.70 & 26.94 & 8.76 & 23.36 & 88.41 & 20.37 & 28.59 & 9.62 & 24.04 \\
HandFlow~\cite{S-xu2026handflow} & 82.22 & 14.21 & 23.07 & 7.98 & 4.14 & 111.30 & 10.21 & 18.63 & 8.41 & 4.52 \\
MINT~\cite{S-zhu2026mint} & 60.03 & 16.28 & 22.38 & 7.93 & 9.38 & 44.94 & 11.28 & 17.71 & 6.87 & 7.65 \\
PAD-Hand~\cite{S-ismayilzada2026padhand} & 51.70 & 17.80 & 25.11 & 8.82 & 10.28 & 44.38 & 16.70 & 41.48 & 7.84 & 7.83 \\
InfiniHand~\cite{S-ren2026infinihand} & 46.15 & 9.39 & 17.65 & 7.64 & 14.25 & 89.43 & 9.58 & 18.40 & 7.11 & 16.11 \\
\midrule
\multicolumn{11}{@{}l}{\textit{Video diffusion with depth regression}} \\
ViDiHand$^\ast$~\cite{S-wang2026vidihand} & 32.45 & 8.82 & 17.10 & 7.93 & 2.64 & 99.33 & 11.48 & 26.02 & \best{6.60} & 3.18 \\
ACE-Ego-Hand$^\ast$~\cite{S-liu2026aceegohand} & 20.32 & 6.03 & 9.43 & 5.05 & 1.34 & 80.96 & 22.59 & 18.70 & 7.38 & 1.43 \\
\midrule
\multicolumn{11}{@{}l}{\textit{Video diffusion with hand geometry}} \\
\textbf{\ours (Ours)}$^\ast$ & \best{17.34} & \best{4.20} & \best{8.87} & \best{4.71} & \best{1.22} & \best{24.08} & \best{6.32} & \best{17.12} & 6.65 & \best{1.29} \\
\bottomrule
\end{tabular}
 }
\end{table}

\FloatBarrier
\section{More Qualitative Results}
\label{S-sec:supp_qual}
\begingroup
\setlength{\intextsep}{4pt}
\captionsetup{skip=3pt,font=footnotesize}
\newcommand{\qualw}{0.94}   %
The figures compare all methods of Tab.~\ref{tab:main}. Long-range motion is followed by single-frame examples from Gen-HumanEgo, Ego4D~\cite{S-grauman2022ego4d}, EgoDex~\cite{S-hoque2026egodex} and Xperience~\cite{S-ropedia2026xperience}, in that order, and then in-domain examples. \Cref{S-sec:supp_ruler} closes the section with a phone video of a hand-held ruler, whose known length serves as 3D ground truth. Cases whose source clip appears in the main paper are placed first within each dataset when present. ViDiHand2 is marked by a green border.

\paragraph{Frame selection.}
These are selected qualitative examples, checked visually and deduplicated by source recording within each single-frame dataset. The quantitative tables report aggregate performance.

\newcommand{\groupqual}[3][0pt]{%
  \par\begin{figure}[H]\centering
  \begingroup
  \let\qualoriginalincludegraphics\includegraphics
  \renewcommand{\includegraphics}[2][]{\qualoriginalincludegraphics[width=\qualw\textwidth]{##2}}%
  #2\par
  \endgroup
  #3
  \end{figure}
}
\newcommand{\boundaryqual}[5]{%
  \par\begin{figure}[H]
  \begingroup
  \captionsetup{skip=1pt}
  \let\qualoriginalincludegraphics\includegraphics
  \renewcommand{\includegraphics}[2][]{\qualoriginalincludegraphics[width=\qualw\textwidth]{##2}}%
  {\centering #1\par}
  #2
  \vspace{-6pt}
  {\raggedright #3\par}
  \vspace{2pt}
  {\centering #4\par}
  #5
  \endgroup
  \end{figure}
}
\subsection{Long-Range Hand Motion}
\label{S-sec:supp_qual_long}
The six windows come from six recordings. Each cell shows the final hands and wrist paths in a three-quarter 3D view, blue for the left hand and orange for the right. Hands in light colors with a dark outline show the ground truth.
\groupqual{
\includegraphics[width=\textwidth]{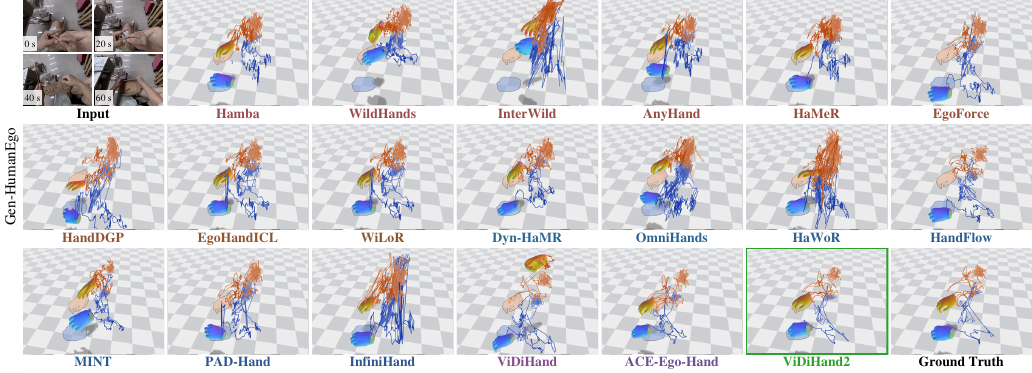}\par\vspace{8pt}
\includegraphics[width=\textwidth]{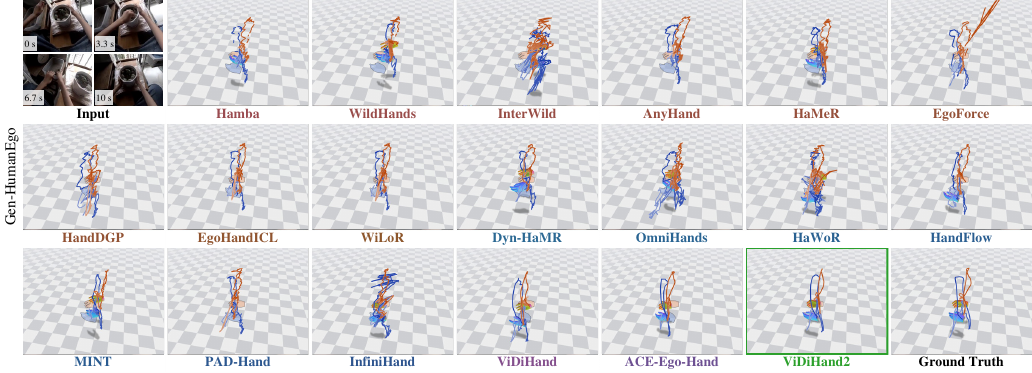}
}{
\caption{\textbf{Qualitative comparison on Gen-HumanEgo.} \textbf{Top:} Both hands work on a woven basket on the lap. WildHands brings the two hands too close together, while InterWild produces long outlying wrist paths. EgoForce and HandFlow fail to detect the current right hand. HaWoR shows a large trajectory excursion, InfiniHand produces dense long outlying paths, and ViDiHand shifts the right hand far from the ground truth. ViDiHand2 keeps both hands and wrist paths close to the ground truth. \textbf{Bottom:} Both hands handle small objects over a pot near a window. Hamba overlaps the current hands, HandDGP and PAD-Hand send the right-wrist path far downward, while InterWild produces dense back-and-forth wrist excursions. EgoForce has a long right-wrist outlier, and InfiniHand produces wide lateral left-wrist excursions. HaWoR also shows large trajectory excursions. ViDiHand2 keeps both hands and the trajectory shape closer to the ground truth.}
\label{S-fig:supp_long0}
}

\groupqual{
\includegraphics[width=\textwidth]{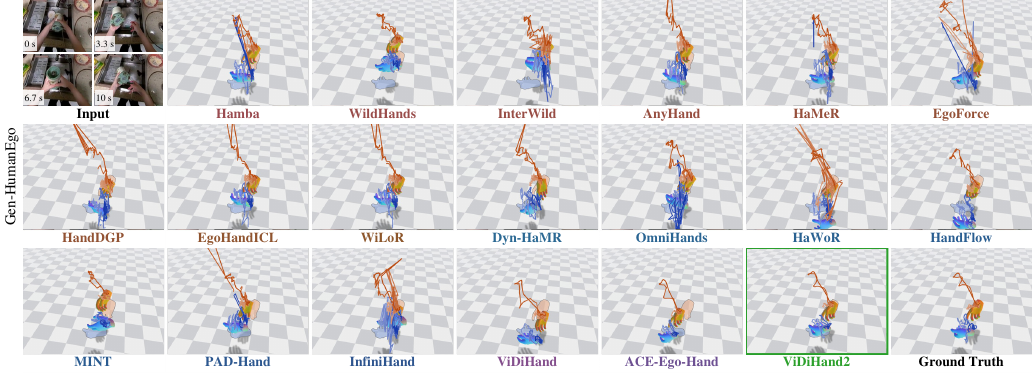}\par\vspace{8pt}
\includegraphics[width=\textwidth]{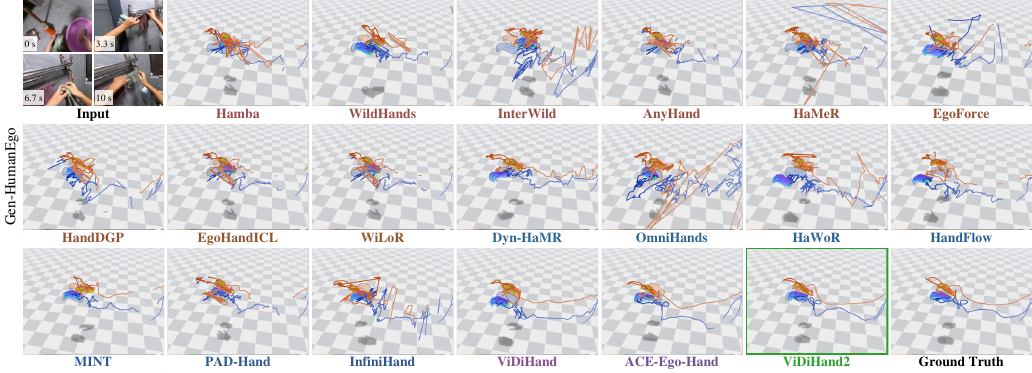}\par\vspace{8pt}
\includegraphics[width=\textwidth]{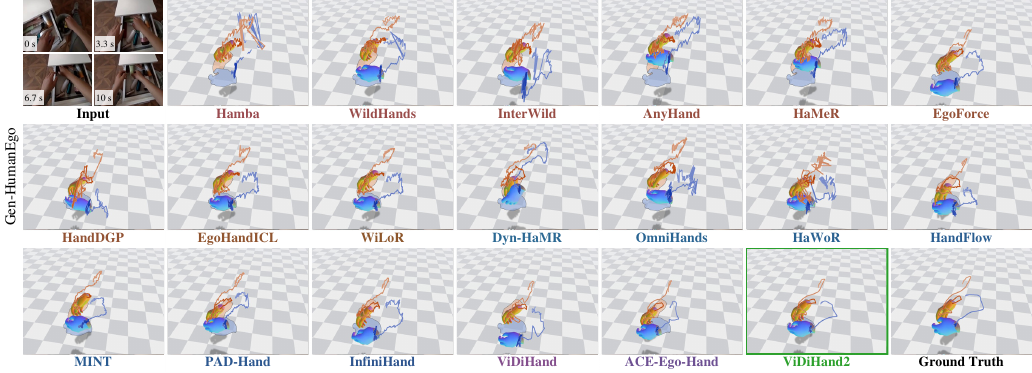}
}{
\caption{\textbf{Qualitative comparison on Gen-HumanEgo.} \textbf{Top:} Both hands wash a blender cup over a kitchen sink. Hamba, AnyHand and EgoForce produce long wrist-path outliers. HandDGP, EgoHandICL, WiLoR, PAD-Hand and Dyn-HaMR extend the right-hand path far above the ground-truth motion. HaWoR adds repeated large excursions around the hands. ViDiHand2 follows the compact ground-truth wrist paths and the current two-hand arrangement. \textbf{Middle:} Both hands wring cloths over a purple basin and hang them on a metal towel rail. InterWild and EgoForce produce long outlying wrist paths. HaMeR and OmniHands add large excursions across the scene, while HandDGP produces long left-wrist outliers and Dyn-HaMR produces jittery wrist paths. HaWoR separates the current hands too far. ViDiHand2 places both hands close to the ground truth with fewer large path deviations. \textbf{Bottom:} Both hands reach into an open drawer. Hamba omits the current left hand and produces long wrist-path outliers. WildHands and InterWild shift the left hand toward the center, while AnyHand brings both hands too close together. Dyn-HaMR misorients the left hand, and HaWoR produces jittery wrist paths. ViDiHand2 follows the ground-truth hand arrangement and the more compact motion paths.}
\label{S-fig:supp_long}
}

{\renewcommand{\qualw}{0.90}%
\boundaryqual{
\includegraphics[width=\textwidth]{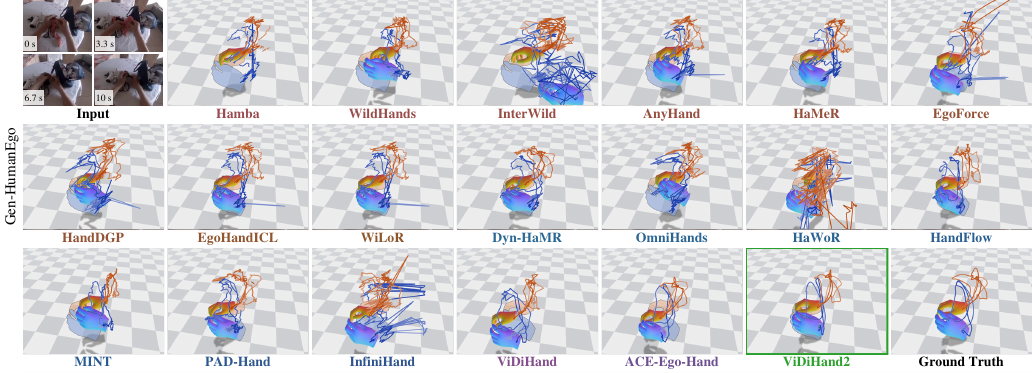}
}{
\caption{\textbf{Qualitative comparison on Gen-HumanEgo.} Both hands sort small items beside a bag on a round table. Hamba omits the current left hand. InterWild shifts the left-hand trajectory and the current left hand far from the ground truth, while EgoForce adds long outliers for both wrists. HaWoR shows large repeated motion excursions. PAD-Hand spreads the right-wrist path laterally, and InfiniHand sends the left path far across the scene. ViDiHand2 keeps the hands and their motion paths closer to the ground-truth region.}
\label{S-fig:supp_long1}

}{
\subsection{In-the-Wild Frames with Ground Truth}
\label{S-sec:supp_qual_wild}
The upper-left inset shows the hands and wrist paths in 3D, with the ground truth in light colors. The in-the-wild cases of
Fig.~\ref{fig:qual_nogt} of the main paper are in \cref{S-sec:supp_qual_nogt}.
}{
\includegraphics[width=\textwidth]{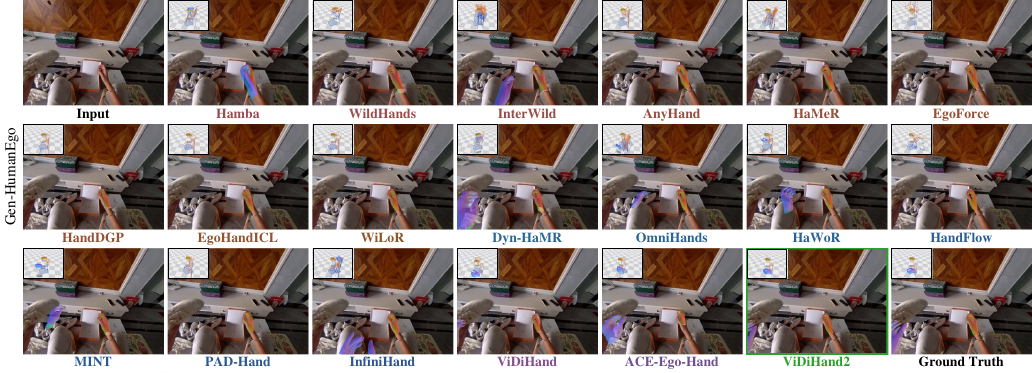}\par\vspace{4pt}
\includegraphics[width=\textwidth]{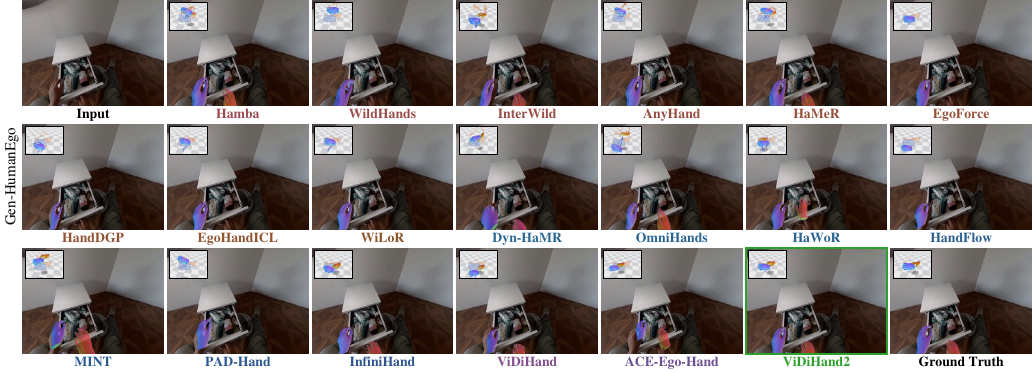}
}{
\caption{\textbf{Top:} \textbf{Qualitative comparison on Gen-HumanEgo.} The left hand lifts a sneaker at the image edge while the right hand holds an open shoebox. WildHands, AnyHand, HaMeR, EgoForce, HandDGP, EgoHandICL, WiLoR, PAD-Hand and HandFlow fail to detect this partly visible left hand, and Hamba assigns a left-hand mesh to the visible right hand. Dyn-HaMR inflates the left hand at the image edge, InfiniHand places a large left hand in the foreground, HaWoR places it by the paper, and ViDiHand and ACE-Ego-Hand misplace it. InterWild, MINT and OmniHands place the left hand in the middle of the image. ViDiHand2 recovers the partly visible hand at the image edge and places the right hand close to the ground truth. \textbf{Bottom:} \textbf{Qualitative comparison on Gen-HumanEgo.} Both hands reach into an open drawer, with the right hand in dim light. WildHands, AnyHand, EgoForce, HandDGP, EgoHandICL, WiLoR, HandFlow and PAD-Hand fail to detect the right hand, and MINT draws it far too large. HaWoR misplaces the right hand, ACE-Ego-Hand both hands and ViDiHand the left hand. ViDiHand2 places both hands close to the ground truth.}
\label{S-fig:distinct_gen_humanego_1h_01}
\label{S-fig:distinct_gen_humanego_1h_02}
\label{S-fig:distinct_gen_humanego_1h_03}
\label{S-fig:supp_main_drawer}
}
}
\groupqual{
\includegraphics[width=\textwidth]{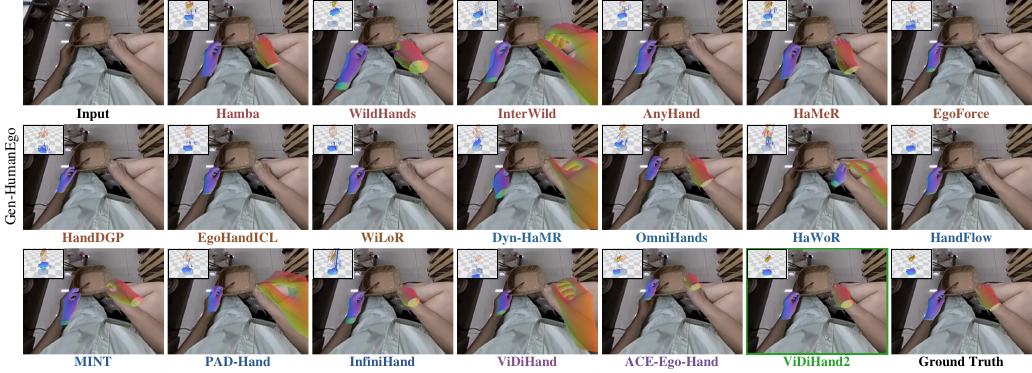}\par\vspace{8pt}
\includegraphics[width=\textwidth]{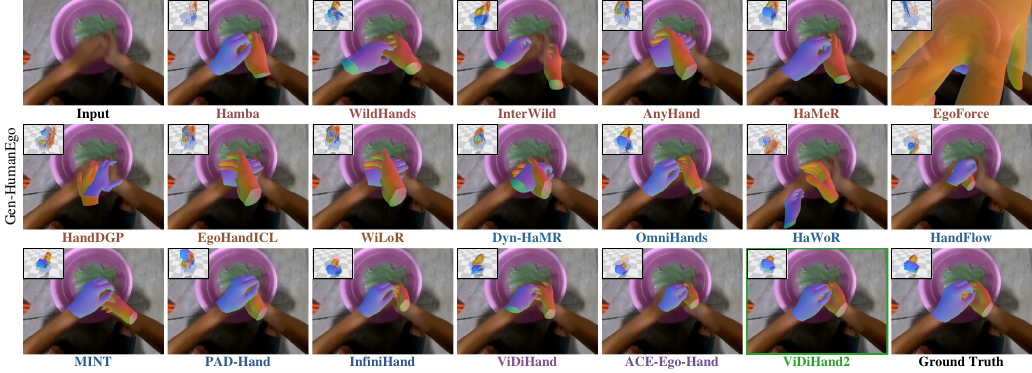}\par\vspace{8pt}
\includegraphics[width=\textwidth]{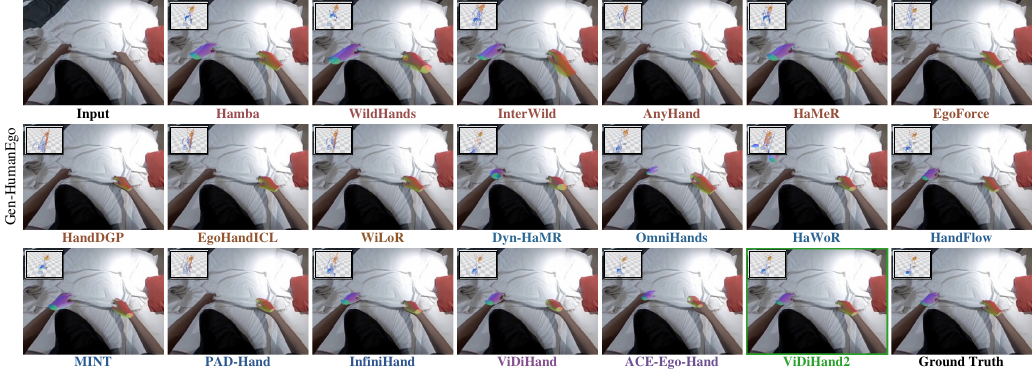}
}{
\caption{\textbf{Top:} \textbf{Qualitative comparison on Gen-HumanEgo.} Both hands hold a woven basket over the lap. AnyHand, EgoForce, HandDGP, EgoHandICL, WiLoR and HandFlow fail to detect the right hand. InterWild, PAD-Hand, Dyn-HaMR and ViDiHand draw the right hand far too large, and HaWoR also enlarges it. HaWoR puts the left hand on the right leg. ViDiHand2 keeps both hands near the ground-truth size and grip locations. \textbf{Middle:} \textbf{Qualitative comparison on Gen-HumanEgo.} Both hands handle a cloth in a purple bowl. EgoForce produces a mesh that covers most of the image. WildHands displaces the left palm across the bowl, AnyHand, EgoHandICL, WiLoR and Dyn-HaMR misplace the overlapping hands, and HandDGP swaps them. InterWild shifts the right hand down the forearm. HaWoR separates the left hand from the contact region. ViDiHand2 follows the ground-truth hand size and the contact location above the cloth. \textbf{Bottom:} \textbf{Qualitative comparison on Gen-HumanEgo.} Both hands pull the edge of a bedsheet. AnyHand, EgoForce, HandDGP, EgoHandICL, WiLoR and PAD-Hand fail to detect the left hand despite its visible grip, and HaWoR places it far above on the bedsheet. ViDiHand2 reconstructs both hands at the sheet edge, with placement close to the ground truth.}
\label{S-fig:distinct_gen_humanego_1h_04}
}

\groupqual{
\includegraphics[width=\textwidth]{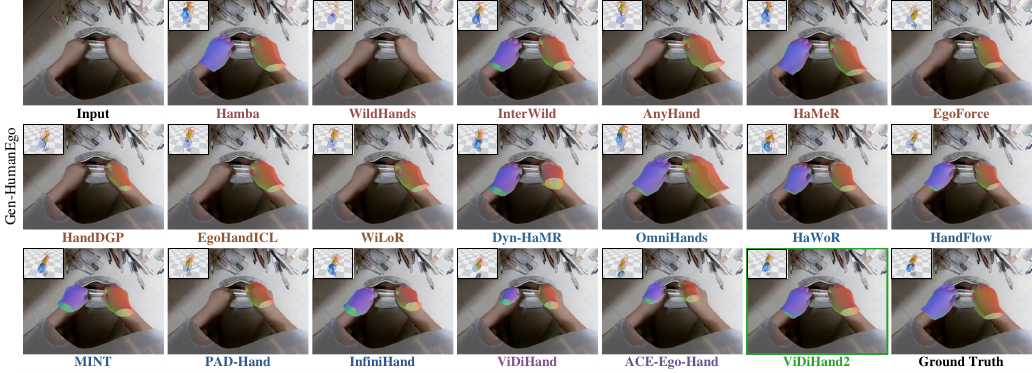}\par\vspace{8pt}
\includegraphics[width=\textwidth]{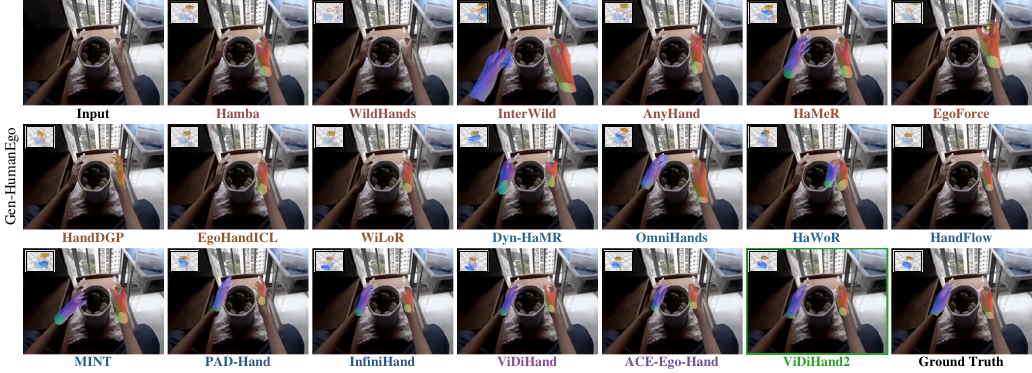}\par\vspace{8pt}
\includegraphics[width=\textwidth]{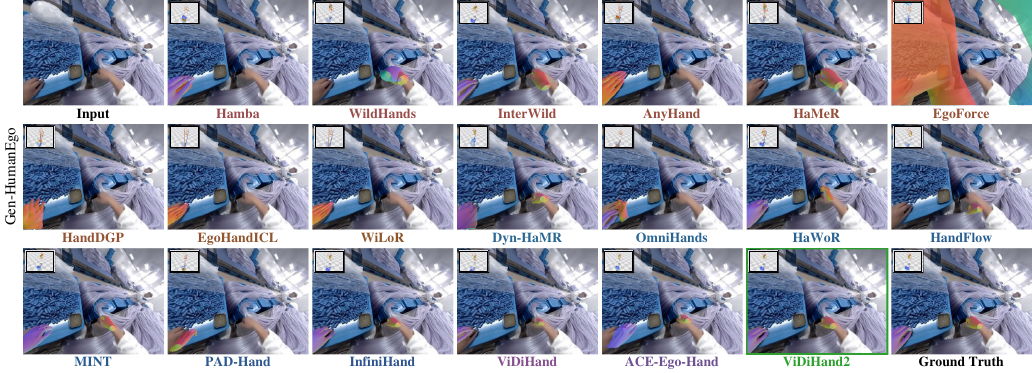}
}{
\caption{\textbf{Top:} \textbf{Qualitative comparison on Gen-HumanEgo.} Both hands grip a small pouch over the lap. WildHands shows no hand meshes, and AnyHand, EgoForce, HandDGP, EgoHandICL, WiLoR and PAD-Hand fail to detect the left hand. ViDiHand2 retains both hands and follows the ground-truth grip locations on either side of the pouch. \textbf{Middle:} \textbf{Qualitative comparison on Gen-HumanEgo.} Both hands hold a pot near a window. WildHands shows no hand meshes, and Hamba, AnyHand, EgoForce, HandDGP, EgoHandICL, WiLoR and HandFlow fail to detect the left hand. EgoForce also enlarges the right hand and shifts it upward, and HaWoR moves the left hand beside the right. InterWild reconstructs it too large and extends it down the forearm. ViDiHand2 retains both hands at the pot rim, near their ground-truth placement. \textbf{Bottom:} \textbf{Qualitative comparison on Gen-HumanEgo.} The left hand rests on a tray of blue tubes while the right hand gathers a bundle of lavender strands. EgoForce produces a mesh spanning much of the image. Hamba, AnyHand, HandDGP, EgoHandICL, WiLoR, PAD-Hand and OmniHands place a right-hand mesh on the left hand and leave the right hand empty, while WildHands, HaMeR and HandFlow fail to detect the left hand. ViDiHand2 retains both hands at the tray and strands, with size and placement close to the ground truth.}
\label{S-fig:distinct_gen_humanego_1h_05}
\label{S-fig:distinct_gen_humanego_1h_06}
}

\groupqual{
\includegraphics[width=\textwidth]{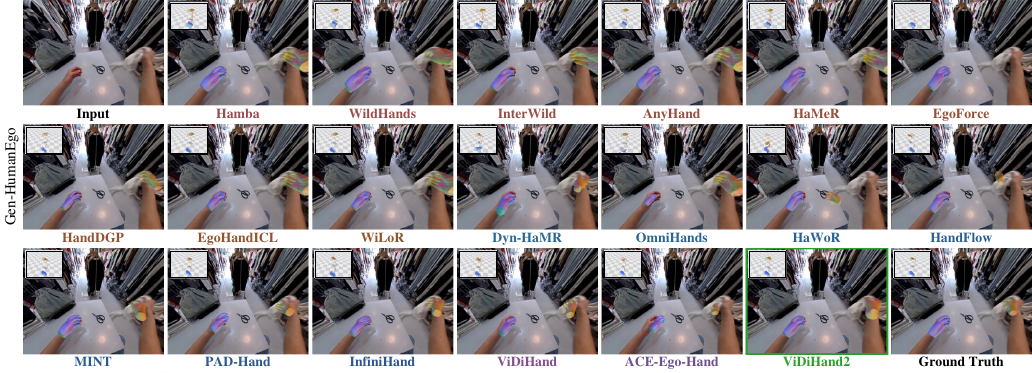}\par\vspace{8pt}
\includegraphics[width=\textwidth]{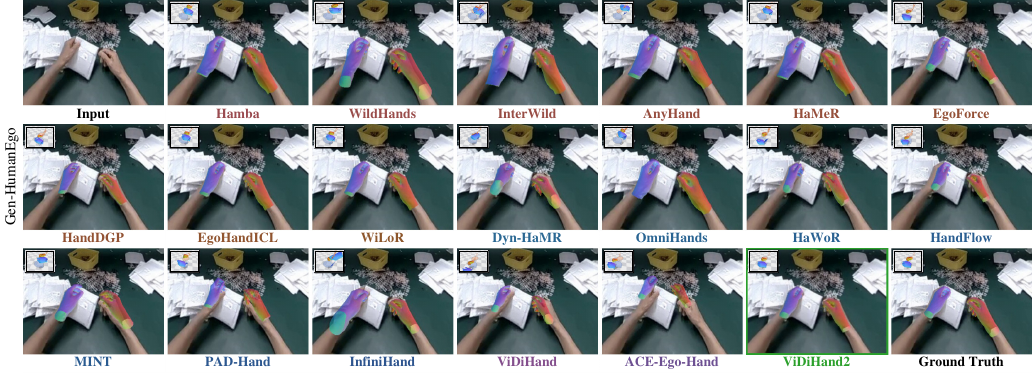}\par\vspace{8pt}
\includegraphics[width=\textwidth]{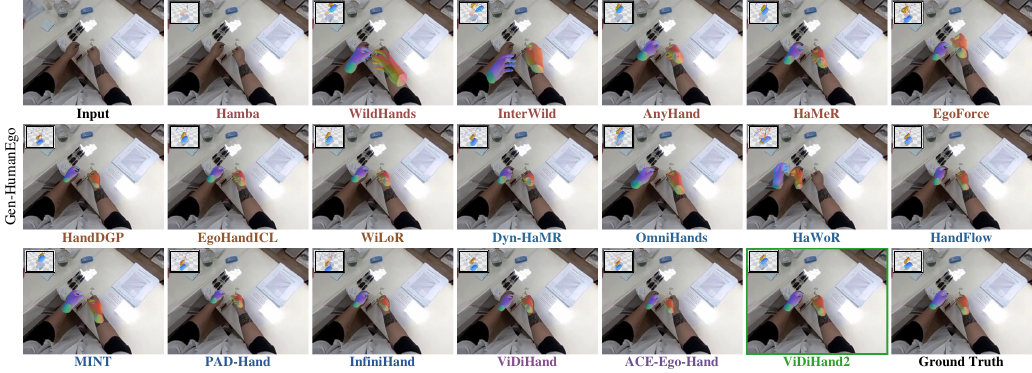}
}{
\caption{\textbf{Top:} \textbf{Qualitative comparison on Gen-HumanEgo.} The left hand reaches toward scissors on a white cutting table while the right hand holds white fabric, between racks of clothing. EgoForce omits the right-hand overlay. WildHands, InterWild, AnyHand, HaMeR, Hamba, EgoHandICL, WiLoR and OmniHands spread the right-hand fingers beyond the visible grip. HaWoR displaces the right hand onto the tabletop. ViDiHand2 follows both hands. \textbf{Middle:} \textbf{Qualitative comparison on Gen-HumanEgo.} Both hands pack small parts into white bags at a cluttered workbench. WildHands and AnyHand elongate both hand meshes down the forearms, and Hamba, InterWild and OmniHands elongate the left-hand mesh. InfiniHand makes the same error more strongly, shifting the palm away from the visible hand. ViDiHand2 keeps the left palm and the right-hand grip closer to the ground truth. \textbf{Bottom:} \textbf{Qualitative comparison on Gen-HumanEgo.} Both hands handle small test tubes beside papers on a laboratory bench. Hamba fails to detect both hands. WildHands and InterWild displace the left palm toward the foreground, while HaWoR shifts it sideways and misplaces the right hand. EgoForce raises the right-hand mesh above the visible grip. ViDiHand2 retains both hands near the ground-truth contact locations.}
\label{S-fig:distinct_gen_humanego_1h_07}
}

\boundaryqual{
\includegraphics[width=\textwidth]{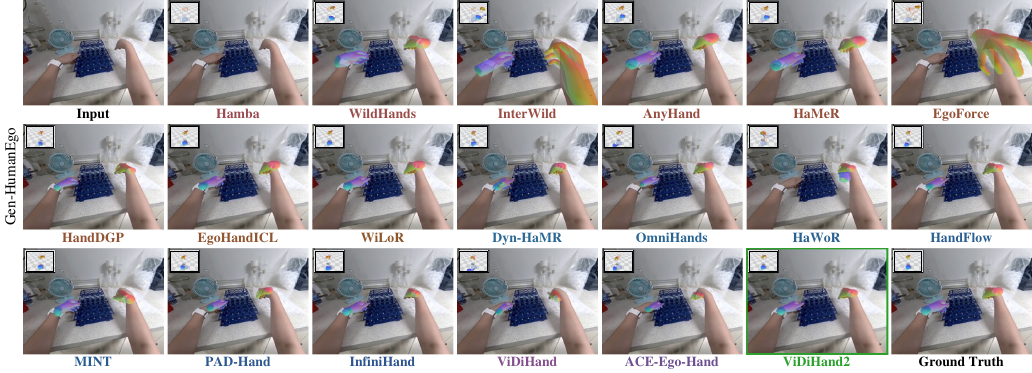}\par\vspace{8pt}
\includegraphics[width=\textwidth]{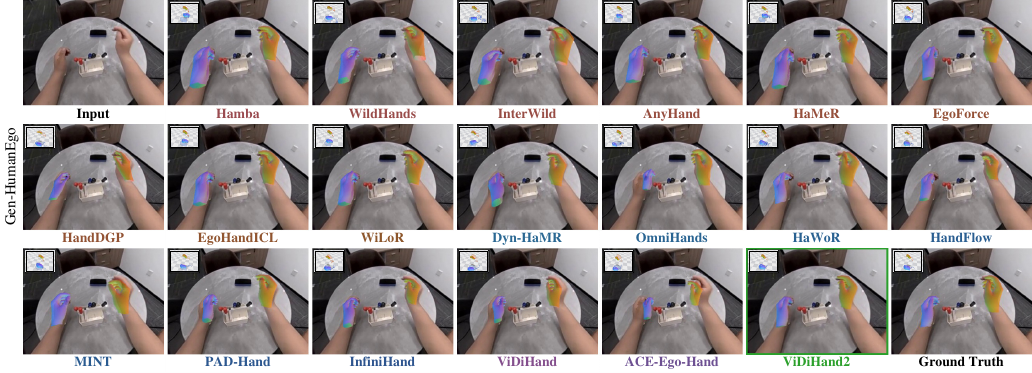}
}{
\caption{\textbf{Top:} \textbf{Qualitative comparison on Gen-HumanEgo.} The left hand touches a blue tube rack on a laboratory bench while the right hand is raised. Hamba fails to detect both hands, while HaWoR places both meshes on the right hand. EgoForce omits the left hand and, like InterWild, greatly enlarges the right hand. ViDiHand2 retains both hands with size and placement close to the ground truth. \textbf{Bottom:} \textbf{Qualitative comparison on Gen-HumanEgo.} Both hands handle small objects above a round table. Hamba, AnyHand, HaMeR, WildHands and WiLoR make the left palm too broad and extend it down the wrist. ViDiHand2 better preserves the ground-truth palm size and the two manipulation locations.}

}{
\subsection{In-the-Wild Frames without Reliable Ground Truth}
\label{S-sec:supp_qual_nogt}
ViDiHand2 is highlighted with a green border. Without reliable ground truth, the overlays show image alignment only. No 3D inset is shown.
}{
\includegraphics[width=\textwidth]{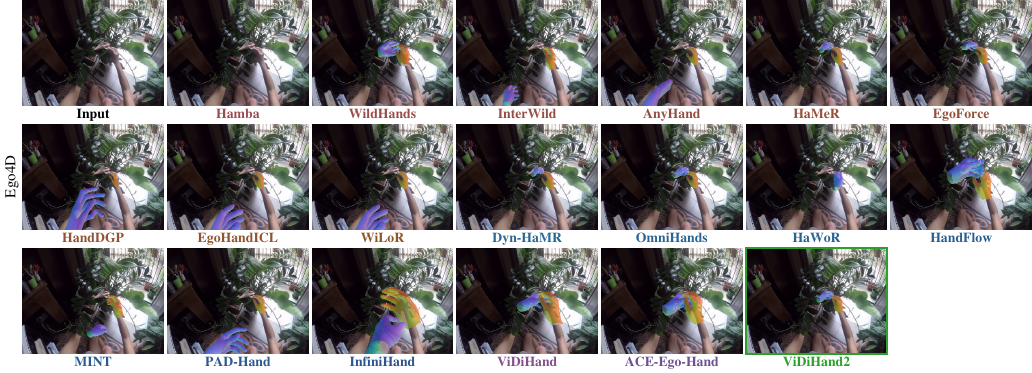}
}{
\caption{\textbf{Qualitative comparison on Ego4D.} Without reliable ground truth, the overlays show image alignment only. Both hands reach into a plant. Hamba shows no hand meshes. WiLoR, InterWild, AnyHand, HandDGP, EgoHandICL and PAD-Hand place the left hand on the leg, and MINT places it below the plant. HaWoR detects only one hand. InfiniHand draws both hands far too large, HandFlow enlarges both hands, and ViDiHand and ACE-Ego-Hand enlarge both hands. ViDiHand2 places both hands at the visible contact region among the leaves.}
\label{S-fig:distinct_ego4d_01}
\label{S-fig:distinct_ego4d_07}
}

\groupqual{
\includegraphics[width=\textwidth]{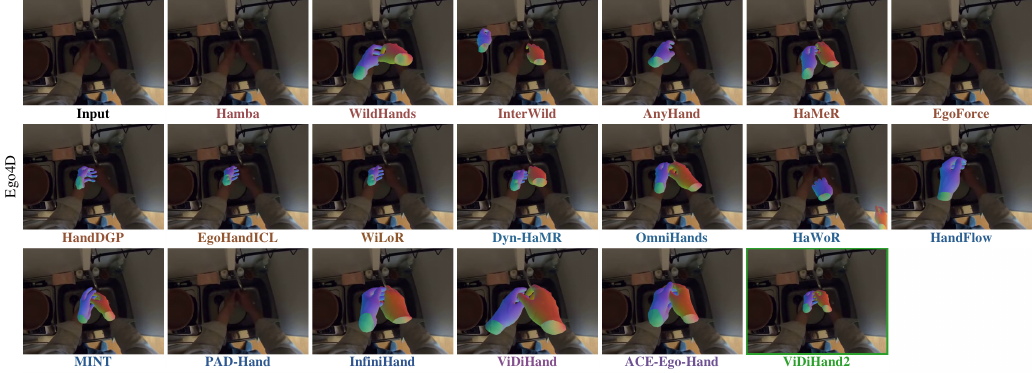}\par\vspace{8pt}
\includegraphics[width=\textwidth]{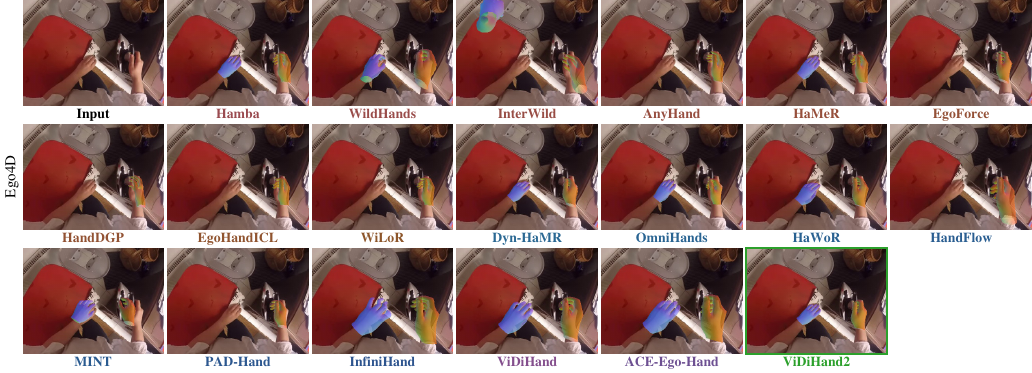}\par\vspace{8pt}
\includegraphics[width=\textwidth]{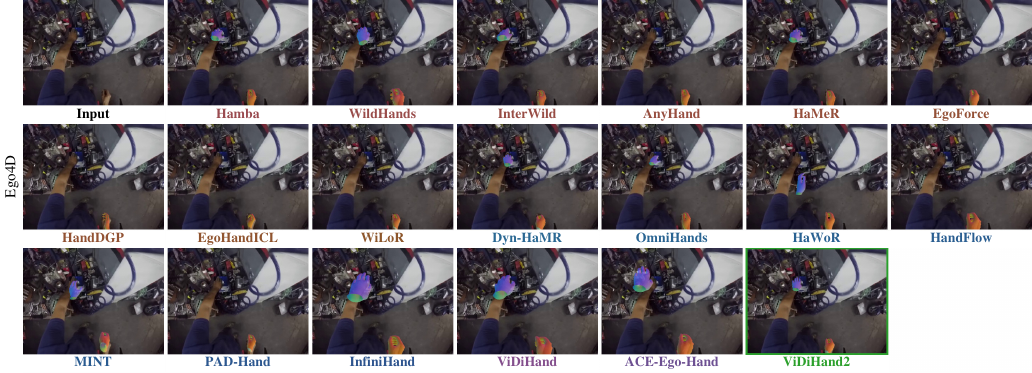}
}{
\caption{\textbf{Top:} \textbf{Qualitative comparison on Ego4D.} Without reliable ground truth, the overlays show image alignment only. Both hands wash an object in a sink. Hamba, EgoForce and PAD-Hand show no hand meshes, and AnyHand, HandDGP, EgoHandICL, WiLoR and HandFlow omit the right hand. InterWild shifts the left hand onto the counter, while HaWoR separates the hands from the washing region. InfiniHand, ViDiHand and ACE-Ego-Hand draw both hands too large. ViDiHand2 keeps both meshes at the visible hand locations and image scale. \textbf{Middle:} \textbf{Qualitative comparison on Ego4D.} The left hand holds patterned fabric beside an ironing board while the right hand holds an iron. AnyHand, EgoForce, HandDGP, EgoHandICL, WiLoR, PAD-Hand and HandFlow omit the left-hand overlay. InterWild places it far above the visible hand. InfiniHand, ViDiHand and ACE-Ego-Hand greatly enlarge both hands. ViDiHand2 follows the visible left-hand grip and the right hand holding the iron. \textbf{Bottom:} \textbf{Qualitative comparison on Ego4D.} A hand reaches into a cluttered pile of tools and engine parts beside a scooter. AnyHand, EgoForce, HandDGP, EgoHandICL, WiLoR, PAD-Hand and HandFlow omit the left-hand overlay. HaWoR shifts the left hand downward and rotates its fingers away from the visible reach. InfiniHand, ViDiHand and ACE-Ego-Hand enlarge the left hand beyond its image outline. ViDiHand2 follows the extended left hand and the partly visible right hand.}
}

\groupqual{
\includegraphics[width=\textwidth]{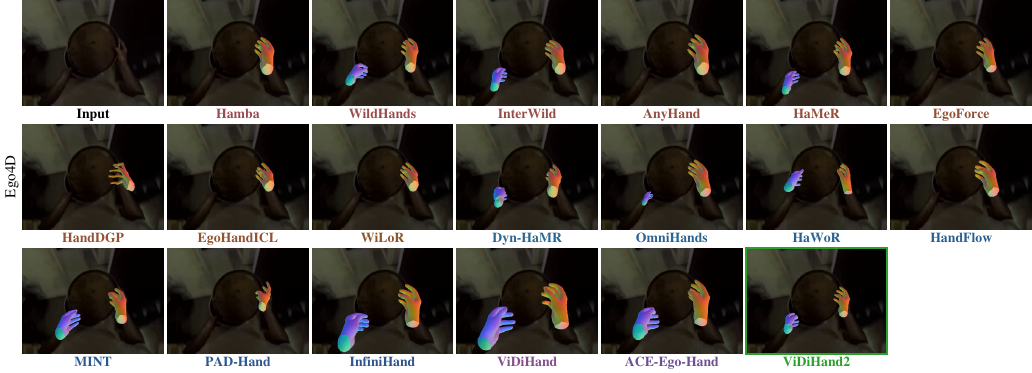}\par\vspace{8pt}
\includegraphics[width=\textwidth]{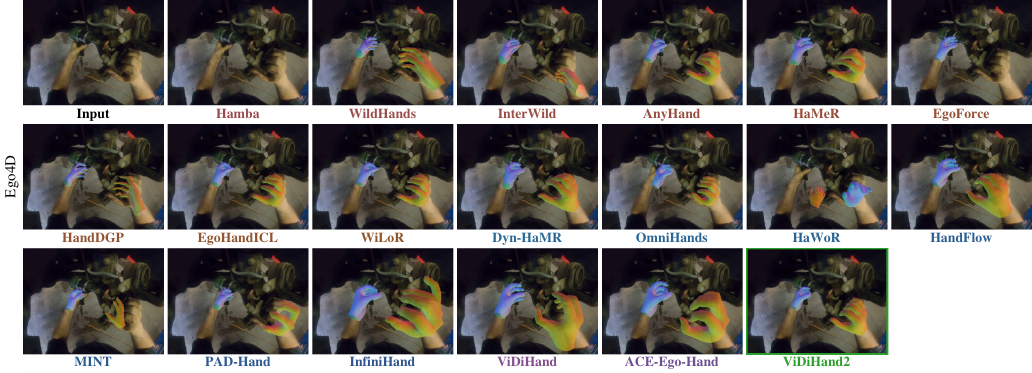}\par\vspace{8pt}
\includegraphics[width=\textwidth]{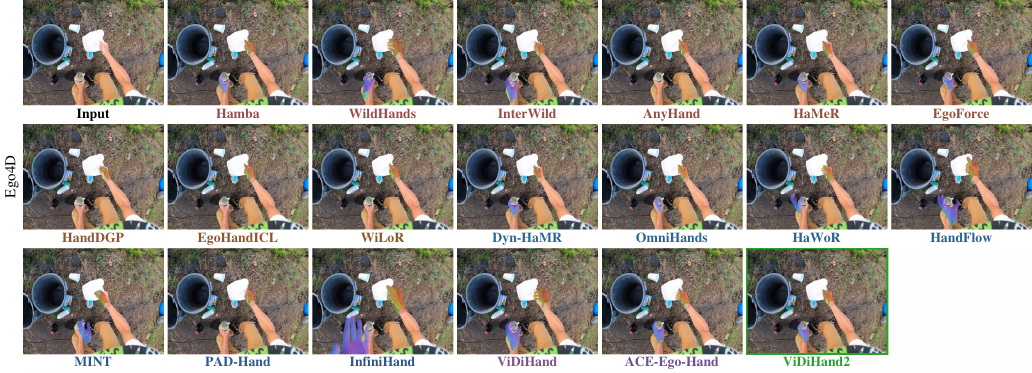}
}{
\caption{\textbf{Top:} \textbf{Qualitative comparison on Ego4D.} Without reliable ground truth, the overlays show image alignment only. Both hands carry a round tray through a dim doorway. Hamba, AnyHand, EgoForce, HandDGP, EgoHandICL, WiLoR, HandFlow and PAD-Hand omit the left-hand overlay. InfiniHand, ViDiHand and ACE-Ego-Hand enlarge the left hand beyond the visible grip. ViDiHand2 retains both hands at the tray rim, following the visible grip locations and hand scale. \textbf{Middle:} \textbf{Qualitative comparison on Ego4D.} The left hand holds a power tool against a mechanical assembly that the right hand steadies. Hamba shows no hand meshes, and EgoForce omits the right hand. HaWoR displaces and reverses the apparent hand locations. WildHands, InfiniHand and ACE-Ego-Hand enlarge the right hand, while ViDiHand extends it below the visible wrist. ViDiHand2 places both meshes over the hands beside the assembly. \textbf{Bottom:} \textbf{Qualitative comparison on Ego4D.} Outdoors, the left hand holds a cup while the right hand holds a white plastic container. AnyHand, HandDGP, EgoHandICL, WiLoR and PAD-Hand omit the left-hand overlay. InfiniHand places a large open hand below the visible cup grip, while HaWoR and HandFlow shift the left hand slightly away from the cup. ViDiHand2 follows both the cup-holding hand and the hand holding the container.}
}

\groupqual{
\includegraphics[width=\textwidth]{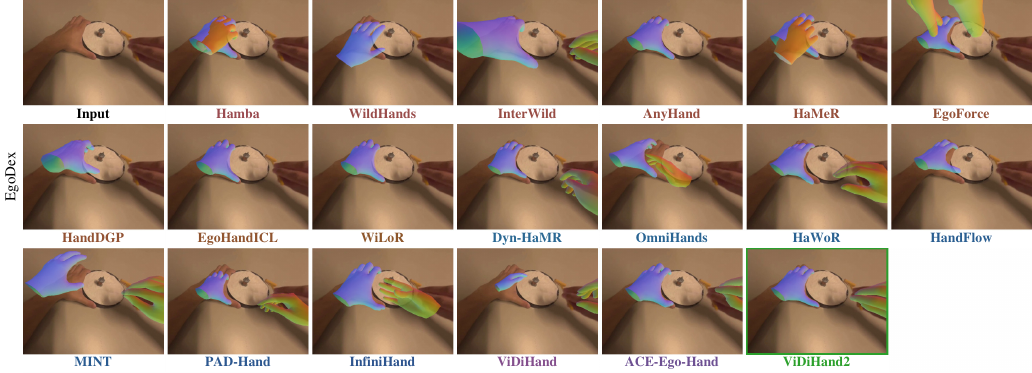}\par\vspace{8pt}
\includegraphics[width=\textwidth]{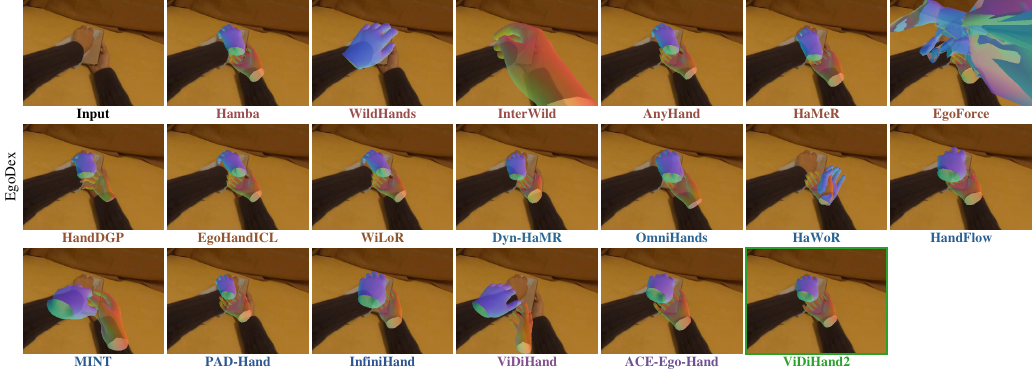}\par\vspace{8pt}
\includegraphics[width=\textwidth]{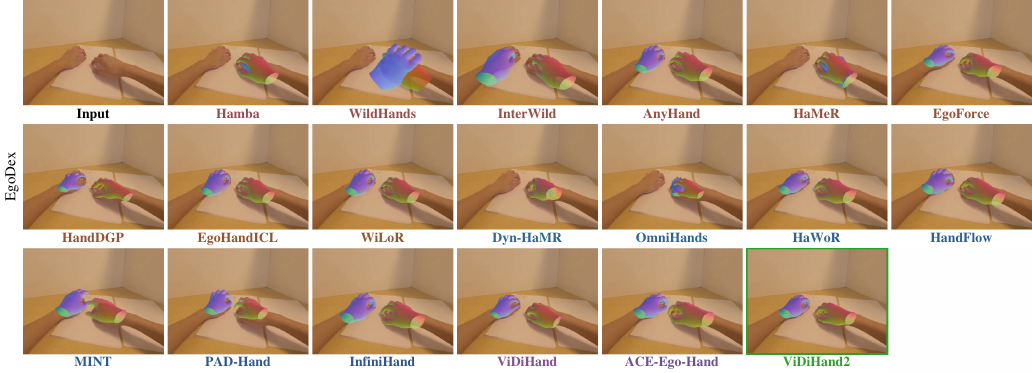}
}{
\caption{\textbf{Top:} \textbf{Qualitative comparison on EgoDex.} Without reliable ground truth, the overlays show image alignment only. The left hand holds a small sand tray while the right hand holds a rake. Hamba and HaMeR place both meshes over the left hand, and WildHands, AnyHand, HandDGP, EgoHandICL, WiLoR and HandFlow fail to detect the right hand. EgoForce adds two spurious arms from the top of the image. OmniHands and InfiniHand shift the right hand onto the tray, and HaWoR mis-poses it. InterWild makes the left hand much too large, and ViDiHand and ACE-Ego-Hand misplace or mis-pose a hand. ViDiHand2 follows the tray-holding hand and the visible rake grip. \textbf{Middle:} \textbf{Qualitative comparison on EgoDex.} On a yellow surface, the left hand wipes a flat case with a cloth while the right hand holds it. WildHands and MINT enlarge the left hand, and WildHands also fails to detect the right hand. InterWild places an oversized right hand across the wiping hand, while EgoForce produces fragmented meshes far beyond both hands. HaWoR overlaps both predictions on the supporting hand, and ViDiHand enlarges the left hand and extends the right wrist down the forearm. ViDiHand2 follows both the wiping hand and the supporting grip at the visible scale. \textbf{Bottom:} \textbf{Qualitative comparison on EgoDex.} Both hands rest on a stack of paper, the right hand holding a small transparent object. Hamba and HaMeR omit the left-hand overlay and place an extra mesh on the right hand, and InterWild enlarges the left hand. WildHands overlaps a large left-hand mesh with the right hand, while OmniHands also merges both predictions at the right grip. Dyn-HaMR also omits the left hand. ViDiHand2 places the left hand on the page and follows the right-hand grip on the object.}
}

\groupqual{
\includegraphics[width=\textwidth]{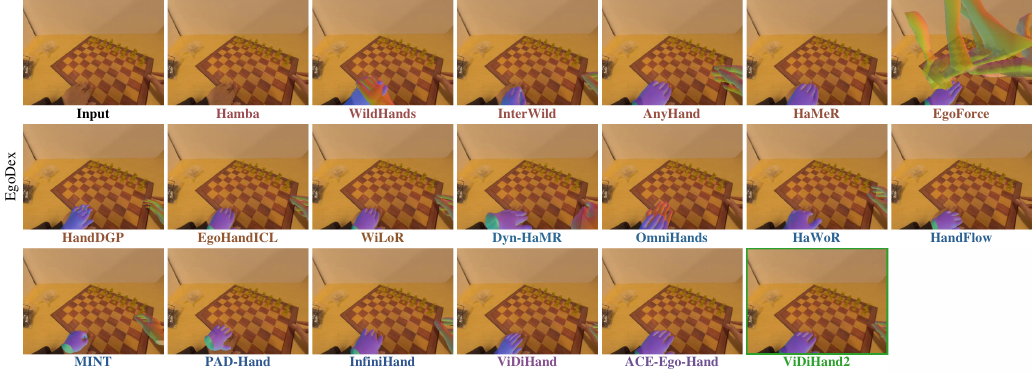}\par\vspace{8pt}
\includegraphics[width=\textwidth]{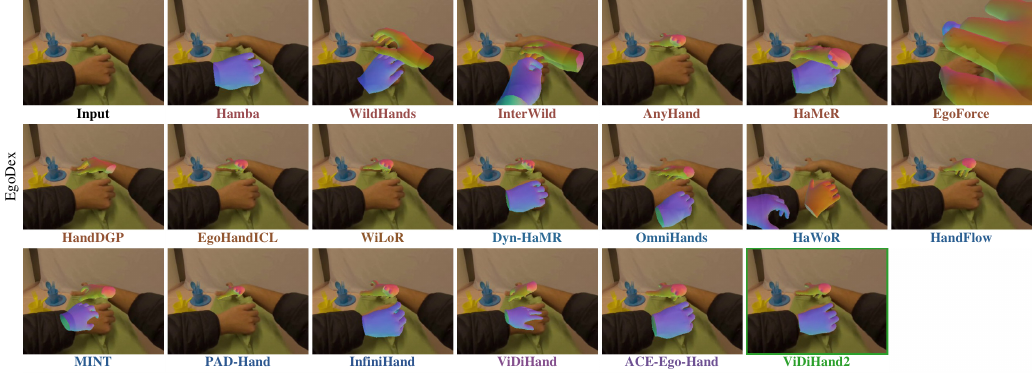}\par\vspace{8pt}
\includegraphics[width=\textwidth]{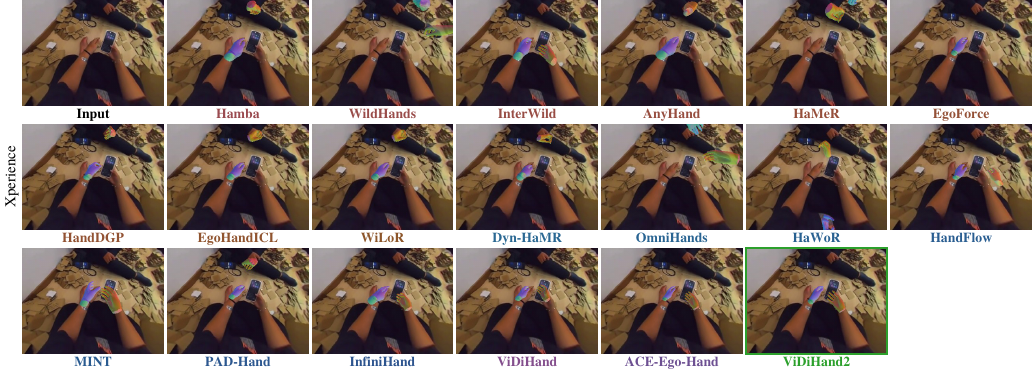}
}{
\caption{\textbf{Top:} \textbf{Qualitative comparison on EgoDex.} Without reliable ground truth, the overlays show image alignment only. Both hands hold a chessboard at its near edges. Hamba shows no hand meshes, and HaMeR, HandFlow, PAD-Hand and ACE-Ego-Hand fail to detect the right hand at the image edge. WildHands and OmniHands overlap both predictions near the left hand, while EgoForce produces an oversized mesh above the board. Dyn-HaMR shifts the left hand inward and rotates the right palm toward the camera. ViDiHand2 follows the two visible hands at the board edges. \textbf{Middle:} \textbf{Qualitative comparison on EgoDex.} Both hands work with green fabric on a table. Hamba omits the right-hand mesh, while AnyHand, HandDGP, EgoHandICL, WiLoR, PAD-Hand and HandFlow omit the left. EgoForce produces a large mesh covering the right side of the image. InterWild extends both wrists, and HaWoR separates and reverses the apparent hand locations. ViDiHand2 follows both visible hands on the fabric. \textbf{Bottom:} \textbf{Qualitative comparison on Xperience.} The left hand is held open next to a phone and the right hand holds the phone. Several detect-then-reconstruct baselines, including Hamba, HandDGP, WiLoR and PAD-Hand, fail to detect the right hand and place a right-hand mesh far above it, and EgoForce shows no right hand. HaMeR overlays neither visible hand and places its meshes far up the frame. HaWoR fails to detect both hands, putting one mesh on the floor above the phone and one on the leg. ViDiHand draws the right hand on the screen, and ACE-Ego-Hand spreads its fingers over the phone. ViDiHand2 aligns with both hands.}
\label{S-fig:supp_main_phone}
}

\groupqual{
\includegraphics[width=\textwidth]{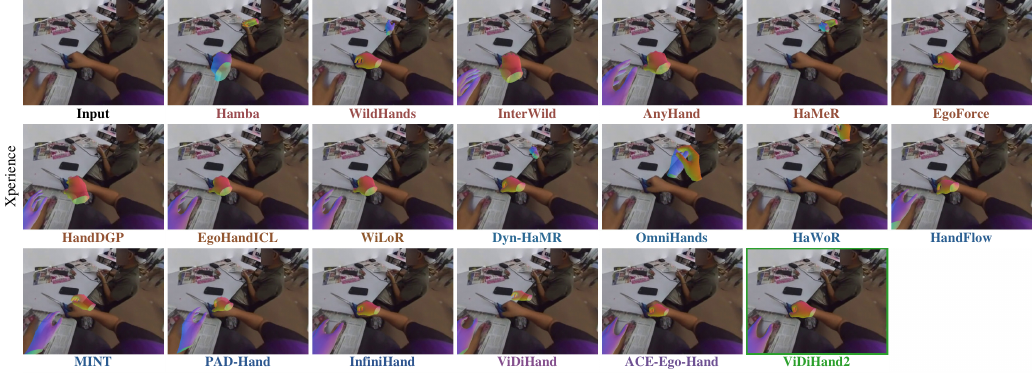}\par\vspace{8pt}
\includegraphics[width=\textwidth]{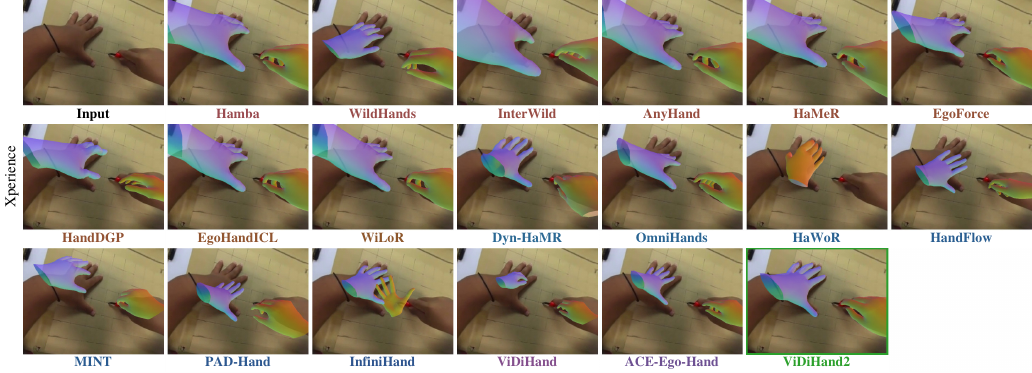}\par\vspace{8pt}
\includegraphics[width=\textwidth]{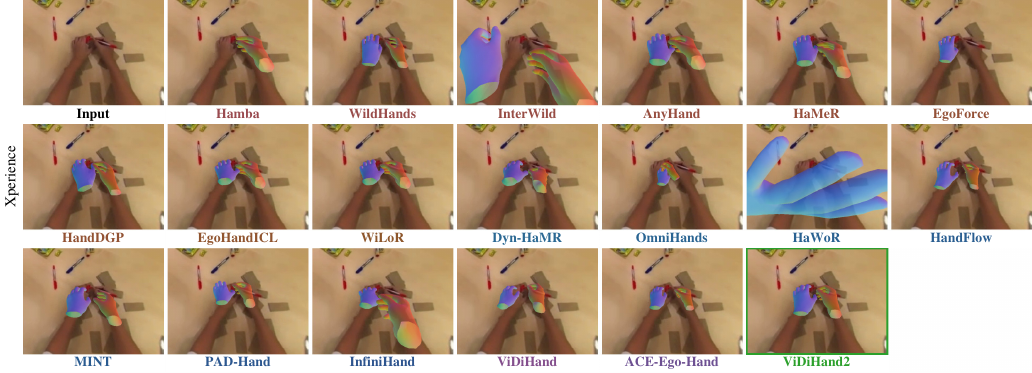}
}{
\caption{\textbf{Top:} \textbf{Qualitative comparison on Xperience.} Without reliable ground truth, the overlays show image alignment only. A person cuts paper with scissors at a shared table. Hamba, HaMeR and Dyn-HaMR place meshes on another person's hand, and Hamba also puts a left-hand mesh on the scissor-holding hand. WildHands places the left-hand mesh on the other person's hand, while OmniHands moves both meshes above the scissors. HaWoR places a hand near the other person's face, and ViDiHand shifts the right hand away from the scissors. ViDiHand2 follows the paper-holding hand and the scissor grip. \textbf{Middle:} \textbf{Qualitative comparison on Xperience.} A person marks a grid while holding the sheet flat. Most image-based baselines, including Hamba, InterWild, AnyHand, HaMeR, EgoForce, HandDGP, EgoHandICL, WiLoR and OmniHands, make the left hand much too large. HaWoR places a right-hand mesh over the visible left hand and omits the writing-hand overlay. InfiniHand overlaps an open right-hand mesh with the left hand, while ViDiHand makes the left hand too small. ViDiHand2 follows the left-hand spread and the right-hand pen grip. \textbf{Bottom:} \textbf{Qualitative comparison on Xperience.} A person marks small cards with a pen. Hamba omits the left-hand mesh, while EgoForce omits the right. InterWild enlarges and separates both hands, and HaWoR produces a large mesh covering the foreground. InfiniHand makes the right hand much too large. ViDiHand2 places both hands at the card and pen grip.}

}

\groupqual{
\includegraphics[width=\textwidth]{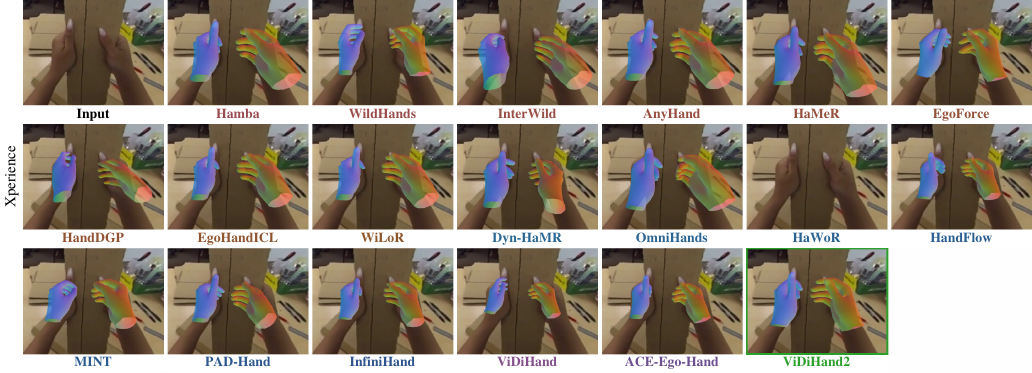}\par\vspace{8pt}
\includegraphics[width=\textwidth]{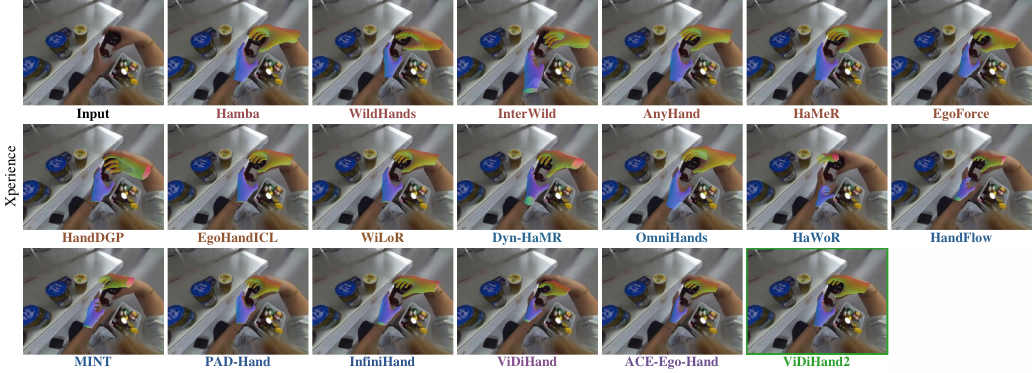}\par\vspace{8pt}
\includegraphics[width=\textwidth]{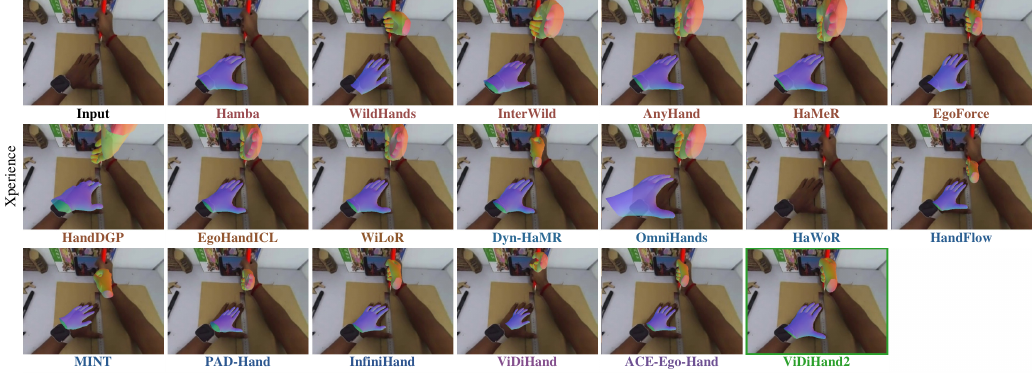}
}{
\caption{\textbf{Top:} \textbf{Qualitative comparison on Xperience.} Without reliable ground truth, the overlays show image alignment only. Both hands hold a vertical cardboard piece. HaWoR shows no hand meshes. Hamba, InterWild, OmniHands, AnyHand, HaMeR, EgoHandICL and WiLoR enlarge the right hand beyond the visible grip, while ViDiHand makes both hands too small. ViDiHand2 preserves both grips despite the long nails. \textbf{Middle:} \textbf{Qualitative comparison on Xperience.} Both hands twist a lid on a dark container. Several baselines, e.g., Hamba and AnyHand, extend the right hand beyond the visible wrist, and InterWild stretches the left hand down the forearm. HaWoR separates the left hand from the container, and ViDiHand makes the left grip too small. ViDiHand2 keeps both meshes around the container and follows the visible lid grip. \textbf{Bottom:} \textbf{Qualitative comparison on Xperience.} A person holds a ruler against cardboard with the left hand and a red tool with the right. Hamba omits the right-hand overlay, while HaWoR shows neither hand. HandDGP and OmniHands enlarge the left hand beyond its visible outline. InterWild, HaMeR, AnyHand, HandDGP, EgoHandICL, WiLoR and OmniHands extend the right-hand mesh above the visible grip. ViDiHand2 follows the supporting left hand and the right-hand grip.}
\label{S-fig:distinct_xperience_05}
}

\groupqual{
\includegraphics[width=\textwidth]{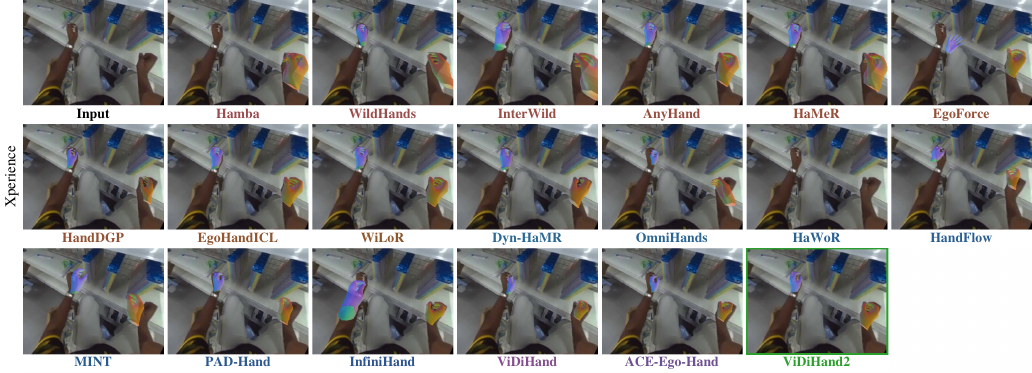}\par\vspace{8pt}
\includegraphics[width=\textwidth]{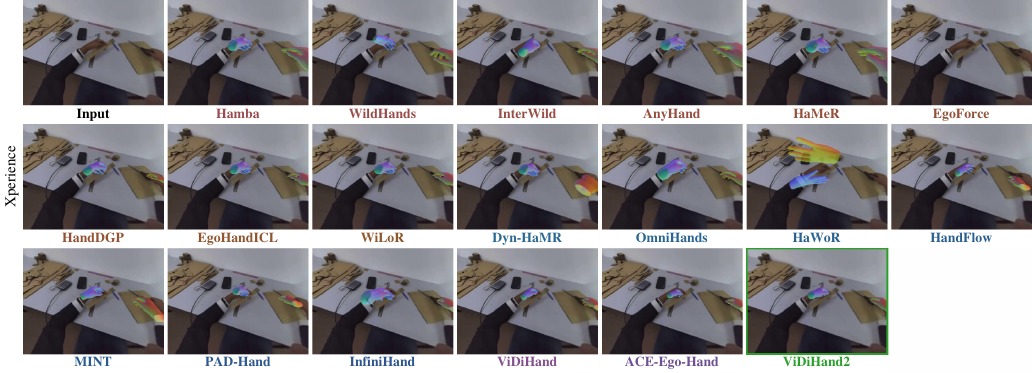}\par\vspace{8pt}
\includegraphics[width=\textwidth]{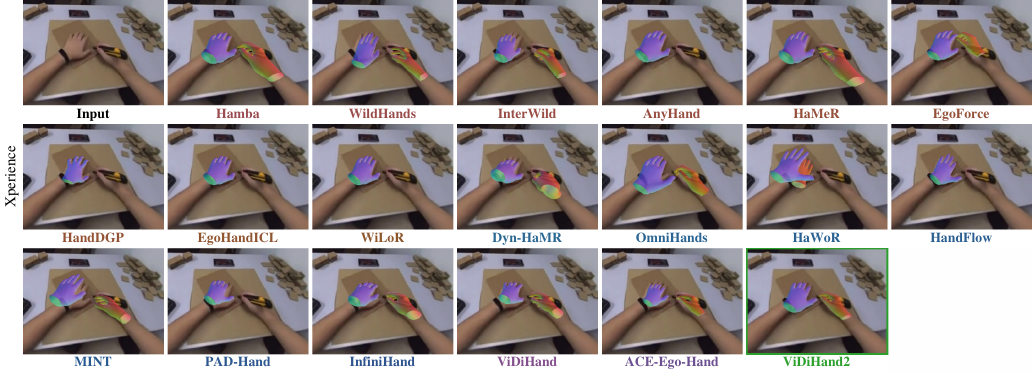}
}{
\caption{\textbf{Top:} \textbf{Qualitative comparison on Xperience.} Without reliable ground truth, the overlays show image alignment only. A person reaches toward items on a shelf. Hamba omits the left-hand overlay, while HaWoR shows neither hand. EgoForce shifts the left hand to the right of the visible reach. InfiniHand enlarges and lowers it, and MINT shifts the right hand inward from the visible fist. ViDiHand2 follows the reaching hand and the right fist. \textbf{Middle:} \textbf{Qualitative comparison on Xperience.} Both hands handle tools and cardboard on a table. EgoForce shows no meshes. HaWoR greatly enlarges and separates both hands, while Dyn-HaMR displaces the right hand to the lower-right image edge. MINT enlarges the right hand, and InfiniHand makes the left hand too broad. ViDiHand2 follows both visible hand positions at the work surface. \textbf{Bottom:} \textbf{Qualitative comparison on Xperience.} A person cuts cardboard with a utility knife while holding it flat. AnyHand, HandDGP, EgoHandICL, WiLoR, PAD-Hand and HandFlow omit the right-hand overlay. HaWoR places both meshes at the left hand instead of the knife grip. EgoForce shifts the right hand upward and left, while Hamba and HaMeR stretch it down the forearm. ViDiHand2 follows the supporting palm and the knife-holding hand.}
\label{S-fig:distinct_xperience_06}
}

\groupqual{
\includegraphics[width=\textwidth]{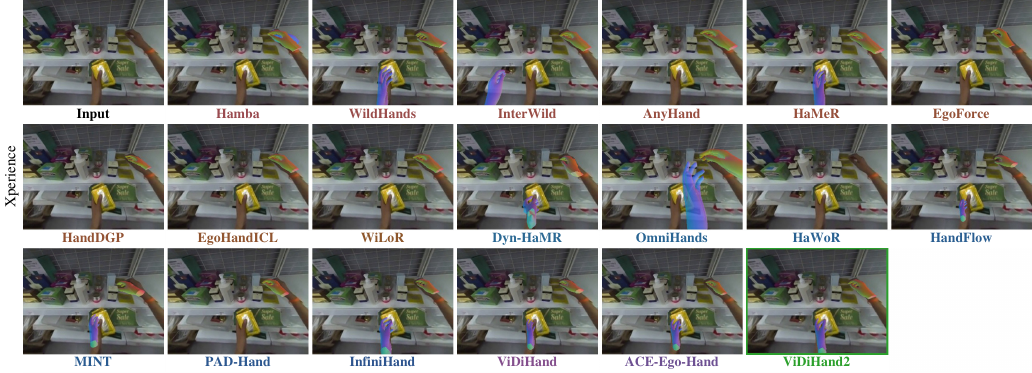}\par\vspace{8pt}
\includegraphics[width=\textwidth]{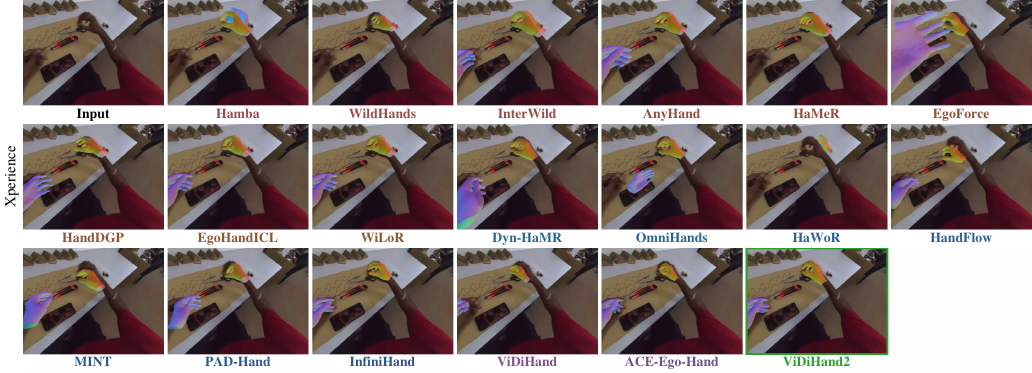}\par\vspace{8pt}
\includegraphics[width=\textwidth]{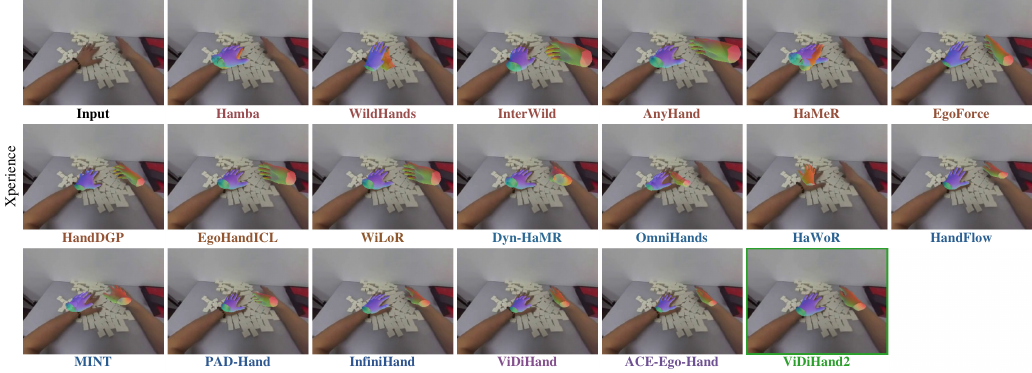}
}{
\caption{\textbf{Top:} \textbf{Qualitative comparison on Xperience.} Without reliable ground truth, the overlays show image alignment only. A person holds a yellow packet while reaching toward shelf items. HaWoR shows no hand meshes, and Hamba, AnyHand, EgoForce, HandDGP, EgoHandICL, WiLoR and PAD-Hand omit the left-hand overlay. InterWild shifts it away from the packet, while OmniHands makes it much too large and extends it up the shelf. ViDiHand2 follows the packet-holding hand and the right-hand reach. \textbf{Middle:} \textbf{Qualitative comparison on Xperience.} Both hands manipulate small pieces beside a cardboard pattern. Hamba overlaps both meshes on the right hand, and HaMeR and WildHands omit the left-hand overlay. EgoForce greatly enlarges it over the work surface, while Dyn-HaMR also makes it too large. HaWoR omits the left hand and shrinks the right, and OmniHands shifts the left hand across the phone. ViDiHand2 follows both hands at their separate working positions. \textbf{Bottom:} \textbf{Qualitative comparison on Xperience.} Both hands sort small white pieces on a table. Hamba omits the right-hand mesh. WildHands, HaMeR and OmniHands overlap both predictions at the visible left hand. HaWoR puts a right-hand mesh there and omits the other hand. AnyHand, EgoHandICL, WiLoR and InterWild greatly enlarge the right hand. ViDiHand2 follows both hands at their separate positions among the pieces.}

}

\boundaryqual{
\includegraphics[width=\textwidth]{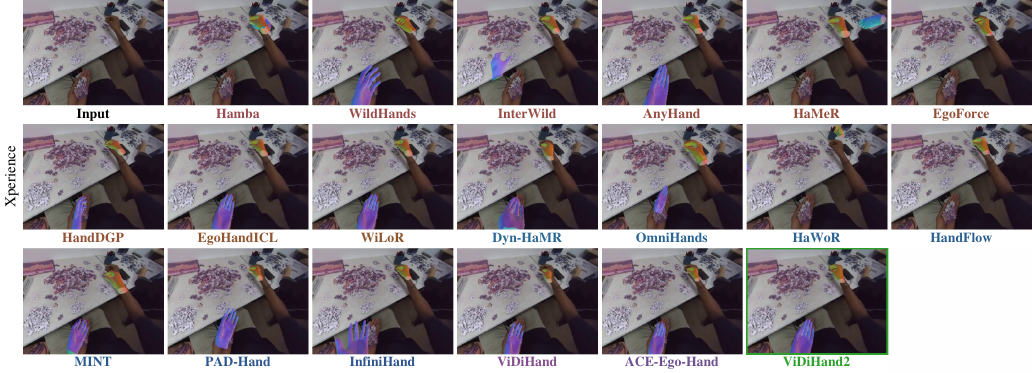}\par\vspace{4pt}
\includegraphics[width=\textwidth]{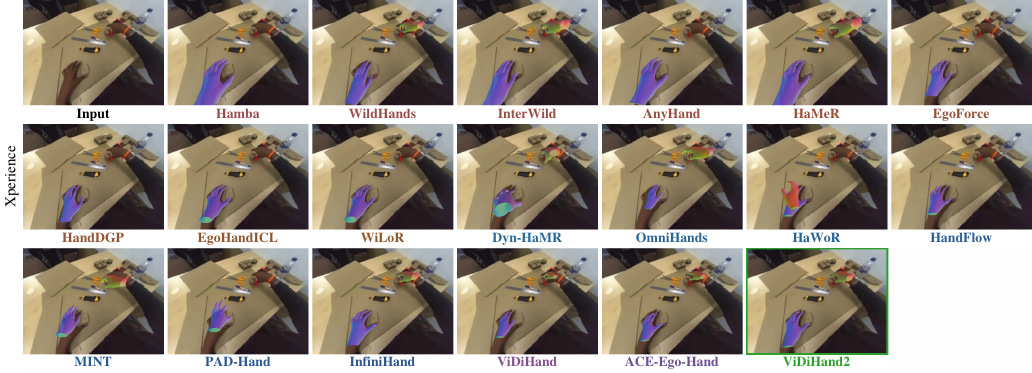}
}{
\caption{\textbf{Top:} \textbf{Qualitative comparison on Xperience.} Without reliable ground truth, the overlays show image alignment only. A person holds puzzle pieces in the left palm while reaching across a shared table. Hamba, HaMeR, EgoForce, HaWoR and HandFlow omit the left-hand overlay. HaMeR places a mesh on another person's hand, while HaWoR shifts the right hand upward onto the pen holder. InterWild shifts the left hand onto the table, and InfiniHand spreads its fingers beyond the visible palm. ViDiHand2 follows the piece-holding palm and the right-hand reach. \textbf{Bottom:} \textbf{Qualitative comparison on Xperience.} The left hand rests on a sheet while the right hand reaches toward small cut blocks. Hamba, AnyHand, EgoForce, HandDGP, EgoHandICL, WiLoR, PAD-Hand and HandFlow fail to detect the right hand. Hamba, AnyHand, HaMeR and InterWild also elongate the left wrist into the foreground. HaWoR overlaps both meshes at the left hand, and Dyn-HaMR distorts the left palm. ViDiHand2 follows the supporting hand and the reaching right hand.}

}{
\subsection{In-Domain Datasets}
\label{S-sec:supp_qual_indomain}
The inset shows the hands and wrist paths, with the ground truth in light colors. ViDiHand2 is marked by a green border.
}{
\includegraphics[width=\textwidth]{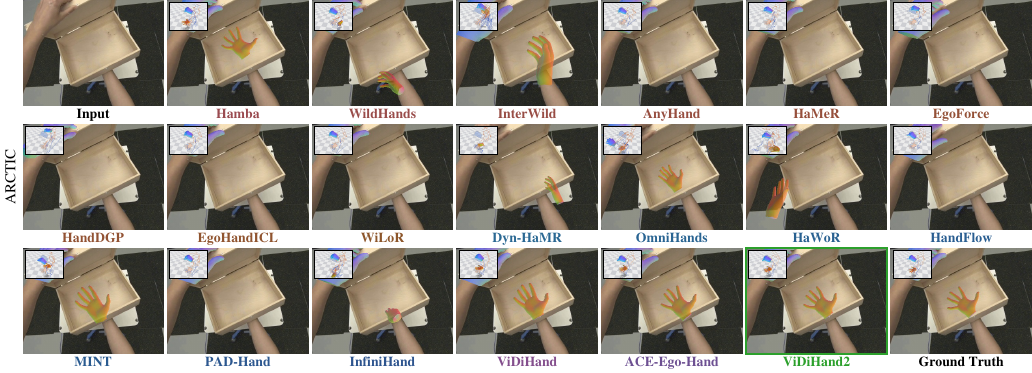}
}{
\caption{\textbf{Qualitative comparison on ARCTIC.} The box is tilted toward the camera, fully occluding the right hand behind its base. ViDiHand2 recovers the hidden palm and spread fingers close to the ground truth. InfiniHand and WildHands shrink this hand and shift it toward the lower box edge, HaWoR places it outside the box, and AnyHand, HaMeR, EgoForce, HandDGP, EgoHandICL, WiLoR, HandFlow and PAD-Hand omit it. The meshes are overlaid on the box to show the occluded hand.}
\label{S-fig:indomain_arctic_01}
}

\groupqual{
\includegraphics[width=\textwidth]{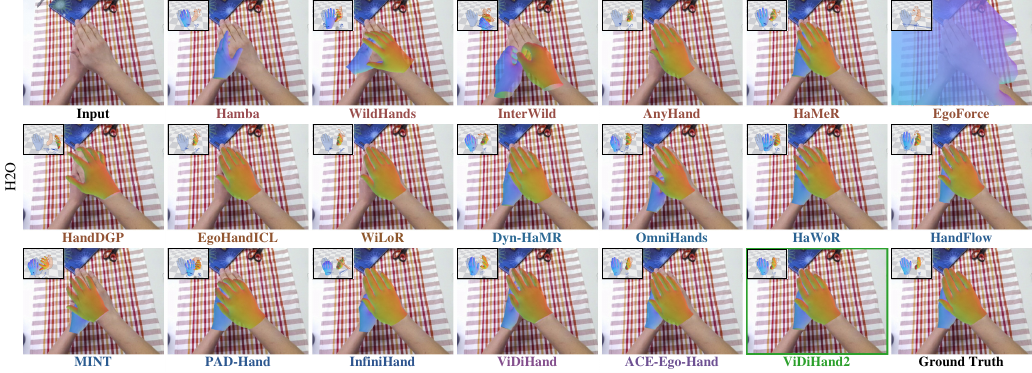}\par\vspace{8pt}
\includegraphics[width=\textwidth]{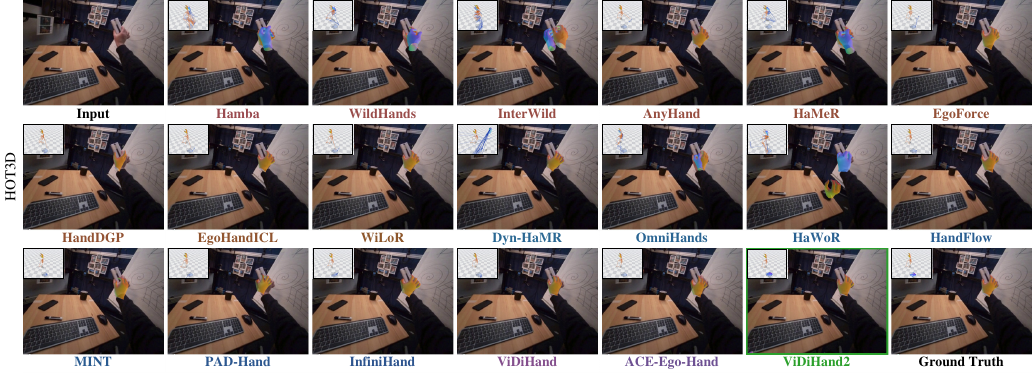}\par\vspace{8pt}
\includegraphics[width=\textwidth]{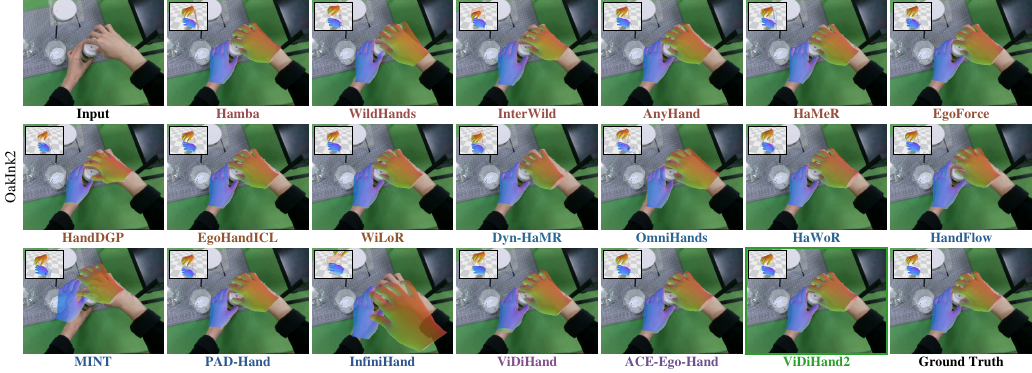}
}{
\caption{\textbf{Top:} \textbf{Qualitative comparison on H2O.} One hand rests over the other on a checkered cloth. Hamba fails to detect the right hand. AnyHand, HandDGP, EgoHandICL and WiLoR fail to detect the left. InterWild separates the hands, and EgoForce produces a single oversized mesh covering the image. ViDiHand2 recovers both hands near the ground-truth overlap. \textbf{Middle:} \textbf{Qualitative comparison on HOT3D.} A raised hand holds a small object beside a desk. Hamba, WildHands and HaMeR assign a left-hand mesh to the visible right hand. InterWild and OmniHands add an overlapping hand, while HaWoR also assigns a left-hand mesh to the right hand and adds a second hand near the mouse. ViDiHand2 retains the visible right hand and follows its position and grip. \textbf{Bottom:} \textbf{Qualitative comparison on OakInk2.} Both hands cooperate to twist a bottle cap. ViDiHand2 preserves the grasp and follows the visible finger contours more closely. InfiniHand enlarges the right hand and shifts it down and to the right, misaligning it with the cap.}
\label{S-fig:indomain_h2o_01}
\label{S-fig:indomain_hot3d_02}
\label{S-fig:indomain_oakink2_01}
}

\groupqual{
\includegraphics[width=\textwidth]{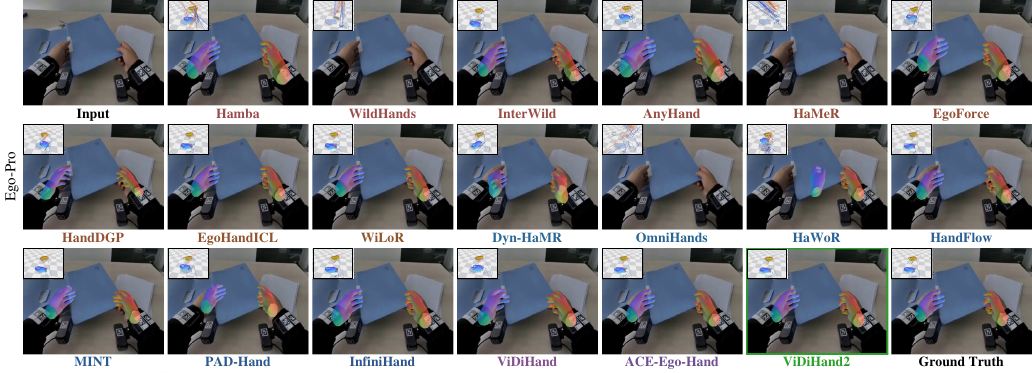}\par\vspace{8pt}
\includegraphics[width=\textwidth]{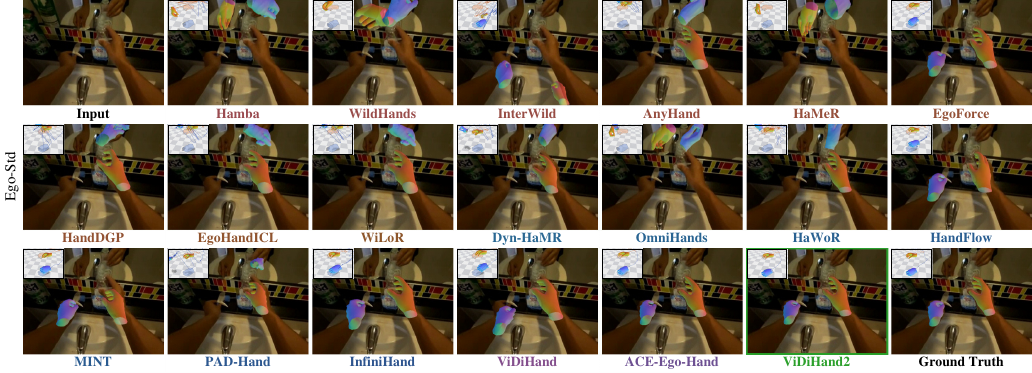}
}{
\caption{\textbf{Top:} \textbf{Qualitative comparison on Ego-Pro.} Both hands hold a blue folder. WildHands, HaMeR and OmniHands omit both overlays. HaWoR shifts the left hand onto the folder. ViDiHand2 follows both grips, close to the ground truth. \textbf{Bottom:} \textbf{Qualitative comparison on Ego-Std.} Hands hold a bottle beside a sink and mirror. Hamba, WildHands, HaMeR and OmniHands place both meshes on the reflected hands, while AnyHand, HandDGP, EgoHandICL, WiLoR, HaWoR, PAD-Hand and Dyn-HaMR place the left-hand mesh on the reflection. InterWild shifts the left hand toward the sink and the right mesh down the forearm. ViDiHand2 follows the foreground hands around the bottle.}
\label{S-fig:indomain_egodemo_egopro_standard_01}
\label{S-fig:indomain_egodemo_egostandard_01}
}

\subsection{In-the-Wild 3D Accuracy with a Ruler as Ground Truth}
\label{S-sec:supp_ruler}
The examples of \cref{S-sec:supp_qual_nogt} have no reliable 3D ground truth. To measure 3D accuracy in the wild, we film a person who
holds a 20\,cm steel ruler between the two thumb tips, so the distance between the thumb tips stays at 20\,cm while the hands
move. The video is recorded with an iPhone camera at 1920$\times$1080 pixels and 29.97\,fps. We score the 8.2\,s clip of
247 frames, during which the ruler is lifted, tilted and turned upright, and each hand in
turn passes behind a cup, for 55 frames in total. All methods use the weights of Tab.~\ref{tab:main} and the same intrinsics. Since a monocular method recovers
metric size only up to the hand size it assumes, each method's distance is multiplied by one factor that sets its median over
a flat hold before the clip to 20\,cm. \Cref{S-fig:supp_ruler} shows the scaled distance of every method over the clip and in
four frames.

\begin{figure}[p]
\centering
\includegraphics[width=\textwidth]{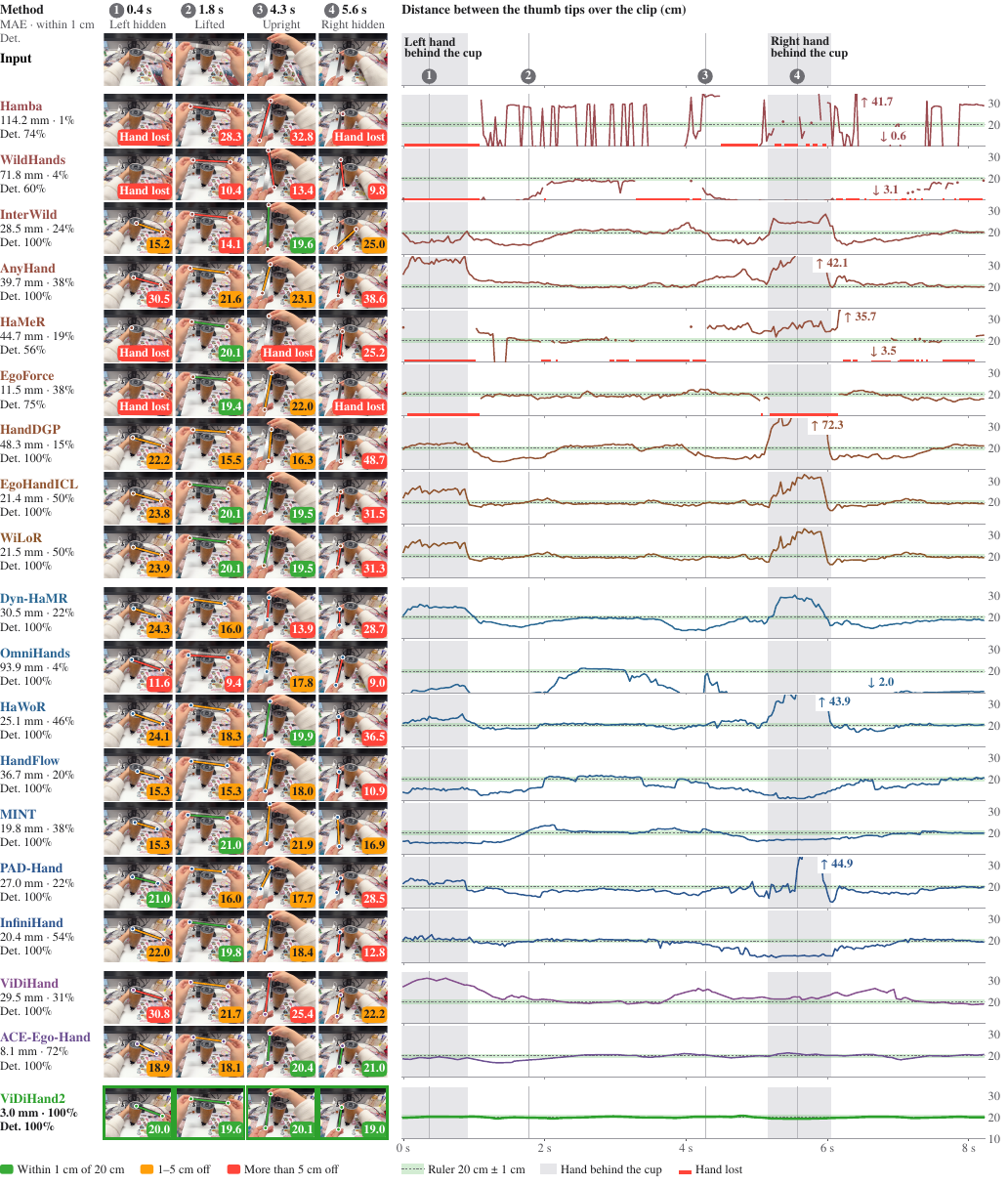}
\caption{\textbf{Thumb distance on a hand-held 20\,cm ruler.} The first row shows the input at four moments: 1, the left hand
behind the cup; 2, the ruler lifted and tilted; 3, the ruler upright; and 4, the right hand behind the cup. The other rows
follow the order of Tab.~\ref{tab:main}, with \ours last in a green border. Each frame shows the predicted thumb tips and the
scaled distance in cm. Under each name are the MAE, the percentage of frames within 1\,cm and Det., the percentage of frames
with both hands detected. The plots show the scaled distance over the clip and print the values beyond the axis.}
\label{S-fig:supp_ruler}
\end{figure}

\begin{table}[tbp]
\centering
\caption{\textbf{Thumb distance on a hand-held 20\,cm ruler,} after scaling each method's distance once as described in
\cref{S-sec:supp_ruler}, over the whole clip from 0 to 8.2\,s and over its frames with a hand behind the cup, 0--0.9\,s
and 5.2--6.0\,s. Det.: percentage of frames in which both hands are detected. MAE: mean absolute error of the distance in mm
over the frames with both hands detected. ${\leq}1$\,cm: percentage of all frames within 1\,cm of 20\,cm, an undetected hand
counted as a miss. Bold marks
the best value of each column, ties included.}
\label{S-tab:supp_ruler}
\setlength{\tabcolsep}{6pt}\renewcommand{\arraystretch}{0.92}\footnotesize
\begin{tabular}{@{}l@{\hspace{6pt}}ccc@{\hspace{14pt}}ccc@{}}
\toprule
& \multicolumn{3}{c}{Whole clip} & \multicolumn{3}{c}{A hand behind the cup} \\
& \multicolumn{3}{c}{247 frames} & \multicolumn{3}{c}{55 frames} \\
\cmidrule(lr){2-4}\cmidrule(l){5-7}
Method & Det.$\uparrow$ & MAE$\downarrow$ & ${\leq}1$\,cm$\uparrow$ & Det.$\uparrow$ & MAE$\downarrow$ & ${\leq}1$\,cm$\uparrow$ \\
\midrule
\multicolumn{7}{@{}l}{\textit{Detect-then-reconstruct, single frame}} \\
Hamba~\cite{S-dong2024hamba} & 74 & 114.2 & 1 & 24 & 62.0 & 2 \\
WildHands~\cite{S-prakash2024wildhands} & 60 & 71.8 & 4 & 51 & 104.0 & 0 \\
InterWild~\cite{S-moon2023interwild} & \best{100} & 28.5 & 24 & \best{100} & 41.5 & 7 \\
AnyHand~\cite{S-si2026anyhand} & \best{100} & 39.7 & 38 & \best{100} & 122.4 & 2 \\
HaMeR~\cite{S-pavlakos2024hamer} & 56 & 44.7 & 19 & 51 & 61.0 & 0 \\
EgoForce~\cite{S-millerdurai2026egoforce} & 75 & 11.5 & 38 & 5 & 10.2 & 4 \\
HandDGP~\cite{S-valassakis2024handdgp} & \best{100} & 48.3 & 15 & \best{100} & 122.4 & 5 \\
EgoHandICL~\cite{S-xie2026egohandicl} & \best{100} & 21.4 & 50 & \best{100} & 63.6 & 0 \\
WiLoR~\cite{S-potamias2025wilor} & \best{100} & 21.5 & 50 & \best{100} & 63.1 & 2 \\
\midrule
\multicolumn{7}{@{}l}{\textit{Detect-then-reconstruct with a motion prior}} \\
Dyn-HaMR~\cite{S-yu2025dynhamr} & \best{100} & 30.5 & 22 & \best{100} & 52.7 & 11 \\
OmniHands~\cite{S-lin2026omnihands} & \best{100} & 93.9 & 4 & \best{100} & 99.1 & 0 \\
HaWoR~\cite{S-zhang2025hawor} & \best{100} & 25.1 & 46 & \best{100} & 79.9 & 4 \\
HandFlow~\cite{S-xu2026handflow} & \best{100} & 36.7 & 20 & \best{100} & 64.2 & 0 \\
MINT~\cite{S-zhu2026mint} & \best{100} & 19.8 & 38 & \best{100} & 38.1 & 0 \\
PAD-Hand~\cite{S-ismayilzada2026padhand} & \best{100} & 27.0 & 22 & \best{100} & 54.4 & 11 \\
InfiniHand~\cite{S-ren2026infinihand} & \best{100} & 20.4 & 54 & \best{100} & 39.7 & 31 \\
\midrule
\multicolumn{7}{@{}l}{\textit{Video diffusion with depth regression}} \\
ViDiHand~\cite{S-wang2026vidihand} & \best{100} & 29.5 & 31 & \best{100} & 58.0 & 0 \\
ACE-Ego-Hand~\cite{S-liu2026aceegohand} & \best{100} & 8.1 & 72 & \best{100} & 9.0 & 58 \\
\midrule
\multicolumn{7}{@{}l}{\textit{Video diffusion with hand geometry}} \\
\textbf{\ours (Ours)} & \best{100} & \best{3.0} & \best{100} & \best{100} & \best{4.5} & \best{100} \\
\bottomrule
\end{tabular}
 \end{table}

As \cref{S-tab:supp_ruler} shows, \ours is within 1\,cm of 20\,cm in every frame, with an MAE of 3.0\,mm over the clip and
4.5\,mm while a hand is behind the cup. The closest baseline, \ace, has an MAE of 8.1\,mm over the clip and 9.0\,mm while a
hand is hidden, and is within 1\,cm in 72\% and 58\% of these frames, respectively. The other baselines that detect both
hands in every frame have MAEs from 19.8 to 93.9\,mm over the clip and from 38.1 to 122.4\,mm while a hand is hidden, where they are
within 1\,cm in at most 31\% of the frames. For each of them the MAE is higher while a hand is hidden. Much of this error is in
depth. At moment 4, for example, WiLoR and HaWoR project the right thumb tip only 25 and 67 pixels from that of \ours
but place it 12.0 and 16.8\,cm deeper after scaling. HaMeR and EgoForce often lose a hand instead: HaMeR returns both hands in only one of the 28 frames with the left hand
behind the cup, and EgoForce in only 3 of the 55 frames with a hand behind the cup. The largest errors of \ours occur while the right
hand is hidden, where its distance is 5 to 10\,mm too short. In this span alone, \ace has a slightly lower MAE, 6.7 against
7.9\,mm, but is within 1\,cm in only 74\% of the frames, against 100\% for \ours.

\endgroup
\section{Implementation Details}
\label{S-sec:supp_impl}

\subsection{Multi-Level Latent Encoder}
The frozen Wan2.2 VAE encodes the whole clip once as one latent stream. As in \ace~\cite{S-liu2026aceegohand}, the first 16 of the 30
blocks of the Wan2.2-Fun-5B-Control transformer~\cite{S-wan2025wan} process the noise-free latents of each chunk without diffusion sampling. LoRA~\cite{S-hu2022lora} adapts the attention and feed-forward layers of these blocks, and the pretrained weights remain frozen.
Each latent frame also carries 48 task tokens for the two hands, the 42 joints and four registers. \ace uses the same tokens but reads
them from the output of the last block with a separate decoder. In \ours, the task tokens instead pass through every adapted block with
the weights of that block. Within a block, the task tokens attend to the video tokens of that block and to all task tokens of the
chunk. The video tokens attend only to each other, over the whole chunk in space and time, so the task tokens do not change the
video features. When the task tokens attend to the video tokens, the video keys keep only the rotary encoding of time. \Cref{S-tab:supp_arch} lists the settings of
the backbone, the temporal decoders, the hand, joint and MANO heads and the Joint-Heatmap Refiner.

\begin{table}[!tbp]
\centering
\caption{\textbf{Architecture settings.} Settings of the backbone, the temporal decoders, the hand, joint and MANO heads and the
Joint-Heatmap Refiner of \ours, with transformer blocks counted from 0.}
\label{S-tab:supp_arch}
\setlength{\tabcolsep}{3pt}
\footnotesize
\begin{tabular}{@{}p{0.2\textwidth}p{0.77\textwidth}@{}}
\toprule
Module & Setting \\
\midrule
\multicolumn{2}{@{}l}{\textit{Backbone}} \\
Transformer & Wan2.2-Fun-5B-Control, width 3072, 24 heads, 30 blocks, blocks 0--15 are used at noise level 0, with blocks 12--15 evaluated twice when the cache is active \\
VAE & Frozen, $16\times$ in space and $4\times$ in time into 48 channels, 81 frames give 21 latent frames \\
Input & Short side of 480 pixels, long side a multiple of 32, intrinsics scaled accordingly \\
Video tokens & $1{\times}2{\times}2$ patches, one token per $32{\times}32$ pixels and latent frame, control channels zero, a fixed caption
embedding as text prompt \\
LoRA & Rank 64 and scale 1 on the self- and cross-attention projections and both feed-forward layers of blocks 0--15 \\
Patch embedding & A trained $1{\times}2{\times}2$ convolution on the 48 latent channels, added to the frozen patch embedding \\
Task tokens & 48 per latent frame, 2 for the hands, 42 for the joints and 4 registers, linear adapters between widths 384 and 3072 \\
Task-token attention & In every block, to the video tokens of that block with rotary encodings of time only, and to all task tokens
of the chunk \\
Chunk cache & Blocks 12--15 also attend to the keys and values of the previous chunk within 21 latent frames, through a learned
bias on the attention logits per block and head, initialized at $-4$ \\
\midrule
\multicolumn{2}{@{}l}{\textit{Decoders}} \\
Inputs & Hand states, joint features, coarse joint locations with variances, the projected feature map and the heat-map features \\
Layers & Topology of the Wan2.2 VAE decoder with temporal kernels of size 3 only: input projection, convolution, two residual blocks,
two stages that double the frame rate with separate kernels for the two output frames, a head of RMS normalization, SiLU and
convolution, output projection \\
Linear path & The same two doubling stages without nonlinearity, initialized as Catmull--Rom interpolation between latent frames \\
Size & Width 128, 3.8M parameters for all five decoders \\
\midrule
\multicolumn{2}{@{}l}{\textit{Hand and joint heads}} \\
Feature map & Linear layer $3072{\to}384$ with LayerNorm on the block-15 video tokens, a learned spatial encoding and a ray encoding of
$\K$, an MLP $32{\to}384{\to}384$ on 8 Fourier octaves of the azimuth and elevation of the ray of each cell \\
Hand attention & In every video frame, one cross-attention with 8 heads from each decoded hand state to the decoded feature map \\
Joint attention & In every latent frame, one cross-attention with 8 heads from each joint token to the feature map \\
Coarse location & Soft-argmax and second moment of the attention map of each joint token \\
\midrule
\multicolumn{2}{@{}l}{\textit{Joint-Heatmap Refiner}} \\
Heat-map cells & 4 pixels, two-layer stride-2 RGB stem $3{\to}32{\to}64$ plus the bilinearly upsampled decoded transformer features,
two convolutions of width 64 \\
Heat-map U-Net & Widths 32, 64, 96 and 128, GroupNorm, dilated bottleneck, per-frame global branch, depthwise temporal
convolution of kernel 5 with a masked mean over the chunk at the three coarsest levels \\
Heat-map location & Soft-argmax within the $7{\times}7$ cells around the arg-max cell \\
Regression branch & Unbounded MLP offset from the decoded coarse location \\
\midrule
\multicolumn{2}{@{}l}{\textit{MANO heads}} \\
Per frame & 6D rotations of the wrist and the 15 finger joints, shape, and 2D and 3D presence of each hand \\
Shape & Median of the per-frame shapes over the chunk in training and over the clip at inference, plus a binned residual \\
Residual heat maps & 65 bins over $\pm0.5$\,rad per rotation-vector component, 65 bins over $\pm2$ per shape coefficient \\
\bottomrule
\end{tabular}
\end{table}

\subsection{MANO Heads}
In every frame, heads on the decoded hand state regress the rotations of the wrist and the 15 finger joints in the 6D
representation~\cite{S-zhou2019continuity}. The same heads regress the shape and the 2D and 3D presence of each hand. Residual heat maps add the expectation of a
distribution over bins to every rotation-vector component and every shape coefficient. The shape of a hand is the median of the
per-frame shapes over the chunk in training and over the clip at inference, plus one residual. This residual uses the
presence-weighted geometric mean of the per-frame shape distributions, so each hand has one shape per clip, as in
\ace~\cite{S-liu2026aceegohand}. The expectation over bin centers follows AdaBins~\cite{S-bhat2021adabins}. The predicted shape is supervised only by
the ground-truth shape. The MANO poses are used frame by frame without smoothing.

\subsection{Joint-Heatmap Refiner and Clip-Level Ray Fitting}
\label{S-sec:supp_lifting}

\paragraph{Hand and joint tokens.}
In every video frame, the decoded state of each hand attends once to the decoded feature map of that frame. We regress the
MANO pose, shape and presence of the hand from the attended state. Each joint token instead attends to the feature map of its latent frame. The soft-argmax of
this attention map gives a coarse location with a variance from the second moment. An auxiliary head on the attended joint feature
predicts wrist-relative joints and is supervised by $\ell_\text{3d}$. The decoders of \cref{S-sec:supp_temporal} bring the coarse
locations, their variances and the joint features to every video frame. The heat maps and the regression branch use these per-frame joint features $\vect{e}_j$.

\paragraph{Heat maps.}
On a grid of 4-pixel cells, the heat-map features are the sum of a two-layer convolutional stem on the RGB frame and the bilinearly
upsampled decoded transformer features. The logit of joint $j$ at a cell is the scaled dot product of the projected joint feature with the
cell feature. The null logit
\begin{equation}
\eta_j=\vect{w}_n^\top\mathrm{LN}(\vect{e}_j)+b_n+\log|C|-2
\end{equation}
represents joint $j$ being outside the image or absent, where $|C|$ is the number of cells and LN a LayerNorm. A U-Net refines the 42 maps jointly from their
log-softmax, the null probabilities, the heat-map features and a Gaussian at the coarse location of each joint. The
variance of this Gaussian is that of the coarse location plus one cell. At the three coarsest levels of the U-Net, a depthwise
convolution over the frames of the chunk adds temporal context. The output of the U-Net changes only the distribution of each map
over the cells, not the null probability. The heat-map location $\hat{\vect{u}}_j$ is the soft-argmax within a neighborhood of cells
around the arg-max cell, with variance $\sigma^2_{\text{hm},j}$ and neighborhood mass $m_j$.

\paragraph{Regression branch and confidence.}
The regression branch predicts $\tilde{\vect{u}}_j$ as the coarse location plus an unbounded offset, so $\tilde{\vect{u}}_j$ may lie
beyond the image border. The weight $\alpha_j$ of \cref{eq:read} falls linearly to zero as the joint approaches the image border,
as the null probability of its heat map rises and as the mass around the heat-map peak drops, so the regressed location takes over
wherever the heat map is unreliable. The variance of the regressed location grows with its distance beyond the border. The uncalibrated variance
$\bar{\sigma}_j^2$ of $\vect{u}_j$ blends the variances of the two locations with the weight $\alpha_j$ and grows when the two locations
disagree. The confidence $c_j$ is a logistic regression on the statistics of the heat map and the joint feature $\vect{e}_j$,
trained to predict whether $\hat{\vect{u}}_j$ lies within 16 pixels of the true location.

\paragraph{Clip-Level Ray Fitting.}
Fine-Grained Geometric Lifting and Clip-Level Ray Fitting minimize \cref{eq:rayfit} over the wrist trajectory of each hand. The joint rays $\vect{q}_j=\K^{-1}[\vect{u}_j;1]$ have a unit
component along the optical axis, so that, with the relative depths $d_j$ of the MANO joints, the bone term places every joint at
$(Z_t+d_j)\,\vect{q}_j$ as in \cref{eq:bone}. Its 41 pairs are the 20 bones of the hand, the six pairs of the four finger knuckles
and the 15 pairs of the wrist with a non-adjacent joint.

\looseness=-1 \emph{Variance of a 2D location.} The heat-map location has the second moment of the heat map in the $7{\times}7$ cells around its
peak as variance $\sigma^2_{\text{hm},j}$, at least 1\,px$^2$. The regressed location has the variance
$\sigma^2_{\text{reg},j}=\big(16\,(1+\max(-\tilde{\xi}_j,0)/64)\big)^2$ plus the variance of the coarse location, where $\tilde{\xi}_j$ is its signed distance to
the image border in pixels, negative beyond the border. The weight of \cref{eq:read} is
\begin{equation}
\alpha_j=\Big[\tfrac{\xi_j-6}{10}\Big]_0^1\cdot\Big[\tfrac{0.7-p^\text{null}_j}{0.4}\Big]_0^1\cdot\Big[\tfrac{m_j-0.3}{0.4}\Big]_0^1 ,
\end{equation}
where $[\cdot]_0^1$ clamps to $[0,1]$, $\xi_j$ is the smaller border distance of the two locations, $p^\text{null}_j$ the null
probability and $m_j$ the share of the probability mass around the peak. The variance of the blended location is
\begin{equation}
\bar{\sigma}_j^2=(\sigma^2_{\text{hm},j})^{\alpha_j}(\sigma^2_{\text{reg},j})^{1-\alpha_j}+\alpha_j(1-\alpha_j)\,\delta_j^2+\alpha_j\max\!\big(\delta_j^2-9\,(\sigma^2_{\text{hm},j}+\sigma^2_{\text{reg},j}),0\big) ,
\end{equation}
with $\delta_j$ the distance between the two locations, so that the variance grows where they disagree. The fit uses
\mbox{$\sigma_j^2=\bar{\sigma}_j^2\,c_j^{-2\alpha_j}\exp(\zeta_j)$}, with the confidence $c_j$ clipped to $[0.05,1]$ and a learned correction
$\zeta_j\in[-10,10]$.

\emph{Residuals and weights.} The weights $w^{\mathrm{bone}}$ and $w^{\mathrm{proj}}$ of \cref{eq:rayfit} are inverse variances,
and the Huber penalty $\psi$ of each term turns linear at three standard deviations, so that each weighted term is the Huber penalty
of its residual in units of its standard deviation. A learned scalar $s_j$ per joint, in
meters, models the error of the MANO joint. It is zero at the wrist and initialized between 1.5 and 8.1\,mm. The reprojection
residual, in pixels, has the covariance $\mathbf{C}_j=\sigma_j^2\,\mathbf{I}+s_j^2\,\mathbf{J}_j\mathbf{J}_j^\top$, where $\mathbf{J}_j$ is the
Jacobian of the projection with respect to the joint. The residual is whitened with $\mathbf{C}_j$, and
\cref{eq:rayfit} writes the case $s_j=0$, with $w^{\mathrm{proj}}=1/\sigma_j^2$. The bone residual of the pair
$(a,p)$, in meters, has the variance
\begin{equation}
(D_a^2\sigma_a^2+D_p^2\sigma_p^2)\Big(\tfrac{n_x^2}{f_x^2}+\tfrac{n_y^2}{f_y^2}\Big)+s_a^2\,\|\partial\rho/\partial\vect{P}_a\|^2+s_p^2\,\|\partial\rho/\partial\vect{P}_p\|^2 ,
\end{equation}
where $D$ is the depth of a joint, $\vect{n}$ the unit direction of the pair, $f_x$ and $f_y$ the focal lengths and $\partial\rho/\partial\vect{P}$ the
derivative of the residual with respect to the MANO joint, and $w^{\mathrm{bone}}$ is its inverse. A variance in square pixels thus becomes a
variance in square meters through the depth and the focal length. All
variances are computed once at the start point of the fit and then kept fixed. A joint beyond the image border thus has a large
variance and constrains the depth far less than a sharp heat-map location.

\emph{Temporal term.} The weight $\vect{\Lambda}_t$ is diagonal. For a frame rate $\nu$ and an acceleration scale $\varsigma$ in
m/s$^2$, its entries are $\nu^2/\varsigma$, so the term measures the acceleration of the wrist in units of $\varsigma$. The scale is
1.5\,m/s$^2$ along the optical axis and is predicted for the two lateral axes in every frame, initialized at 4.5\,m/s$^2$. The term
couples all frames between the first and the last frame in which the hand is detected. A frame in between without a detection has no
data term and is placed by the temporal term alone. The first and the last frame have no temporal residual of their own.

\emph{Learned heads.} Two small heads refine these weights. The first is an MLP with one hidden layer of width 32 that predicts the
correction $\zeta_j$ of every 2D location from nine cues: the confidence, the variance, the null probability, the border distance, the
weight $\alpha_j$, the residual of the per-frame initialization below and three averages over the hand. The second, of the same size,
predicts the lateral acceleration scale of every hand-frame from the speed and the acceleration of the 2D locations relative to the
image size of the hand, the change of this size, and the mean confidence, weight $\alpha$ and null probability of the hand. Both are
trained in the first stage together with the rest of the model, with the loss $\ell_\text{e2e}$ of \cref{S-tab:supp_loss} on the output
hand.

\emph{Initialization and solver.} The fit runs in 64-bit floating point in three steps. First, every frame alone gets a closed-form
start: for every pair of joints, the depth at which the pair has its MANO length on its two rays~\cite{S-iqbal2018latent25d}, the
median over the pairs as the depth, and the median over the joints as the lateral position. Second, a fit of the depth alone over the
clip, with the bone residuals along the wrist ray and the temporal term on the depth, refines this start by at most 200 damped
Gauss--Newton iterations. Third, the full fit of \cref{eq:rayfit} runs for at most 40 Levenberg--Marquardt steps, with banded
linear systems for long clips. The damping is $10^{-3}$ times the diagonal of the normal equations, grows tenfold after a rejected step and
shrinks tenfold after an accepted one. The fit stops when every gradient component, scaled by the square root of its diagonal entry,
falls below $10^{-4}$. Every joint with a 2D location is constrained to lie in front of the camera, and a step that leaves this region
is rejected. A frame without coupled neighbors is placed when it has at least two 2D locations, and a run of coupled frames when its
frames together provide at least six constraints. A hand-frame that the fit cannot place is not reported and counts as a detection
failure. The fit has no depth prior and no bound on the depth.

\emph{Gradient through the fit.} At the solution $\hat{\vect{r}}$, the implicit function theorem gives the gradient of a loss
$\mathcal{L}$ on the output hand with respect to the parameters $\vartheta$ of the two heads and the scalars $s_j$ as
$-\partial_\vartheta\langle\vect{\lambda},\nabla_{\vect{r}}E(\hat{\vect{r}};\vartheta)\rangle$, where $\vect{\lambda}$ solves
$\mathbf{H}\vect{\lambda}=\partial\mathcal{L}/\partial\hat{\vect{r}}$ with the exact Hessian $\mathbf{H}$ of the energy, by an undamped Cholesky
factorization. A hand whose fit has not converged, or whose Hessian is not positive definite, contributes no
gradient. The MANO hand and the 2D locations learn from their own losses.

\paragraph{Hand size as a depth cue.}
Across the six training splits, the central 98\% of the wrist depths of the hands in the image lie between 0.19 and 0.64\,m. This
range spans 3.37$\times$ around a median of 0.34\,m. The total bone length of the same hand-frames spans 1.28$\times$, and at
most 1.19$\times$ within one dataset. The total length of the 20 bones is 0.75\,m at the median training shape. Moving any
single MANO shape coefficient to either end of its central 99\% training range changes this size by $-5.19\%$ to $+16.87\%$. If the predicted hand size deviates
by some factor, the fitted depth is scaled by approximately the same factor. For fixed joint rays, this relative depth error is therefore independent of the hand-to-camera distance.
In the wild, \ace, which regresses the depth, predicts hands 7\% too small but
places them 26\% too far in the median, and \vidihand predicts them 2\% too small but 21\% too far. A hand predicted too small should be placed too near, so this regressed depth is inconsistent with the predicted
hand size. \ours predicts the in-the-wild hands 4\% too small and places them 2\% too near.

\subsection{Hand Latent Decoder and Autoregressive Rollout Adaptation}
\label{S-sec:supp_temporal}

\paragraph{Hand Latent Decoder.}
The transformer and the task tokens operate at the latent frame rate, with one latent frame per four video frames after the first. The Hand Latent Decoder contains five small
decoders that bring the latent-rate quantities of \cref{S-tab:supp_arch} to the video frame rate, so the hand states of every frame come from learned temporal decoding. Each decoder follows the Wan2.2 VAE decoder~\cite{S-wan2025wan} but has temporal kernels only. Two stages each double the
frame rate, with separate kernels for the two output frames of each input frame. All temporal kernels are centered and use replicate padding at both
ends of the decoded span. A linear path, initialized as Catmull--Rom interpolation between latent frames, is added to each
decoder.

\paragraph{Autoregressive chunks.}
In the second training stage, each chunk follows two preceding chunks of the same clip, as at inference. The seam loss compares the
change of the predictions across each chunk boundary with that of the ground truth,
\begin{equation}
\ell_\text{seam}=10\,\big\|\Delta\vect{P}-\Delta\vect{P}^{*}\big\|_1+\big\|\Delta\tilde{\vect{u}}-\Delta\tilde{\vect{u}}^{*}\big\|_1
+\big\|\Delta\hat{\vect{u}}-\Delta\hat{\vect{u}}^{*}\big\|_1 ,
\label{S-eq:supp_seam}
\end{equation}
where $\Delta$ is the change from the last frame of the previous chunk to the first frame of the current one. Here, $\vect{P}$ are the
wrist-relative MANO joints, and the superscript $*$ marks the ground truth. The predictions of the previous chunk are detached, and the
loss trains only the cache bias.

\subsection{Confidence-Gated Fusion}
\label{S-sec:supp_fusion}

\paragraph{Gate.}
The five part gates below are five heads that give 11 values per hand and frame. In the gate of \cref{eq:gate}, the joint feature $\vect{e}_j$ passes through a LayerNorm before the linear term
$\vect{v}^\top\vect{e}_j$. The gate has one value for every
joint, hand and frame and is not smoothed over time. Both estimates lie at the depth $Z+d_j$, so the blend moves each joint only
parallel to the image plane, and the projection of $\vect{B}_j$ is $g_j$ times the 2D location plus $1-g_j$ times the
projected MANO joint. For this 2D location, \cref{eq:read} is evaluated with the same weight $\alpha_j$ but with the heat-map location
read from a learned weighted combination of the log heat maps of neighboring frames, while the fit uses the heat map of the frame itself. The blend treats each joint independently, without a constraint on the bone lengths, and is therefore not
output directly but passed to the hierarchical fusion below.

\paragraph{Hierarchical fusion.}
\looseness=-1 Let $\vect{F}_j=\vect{P}_j+\hat{\vect{r}}$ be the MANO joints and $\bar{\vect{F}}$ and $\bar{\vect{B}}$ the centroids of $\vect{F}$ and
$\vect{B}$ over the 21 joints. The Kabsch algorithm gives the rotation $\mathbf{R}$ of the least-squares rigid fit of $\vect{F}$ to $\vect{B}$ by a
singular value decomposition of the cross-covariance about the centroids, without reflections. This rotation and the
translation $\vect{o}=\bar{\vect{B}}-\bar{\vect{F}}$ give the rigidly moved hand
\begin{equation}
\vect{x}_j=\bar{\vect{F}}+\mathbf{R}\,(\vect{F}_j-\bar{\vect{F}})+\tau\,\vect{o},
\label{S-eq:supp_rigid}
\end{equation}
in which the rotation is applied in full and the translation with the share $\tau$. The palm, that is the wrist and the five MCP
joints, keeps $\vect{x}_j$. For a finger $k$ with MCP joint $m$, $\mathbf{R}_k$ is the least-squares rotation about the MCP joint that takes
$\vect{x}_j-\vect{x}_m$ to $\vect{B}_j-\vect{B}_m$ over the PIP, DIP and tip joints $j$ of the finger, and $\vect{\omega}_k$ its
axis-angle vector. The power $\mathbf{R}_k^{\mu_k}=\exp(\mu_k[\vect{\omega}_k]_\times)$ rotates these joints about $\vect{x}_m$ by the
fitted angle times the factor $\mu_k$, $\vect{y}_j=\vect{x}_m+\mathbf{R}_k^{\mu_k}(\vect{x}_j-\vect{x}_m)$, so $\mu_k=1$ applies the fitted rotation, $\mu_k=0$ none and $\mu_k>1$ more, and
each of these joints then takes the share $s_k$ of the remaining offset to the blend,
\begin{equation}
\vect{X}_j=(1-s_k)\,\vect{y}_j+s_k\big(\vect{B}_j-(1-\tau)\,\vect{o}\big),
\label{S-eq:supp_fuse}
\end{equation}
where $\vect{B}_j-(1-\tau)\,\vect{o}$ is the blended joint shifted by the part of the translation that the hand does not take. With
$g_j=0$ for all joints, $\vect{B}$ is the MANO hand and \cref{S-eq:supp_fuse} returns it. Unlike the blend, the rotations of the hierarchical fusion can change the depth of a joint. Both fits depend only on the MANO joints, the
back-projected 2D locations and the gate $g_j$, not on the shares. They are computed in double precision and are not
differentiated through, so the gradients reach the shares, the gate and the two estimates through the centroids and the remaining
offsets.

\paragraph{Part gates.}
The share $\tau$, the factor $\mu_\text{th}$ and the share $s_\text{th}$ of the thumb, and the factor $\mu$ and the share $s$ of the other four
fingers come from five part gates for every hand and frame. The gates of $\mu$ and $s$ are each one head that is evaluated for each of
the four fingers. Every part gate is $\kappa\,\mathrm{sigmoid}\big(\mathrm{MLP}(\cdot)+a\,\bar{z}\big)$, with the cap $\kappa=1$
for the shares and $\kappa=2$ for the factors, a learned scalar $a$, and the mean $\bar{z}$ of $\mathrm{logit}(c_j)$ over the
joints of its part, the palm for $\tau$. The MLP has one hidden layer of 64 units with GELU. It receives the mean of the joint
features $\vect{e}_j$ over the part after the LayerNorm of the gate of \cref{eq:gate}, for the finger and thumb gates also the mean
over the palm, and 16 scalar cues of the hand and frame: the logarithms of $\|\vect{o}\|$, of the rotation angle of $\mathbf{R}$, of the
root mean square of $\vect{B}_j-\vect{x}_j$ at $\tau=1$, of the five finger angles $\|\vect{\omega}_k\|$ and of the depth of
$\bar{\vect{F}}$, the hand presence, and $\bar{z}$ of the four fingers, the thumb and the palm. The part gates hold 0.23M
parameters and act on each frame independently, without smoothing over time.

\paragraph{Gate loss.}
The parameters $\gamma_1$, $\gamma_0$, $\vect{v}$ and the LayerNorm of the gate are trained in the first stage with a loss on the
blended joints, whose gradient reaches only the gate. For a joint $\vect{X}_j$ with ground truth
$\vect{X}^*_j$, let
\begin{equation}
\epsilon(\vect{X}_j)=\big\|\vect{X}_j-\vect{X}^*_j\big\|+\big\|\vect{\Pi}_j\,(\vect{X}_j-\vect{X}^*_j)\big\| ,
\end{equation}
where $\vect{\Pi}_j$ projects onto the plane across the viewing ray of the joint. The second term adds weight to the lateral error, which the gate
changes at a fixed depth. The loss is
\begin{equation}
\ell_\text{gate}=\epsilon(\vect{B}_j)+\max\big(0,\,\epsilon(\vect{B}_j)-\epsilon(\vect{F}_j)\big) ,
\label{S-eq:supp_gatefit}
\end{equation}
averaged over the joints, so a blended joint that is less accurate than the MANO joint $\vect{F}_j$ is penalized twice.
The part gates are trained on the L1 error and the lateral error of the output joints
$\vect{X}_j$ of \cref{S-eq:supp_fuse} and, with a small weight, on their acceleration.

\paragraph{Learned gate.}
\Cref{S-tab:supp_gate} gives the mean gate by joint type. The gate is nearly closed at the wrist and the knuckles and opens toward
the fingertips, so the proximal joints follow the MANO hand and the contribution of the 2D locations increases toward the fingertips. In the wild, the gate is lower or equal for every joint type, consistent with 2D locations that are less reliable there. The part
gates change more between the two settings. On the same hand-frames, averaged over the six in-domain datasets and in the wild, the mean translation share $\tau$ is 0.62 and 0.20, the thumb factor $\mu_\text{th}$ 1.87 and 1.99,
the thumb share $s_\text{th}$ 0.42 and 0.36, the finger factor $\mu$ 1.93 and 0.28 and the finger share $s$ 0.46 and 0.03. In the wild, the
four fingers thus stay close to the rigidly moved MANO hand, while the thumb follows the 2D locations in both settings.

\begin{table}[H]
\centering
\caption{\textbf{Mean gate by joint type.} Mean of the gate $g_j$ of \cref{eq:gate} over the detected hand-frames with ground truth of all 688 test clips,
pooled over the six in-domain datasets, and on \genhuman in the wild. A gate of zero selects the MANO joint for the blend
$\vect{B}_j$, and a gate of one the back-projected 2D location. MCP, PIP and DIP are the metacarpophalangeal, proximal and distal
interphalangeal joints.}
\label{S-tab:supp_gate}
\footnotesize
\setlength{\tabcolsep}{6pt}
\begin{tabular}{@{}lcccccc@{}}
\toprule
& Wrist & MCP & PIP & DIP & Fingertip & All joints \\
\midrule
In-domain & 0.04 & 0.06 & 0.20 & 0.36 & 0.49 & 0.27  \\
In-the-wild & 0.04 & 0.05 & 0.15 & 0.25 & 0.34 & 0.19  \\
\bottomrule
\end{tabular}
\end{table}

\subsection{Losses}
\Cref{S-tab:supp_loss} lists all loss terms, with 2D quantities in normalized image units and joints in meters.

\begin{table}[H]
\centering
\caption{\textbf{Loss terms.} The weight and the supervised quantity of every loss term. The terms below the rule pass their
gradient only to the parameters named in their row. The gate $g_j$ of \cref{eq:gate} is trained with $\ell_\text{gate}$ of
\cref{S-eq:supp_gatefit}, and the part gates of \cref{S-eq:supp_fuse} with $\ell_\text{part}$.}
\label{S-tab:supp_loss}
\setlength{\tabcolsep}{3pt}
\footnotesize
\begin{tabular}{@{}p{0.2\textwidth}rp{0.62\textwidth}@{}}
\toprule
Term & Weight & Supervises \\
\midrule
$\ell_\text{rot}$ & 1 & Geodesic and matrix loss of the global and local rotations \\
$\ell_{\beta,\text{chunk}}$ & 0.1 & L1 of the chunk shape against the ground-truth shape \\
$\ell_{\beta,\text{frame}}$ & 1 & L1 of every per-frame shape against the ground-truth shape \\
$\ell_\text{rel}$ & 10 & L1 of the wrist-relative MANO joints \\
$\ell_\text{3d}$ & 10 & L1 of the auxiliary wrist-relative joints of the joint tokens \\
$\ell_\text{coarse}$ & 1 & L1 of the coarse location of each joint, on joints inside the image \\
$\ell_\text{attn}$ & 1 & KL divergence of each joint-token attention map to a one-cell Gaussian at the label of the joint \\
$\ell_\text{hm}$ & 1 & Heat-map cross-entropy over cells and null with a one-cell Gaussian, and a null target for joints beyond the
image border or absent \\
$\ell_\text{loc}$ & 1 & L1 of the heat-map location \\
$\ell_\text{reg}$ & 1 & L1 of the regressed location within two image widths of the image center \\
$\ell_c$ & 1 & Class-balanced binary cross-entropy of the confidence $c_j$ against a heat-map location within 16\,px of the label \\
$\ell_\text{proj}$ & 1 & L1 of the projected MANO hand at the detached wrist against the 2D labels \\
$\ell_\text{pres,3D}$ & 0.5 & Binary cross-entropy of the 3D presence \\
$\ell_\text{pres,2D}$ & 0.25 & Binary cross-entropy of the 2D presence \\
$\ell_\text{ray}$ & 1 & L1 of an auxiliary head that predicts the camera ray of every pixel, which is not used at test time \\
$\ell_\text{heat,rot}$ & 1 & Cross-entropy of the rotation heat maps against the ground-truth residual \\
$\ell_\text{heat,shape}$ & 0.1 & Cross-entropy of the shape heat maps against the ground-truth residual \\
$\ell_\text{smooth}$ & 0.5 & Acceleration of the wrist-relative joints under the identity global rotation, without ground truth \\
\midrule
$\ell_\text{seam}$ & 1 & Seam loss of \cref{S-eq:supp_seam} on autoregressive rollouts, gradient to the cache bias only \\
$\ell_\text{e2e}$ & 1 & L1 of the output hand, absolute, root-relative, 2D and acceleration, each normalized by its running scale, through the fit by implicit differentiation, gradient to the two weight heads and the learned scalars $s_j$ only \\
$\ell_\text{gate}$ & 1 & Gate loss of \cref{S-eq:supp_gatefit} on the blended joints, gradient to the gate only \\
$\ell_\text{part}$ & 1 & L1 and lateral error of the output joints of \cref{S-eq:supp_fuse} and, with a small weight, their acceleration, gradient to the part gates only \\
\bottomrule
\end{tabular}
\end{table}

\subsection{Training}
\ours is trained in the two stages of \cref{S-tab:supp_stages} on the six in-domain training splits of \cref{S-tab:supp_data},
starting from the released Wan2.2 weights. The training chunks of 21 latent frames come from one causal latent stream per clip,
as at inference, and every frame contributes to the losses. Speed augmentation subsamples 60\% of the
chunks by 2$\times$ or 3$\times$ with equal probability. The fit keeps the original frame rate, so these chunks show faster
hand motion than the recordings. We also paste black occluders into 25\% and other hands into 20\% of the chunks and low-pass
filter 30\% of the chunks. Clips with the smallest and the largest hands are sampled twice as often. Both stages use AdamW with
$\beta=(0.9,0.999)$ and a total batch size of eight chunks on 8 A100 GPUs, and gradients are clipped at a norm of 1.0. The first stage trains the LoRA
adapters, the task tokens, the heads, the decoders and the gates for 50k steps, with a learning rate of $2{\times}10^{-4}$ for the
heads and $10^{-4}$ for LoRA after a warm-up of 200 steps. The second stage trains the cache bias for 3k steps on autoregressive
rollouts.
\ours has 3.0B parameters, of which 107.5M are trained. The LoRA adapters and the trained patch convolution hold 86.6M of them, the task-token adapters
2.4M and the heads 18.5M, including the 3.8M of the decoders and the 0.23M of the part gates. The frozen part is blocks 0--15 of the Wan2.2
transformer with its embeddings and the encoder of the Wan2.2 VAE.

\begin{table}[H]
\centering
\caption{\textbf{Training stages.} The second stage starts from the checkpoint of the first.}
\label{S-tab:supp_stages}
\setlength{\tabcolsep}{3pt}
\footnotesize
\begin{tabular}{@{}lp{0.72\textwidth}r@{}}
\toprule
Stage & What is trained & Steps \\
\midrule
1 & LoRA adapters, task tokens, heads, decoders, the gate of \cref{eq:gate} and the part gates of \cref{S-eq:supp_fuse}, with all losses of \cref{S-tab:supp_loss} except
the seam loss & 50k \\
2 & Only the cache bias, on autoregressive rollouts with the seam loss & 3k \\
\bottomrule
\end{tabular}
\end{table}

\subsection{Inference}
\label{S-sec:supp_clip}
\looseness=-1 The first chunk covers 81 frames and every later chunk the next 84, so
every chunk holds 21 latent frames. A clip that does not end on a latent boundary is padded by repeating the last frame, and the
padded frames are dropped from the output. The chunks are processed autoregressively, and from block 12 on, the
transformer of each
chunk attends to the cached keys and values of the previous chunk. When the cache is used, blocks 12 to 15 run twice, once with and once
without the cache. The pass with the cache provides the task-token states for the pose. The pass without the cache provides the
video features and the states for the shape and the presence. The decoders of each chunk also read the last two latent frames of the previous chunk and the first
latent frame of the next. The chunks do not overlap, but their decoded spans do, and the decoded features of the overlap are averaged. After the last
chunk, a clip-level pass takes the median of the per-frame shapes of each hand and adds the clip-wide shape residual. It then
applies MANO and the Clip-Level Ray Fitting to the whole clip. \Cref{eq:lock} then blends the two estimates of each joint with the learned gate, and
\cref{S-eq:supp_fuse} moves the MANO hand onto the blend to give the output joints. The hand presence confidence threshold is 0.05. Hand-frames that the fit cannot place and hands with a joint
behind the camera are left out of the output.

\section{Ablation Details}
\label{S-sec:supp_abl}
This section defines the variants of Tabs.~\ref{tab:abl_backbone} to~\ref{tab:abl_fusion}, named as in their rows, adds EPE2D-p to each table, and gives the ablation of Autoregressive Rollout Adaptation and the setting of the focal sweep of Tab.~\ref{tab:focal}. All
variants are evaluated on the full test sets of the seven datasets under the protocol of \cref{S-sec:supp_protocol}. The row marked ``(Ours)'' in every
table is the design that \ours uses, and its numbers are those of Tab.~\ref{tab:main}.

\subsection{Backbone}
The rows ``DINOv3'' and ``V-JEPA~2'' keep the heads and the losses of \ours and replace the encoder. LoRA of rank 64 adapts every
block of the frozen encoder, a linear layer maps its features to the width of the heads, and a decoder on these features reads the
task tokens. The row ``Wan2.2, random weights'' trains the architecture of \ours without the pretrained weights of Wan2.2. The row
``Wan2.2, no task tokens in the blocks'' reads the task tokens of the pretrained Wan2.2 with a separate decoder after the last block instead of passing them through every block.

\begin{table}[H]
\centering
\caption{\textbf{Ablation of the backbone with EPE2D-p.} The rows of Tab.~\ref{tab:abl_backbone} with all four metrics. MPJPE-abs-p and MPJPE-p in mm, EPE2D-p in px, Jitter in mm/frame$^2$,
best values in bold. \ours is the most accurate in both settings and the \textbf{best} in every column except the in-the-wild
Jitter.}
\label{S-tab:supp_abl2_backbone}
\footnotesize
\setlength{\tabcolsep}{3pt}
\providecommand{\mh}[2]{#1#2}
\adjustbox{max width=\textwidth}{%
\begin{tabular*}{\linewidth}{@{\extracolsep{\fill}}lcccccccc@{}}
\toprule
& \multicolumn{4}{c}{In-domain} & \multicolumn{4}{c}{In-the-wild} \\
\cmidrule(lr){2-5}\cmidrule(lr){6-9}
Backbone & \mh{MPJPE}{-abs-p$\downarrow$} & \mh{EPE2D}{-p$\downarrow$} & \mh{MPJPE}{-p$\downarrow$} & \mh{Jitter}{$\downarrow$} & \mh{MPJPE}{-abs-p$\downarrow$} & \mh{EPE2D}{-p$\downarrow$} & \mh{MPJPE}{-p$\downarrow$} & \mh{Jitter}{$\downarrow$} \\
\midrule
DINOv3~\cite{S-simeoni2025dinov3} & 27.74 & 7.15 & 13.49 & 1.54 & 36.17 & 7.59 & 20.78 & 1.79 \\
V-JEPA 2~\cite{S-assran2025vjepa2} & 31.59 & 7.46 & 14.15 & 1.59 & 34.91 & 7.23 & 19.74 & 1.81 \\
Wan2.2, random weights & 31.62 & 7.57 & 19.35 & 1.53 & 41.49 & 8.12 & 23.93 & 1.70 \\
Wan2.2, no task tokens in the blocks & 20.50 & 5.56 & 11.73 & 1.30 & 34.52 & 6.72 & 20.37 & \best{1.60} \\
\textbf{Wan2.2, full (Ours)} & \best{19.62} & \best{4.67} & \best{9.50} & \best{1.26} & \best{32.59} & \best{6.50} & \best{18.59} & 1.61 \\
\bottomrule
\end{tabular*}
 }
\end{table}

\subsection{Clip-Level Ray Fitting}
All rows of Tab.~\ref{tab:abl_depth} start from \ours and differ in how the hand depth
is obtained. The rows ``Regressed depth'' and ``Regressed depth over focal length'' add a head that predicts the
logarithm of the wrist depth, directly or after division by the focal length, and the lateral position is solved at this depth.
The head is trained together with the features, so that the features can adapt to the regression, as in \ace, which regresses the depth end to end. With frozen features, the two heads reach 83.54 and 44.78\,mm in the wild, so adapting the features changes little. On the
features adapted to the log-depth head, the fit still reaches 19.37\,mm in-domain and 32.55\,mm in the wild. All other rows use the features, the 2D locations and the MANO hand of \ours unchanged.
``One bone, closed form'' places the wrist and the index knuckle on their joint rays~\cite{S-iqbal2018latent25d}.
``Reprojection fit, per frame'' estimates the MANO translation in every frame alone from the correspondences of its joints with
the 2D locations, as in MeTRAbs~\cite{S-sarandi2021metrabs} and HandDGP~\cite{S-valassakis2024handdgp}. Here, the least-squares
weights decrease with the variances of the 2D locations.
``Reprojection fit, whole clip'' minimizes \cref{eq:rayfit} without the bone term, with the reprojection and temporal
terms, the weights and the initialization of \ours. Its MPJPE-abs-p exceeds that of \ours on every dataset, by 1.73\,mm on ARCTIC,
1.56 on HOT3D, 0.32 on OakInk2, 0.21 on H2O, 0.20 on Ego-Std, 0.17 on Ego-Pro and 0.27 on \genhuman.
``Ray Fitting, per frame'' minimizes \cref{eq:rayfit} in every frame alone, with the bone and reprojection terms but
without the temporal term. ``Clip-Level Ray Fitting (Ours)'' is the fit of \cref{sec:depth} that \ours
uses, with all three terms of \cref{eq:rayfit} over the whole clip.

\begin{table}[H]
\centering
\caption{\textbf{Ablation of Clip-Level Ray Fitting with EPE2D-p.} The rows of Tab.~\ref{tab:abl_depth} with all four
metrics. MPJPE-abs-p and MPJPE-p in mm, EPE2D-p in px, Jitter in mm/frame$^2$, best values in bold. Clip-Level Ray Fitting has the lowest MPJPE-abs-p in both settings.}
\label{S-tab:supp_abl2_depth}
\footnotesize
\setlength{\tabcolsep}{3pt}
\providecommand{\mh}[2]{#1#2}
\adjustbox{max width=\textwidth}{%
\begin{tabular*}{\linewidth}{@{\extracolsep{\fill}}lcccccccc@{}}
\toprule
& \multicolumn{4}{c}{In-domain} & \multicolumn{4}{c}{In-the-wild} \\
\cmidrule(lr){2-5}\cmidrule(lr){6-9}
Hand depth & \mh{MPJPE}{-abs-p$\downarrow$} & \mh{EPE2D}{-p$\downarrow$} & \mh{MPJPE}{-p$\downarrow$} & \mh{Jitter}{$\downarrow$} & \mh{MPJPE}{-abs-p$\downarrow$} & \mh{EPE2D}{-p$\downarrow$} & \mh{MPJPE}{-p$\downarrow$} & \mh{Jitter}{$\downarrow$} \\
\midrule
Regressed depth & 25.81 & 4.91 & 9.64 & 1.43 & 81.24 & 8.33 & \best{18.26} & 2.12   \\
Regressed depth over focal length & 22.33 & 4.83 & 9.60 & 1.41 & 42.52 & \best{6.49} & 18.58 & 1.72   \\
\midrule
One bone, closed form & 23.82 & 5.40 & 9.61 & 4.73 & 38.83 & 7.18 & 18.68 & 4.91   \\
Reprojection fit, per frame & 20.66 & 4.70 & 9.52 & 3.55 & 34.40 & 6.60 & 18.62 & 3.72   \\
\rayfit, per frame & 20.24 & 4.68 & \best{9.50} & 3.39 & 32.69 & 6.53 & 18.59 & 3.65   \\
\midrule
Reprojection fit, whole clip & 20.32 & \best{4.67} & \best{9.50} & \best{1.24} & 32.86 & 6.50 & 18.60 & \best{1.58}   \\
\textbf{Clip-Level \rayfit (Ours)} & \best{19.62} & \best{4.67} & \best{9.50} & 1.26 & \best{32.59} & 6.50 & 18.59 & 1.61   \\
\bottomrule
\end{tabular*}
 }
\end{table}

\subsection{Joint-Heatmap Refiner}
The row ``Coarse, from token attention'' reads each 2D location from the attention of its token over the 32-pixel
tokens, without the heat maps. The row ``Inside the image only'' discards the 2D locations beyond the image border.
The row ``All joints (Ours)'' uses the 2D locations of \cref{eq:read} for every joint, also beyond the border, as \ours does.

\begin{table}[H]
\centering
\caption{\textbf{Ablation of the Joint-Heatmap Refiner with EPE2D-p.} The rows of Tab.~\ref{tab:abl_loc} with all four
metrics. MPJPE-abs-p and MPJPE-p in mm, EPE2D-p in px, Jitter in mm/frame$^2$, best values in bold. The 2D locations of all joints give the lowest MPJPE-abs-p and MPJPE-p in both settings, and the
coarse locations are smoother but far less accurate.}
\label{S-tab:supp_abl2_loc}
\footnotesize
\setlength{\tabcolsep}{3pt}
\providecommand{\mh}[2]{#1#2}
\adjustbox{max width=\textwidth}{%
\begin{tabular*}{\linewidth}{@{\extracolsep{\fill}}lcccccccc@{}}
\toprule
& \multicolumn{4}{c}{In-domain} & \multicolumn{4}{c}{In-the-wild} \\
\cmidrule(lr){2-5}\cmidrule(lr){6-9}
2D locations & \mh{MPJPE}{-abs-p$\downarrow$} & \mh{EPE2D}{-p$\downarrow$} & \mh{MPJPE}{-p$\downarrow$} & \mh{Jitter}{$\downarrow$} & \mh{MPJPE}{-abs-p$\downarrow$} & \mh{EPE2D}{-p$\downarrow$} & \mh{MPJPE}{-p$\downarrow$} & \mh{Jitter}{$\downarrow$} \\
\midrule
Coarse, from token attention & 32.30 & 5.42 & 9.63 & \best{0.97} & 39.71 & 6.65 & 18.65 & \best{1.19} \\
Inside the image only & 20.20 & 4.71 & 9.57 & 1.26 & 32.60 & \best{6.48} & \best{18.59} & 1.60 \\
\textbf{All joints (Ours)} & \best{19.62} & \best{4.67} & \best{9.50} & 1.26 & \best{32.59} & 6.50 & \best{18.59} & 1.61 \\
\bottomrule
\end{tabular*}
 }
\end{table}

\subsection{Confidence-Gated Fusion}
The rows of Tab.~\ref{tab:abl_fusion} use one model and differ only in the output joints. ``Joints only'' sets $g_j=1$ in
\cref{eq:lock} and outputs the back-projected 2D locations at the fitted depth. ``MANO only'' sets the gate to zero and
outputs the MANO joints. ``Gate equal to the confidence'' sets the gate to the confidence $c_j$ and keeps the rest of the fusion.
``Learned gate (Ours)'' is the fusion of \cref{S-sec:supp_fusion} that \ours uses. Beyond the image border, the 2D location is a
regression without image evidence and is less accurate than the MANO joint, so the learned gate keeps the MANO joint there, as the red box
in Fig.~\ref{fig:fusion} of the main paper shows. On the joints whose ground truth projects outside the image, the 2D error of the
detected hands is 20.10\,px for the 2D locations and 15.69\,px for the MANO joints in-domain, and 24.34 against 21.89\,px in the
wild.

\begin{table}[H]
\centering
\caption{\textbf{Ablation of Confidence-Gated Fusion with EPE2D-p.} The rows of Tab.~\ref{tab:abl_fusion} with all four
metrics. MPJPE-abs-p and MPJPE-p in mm, EPE2D-p in px, Jitter in mm/frame$^2$, best values in bold. The learned gate is the \textbf{best} on every accuracy metric in both settings.}
\label{S-tab:supp_abl2_fusion}
\footnotesize
\setlength{\tabcolsep}{3pt}
\providecommand{\mh}[2]{#1#2}
\adjustbox{max width=\textwidth}{%
\begin{tabular*}{\linewidth}{@{\extracolsep{\fill}}lcccccccc@{}}
\toprule
& \multicolumn{4}{c}{In-domain} & \multicolumn{4}{c}{In-the-wild} \\
\cmidrule(lr){2-5}\cmidrule(lr){6-9}
Output joints & \mh{MPJPE}{-abs-p$\downarrow$} & \mh{EPE2D}{-p$\downarrow$} & \mh{MPJPE}{-p$\downarrow$} & \mh{Jitter}{$\downarrow$} & \mh{MPJPE}{-abs-p$\downarrow$} & \mh{EPE2D}{-p$\downarrow$} & \mh{MPJPE}{-p$\downarrow$} & \mh{Jitter}{$\downarrow$} \\
\midrule
Joints only & 19.81 & 4.75 & 10.40 & 1.60 & 32.92 & 6.94 & 20.95 & 2.01 \\
MANO only & 19.79 & 5.03 & 9.75 & \best{1.20} & 32.62 & 6.54 & 18.71 & \best{1.59} \\
Gate equal to the confidence & 19.72 & 4.75 & 9.59 & 1.45 & 32.61 & 6.57 & 18.99 & 1.72 \\
\textbf{Learned gate (Ours)} & \best{19.62} & \best{4.67} & \best{9.50} & 1.26 & \best{32.59} & \best{6.50} & \best{18.59} & 1.61 \\
\bottomrule
\end{tabular*}
 }
\end{table}

\subsection{Autoregressive Rollout Adaptation}
\label{S-sec:supp_abl_ar}
\Cref{S-tab:supp_abl_ar} extends Tab.~\ref{tab:abl_ar} of the main paper with MPJPE-p. It compares the autoregressive chunks of
\ours with independent chunks, in which the same model processes each chunk without the cache of the previous one. Both variants
decode each chunk with its neighboring latent frames and end with the clip-level fit. Boundary jitter is the jitter of the
wrist-relative joints over the three-frame windows that span two chunks. The autoregressive chunks lower it from 0.98 to
0.94\,mm/frame$^2$ in-domain and from 1.34 to 1.29\,mm/frame$^2$ in the wild, and they are the best or tied in every other column.

\begin{table}[H]
\centering
\caption{\textbf{Ablation of Autoregressive Rollout Adaptation.} One model with and without the cache of the previous chunk.
MPJPE-abs-p and MPJPE-p in mm, Jitter and Boundary jitter in mm/frame$^2$, best values in bold. Boundary jitter is the jitter of
the wrist-relative joints over the windows that span two chunks. The autoregressive chunks are the \textbf{best} on every metric.}
\label{S-tab:supp_abl_ar}
\footnotesize
\setlength{\tabcolsep}{3pt}
\adjustbox{max width=\textwidth}{%
\begin{tabular}{@{}lcccccccc@{}}
\toprule
& \multicolumn{4}{c}{In-domain} & \multicolumn{4}{c}{In-the-wild} \\
\cmidrule(lr){2-5}\cmidrule(lr){6-9}
Chunks & MPJPE-abs-p$\downarrow$ & MPJPE-p$\downarrow$ & Jitter$\downarrow$ & Boundary jitter$\downarrow$ & MPJPE-abs-p$\downarrow$ & MPJPE-p$\downarrow$ & Jitter$\downarrow$ & Boundary jitter$\downarrow$ \\
\midrule
Independent & \best{19.62} & 9.53 & 1.27 & 0.98 & 32.62 & 18.64 & \best{1.61} & 1.34 \\
\textbf{Autoregressive (Ours)} & \best{19.62} & \best{9.50} & \best{1.26} & \best{0.94} & \best{32.59} & \best{18.59} & \best{1.61} & \best{1.29} \\
\bottomrule
\end{tabular}
 }
\end{table}

\subsection{Sensitivity to the Focal Length}
\label{S-sec:supp_abl_focal_sec}
Tab.~\ref{tab:focal} of the main paper rescales the test videos of all seven datasets about the principal point by a factor
between 0.60 and 1.60 and gives the rescaled intrinsics to every method. The rescaling leaves the 3D ground truth unchanged. Every
factor is scored as in Tab.~\ref{tab:main} on the hands of that table, so the column of the factor 1.00 repeats
Tab.~\ref{tab:main}. A hand that leaves the image when the video is enlarged is not counted, and a hand that projects only onto the padded border when the video is reduced is not added, as in the evaluation of
\cref{S-sec:supp_protocol}. The in-domain block is the mean of the six in-domain datasets. Over
all factors, the MPJPE-abs-p of \ours stays between 19.62 and 37.96\,mm in-domain and between 29.71 and 45.79\,mm in the wild,
whereas that of \ace ranges from 24.46 to 178.86\,mm in-domain and from 42.59 to 314.15\,mm in the wild, and that of \vidihand
from 36.60 to 107.65\,mm in-domain and from 57.07 to 191.34\,mm in the wild. In the wild, the two baselines improve when the videos are enlarged: the focal length of
the \genhuman camera is 0.32 of the image width, shorter than that of every training camera, 0.41 to 0.86, and a larger factor
moves it into the range of the training cameras, as expected of a regressed depth that depends on the training cameras.
\ours is the most accurate at every factor in both settings.

\subsection{Presence Threshold}
\label{S-sec:supp_abl_thresh}
The hand presence confidence threshold of \ours is 0.05, and the same threshold selects the hand-frames that the fit covers.
\Cref{S-tab:supp_abl_thresh} serves the same model with thresholds from 0.05 to 0.20 on the full test sets. The results are
insensitive to the threshold. The MPJPE-abs-p changes by at most 0.24\,mm in-domain and 0.20\,mm in the wild, the
FAcc by at most 0.09 points, the EPE2D-p by at most 0.03\,px, the MPJPE-p by at most 0.06\,mm and the Jitter by at most 0.01\,mm/frame$^2$.

\begin{table}[H]
\centering
\caption{\textbf{Presence threshold.} \ours with four hand presence confidence thresholds on the full test sets. FAcc in \%,
MPJPE-abs-p and MPJPE-p in mm, EPE2D-p in px, Jitter in mm/frame$^2$, best values in bold. The results are
insensitive to the threshold.}
\label{S-tab:supp_abl_thresh}
\footnotesize
\setlength{\tabcolsep}{3pt}
\adjustbox{max width=\textwidth}{%
\begin{tabular}{@{}lcccccccccc@{}}
\toprule
& \multicolumn{5}{c}{In-domain} & \multicolumn{5}{c}{In-the-wild} \\
\cmidrule(lr){2-6}\cmidrule(lr){7-11}
Threshold & FAcc$\uparrow$ & MPJPE-abs-p$\downarrow$ & EPE2D-p$\downarrow$ & MPJPE-p$\downarrow$ & Jitter$\downarrow$ & FAcc$\uparrow$ & MPJPE-abs-p$\downarrow$ & EPE2D-p$\downarrow$ & MPJPE-p$\downarrow$ & Jitter$\downarrow$ \\
\midrule
\textbf{0.05 (Ours)} & \best{99.93} & \best{19.62} & \best{4.67} & \best{9.50} & \best{1.26} & \best{99.89} & \best{32.59} & \best{6.50} & \best{18.59} & 1.61 \\
0.10 & 99.90 & 19.71 & \best{4.67} & 9.52 & \best{1.26} & 99.85 & 32.69 & 6.52 & 18.60 & \best{1.60} \\
0.15 & 99.87 & 19.79 & 4.68 & 9.54 & \best{1.26} & 99.82 & 32.74 & 6.53 & 18.61 & \best{1.60} \\
0.20 & 99.84 & 19.86 & 4.68 & 9.56 & \best{1.26} & 99.80 & 32.79 & 6.53 & 18.62 & \best{1.60} \\
\bottomrule
\end{tabular}
 }
\end{table}

\section{Runtime}
\label{S-sec:supp_speed}
We time every method on one exclusive A100-SXM4-80GB GPU with 14 CPU threads. All methods process
the same 400-frame EgoDemo Standard test clip at $608{\times}480$, with two hands in every frame. For timing only, hands that the detector fails to detect are
supplemented with boxes from the ground-truth 2D joint locations. The measured time spans from the decoded frames and intrinsics to the joints and
excludes video decoding, file I/O, model loading, warm-up and SLAM-based camera estimation. We measure offline FPS at the fastest
tested batch size and report the median of three runs. \Cref{S-tab:supp_speed} lists every method of Tab.~\ref{tab:main}.
\ours runs at 57.23\,FPS with 14.79\,GiB peak memory, second only to HandDGP.

\begin{table}[H]
\centering
\caption{\textbf{Runtime.} Offline speed in FPS on one A100 GPU for a 400-frame clip at $608{\times}480$ with two hands in every
frame, with the methods in the order of Tab.~\ref{tab:main} and the best value in bold. \ours runs at 57.23\,FPS, second only to
HandDGP.}
\label{S-tab:supp_speed}
\setlength{\tabcolsep}{6pt}
\begin{tabular}{@{}lc@{\hspace{18pt}}lc@{\hspace{18pt}}lc@{\hspace{18pt}}lc@{}}
\toprule
Method & FPS$\uparrow$ & Method & FPS$\uparrow$ & Method & FPS$\uparrow$ & Method & FPS$\uparrow$ \\
\midrule
Hamba~\cite{S-dong2024hamba} & 1.42 & EgoForce~\cite{S-millerdurai2026egoforce} & 9.91 & OmniHands~\cite{S-lin2026omnihands} & 1.59 & InfiniHand~\cite{S-ren2026infinihand} & 7.78 \\
WildHands~\cite{S-prakash2024wildhands} & 1.64 & HandDGP~\cite{S-valassakis2024handdgp} & \best{68.21} & HaWoR~\cite{S-zhang2025hawor} & 8.84 & ViDiHand~\cite{S-wang2026vidihand} & 2.55 \\
InterWild~\cite{S-moon2023interwild} & 24.85 & EgoHandICL~\cite{S-xie2026egohandicl} & 3.46 & HandFlow~\cite{S-xu2026handflow} & 24.43 & ACE-Ego-Hand~\cite{S-liu2026aceegohand} & 56.75 \\
AnyHand~\cite{S-si2026anyhand} & 14.34 & WiLoR~\cite{S-potamias2025wilor} & 21.30 & MINT~\cite{S-zhu2026mint} & 23.10 & \textbf{\ours (Ours)} & 57.23 \\
HaMeR~\cite{S-pavlakos2024hamer} & 1.68 & Dyn-HaMR~\cite{S-yu2025dynhamr} & 6.46 & PAD-Hand~\cite{S-ismayilzada2026padhand} & 11.68 &  &  \\
\bottomrule
\end{tabular}
 \end{table}
 {
    \small
    \bibliographystyle{ieeenat_fullname}

}

\end{document}